\documentclass{article}

\PassOptionsToPackage{sort&compress,numbers}{natbib}

\usepackage[final]{neurips_2021} 
\usepackage[utf8]{inputenc}
\usepackage[T1]{fontenc}
\usepackage{graphics,graphicx,caption,float,subcaption,booktabs,multirow,array,color,ifthen,tabu,colortbl,dblfloatfix,url,xparse,mathtools,patchcmd,algorithm,algorithmic,amssymb,xspace,nicefrac,microtype,amsmath,amsthm,amsfonts,bm,ragged2e,tikz,stackengine,etoolbox,xpatch,enumerate,xstring,setspace,tabularx,makecell,changepage,cuted,titlesec,enumitem,wrapfig,tcolorbox,xfrac}
\usepackage[pagebackref=true,breaklinks=true,colorlinks=true,bookmarks=false,citecolor=blue]{hyperref}
\usepackage[capitalise,noabbrev,nameinlink]{cleveref}
\usepackage[english]{babel}

\usepackage{diagbox}
\usepackage{multirow}
\usepackage{colortbl}

\definecolor{colorcbet}{rgb}{0.12156862745098039, 0.4666666666666667, 0.7058823529411765}
\definecolor{colorcbet_tabula}{rgb}{0.12156862745098039, 0.4666666666666667, 0.7058823529411765}
\definecolor{colorcount}{rgb}{0.5803921568627451, 0.403921568627451, 0.7411764705882353}
\definecolor{colorride}{rgb}{1.0, 0.4980392156862745, 0.054901960784313725}
\definecolor{colorcuriosity}{rgb}{0.17254901960784313, 0.6274509803921569, 0.17254901960784313}
\definecolor{colorrnd}{rgb}{0.4980392156862745, 0.4980392156862745, 0.4980392156862745}
\definecolor{colorvanilla}{rgb}{0.0, 0.0, 0.0}
\definecolor{colorrandom}{rgb}{0.0, 0.0, 0.0}
\definecolor{colorcbet1}{rgb}{1.0, 0.0, 0.0}
\definecolor{colorcbet2}{rgb}{1.0, 0.8431372549019608, 0.0}
\definecolor{colorcbet3}{rgb}{0.0, 0.39215686274509803, 0.0}
\definecolor{colorcbet4}{rgb}{0.5803921568627451, 0.403921568627451, 0.7411764705882353}
\definecolor{colorcbet49}{rgb}{0.5803921568627451, 0.403921568627451, 0.7411764705882353}
\definecolor{colorcbet5}{rgb}{0.5490196078431373, 0.33725490196078434, 0.29411764705882354}
\definecolor{colorcbet6}{rgb}{0.7803921568627451, 0.08235294117647059, 0.5215686274509804}
\definecolor{colorcbet7}{rgb}{0.12156862745098039, 0.4666666666666667, 0.7058823529411765}
\definecolor{colorcbet79}{rgb}{0.12156862745098039, 0.4666666666666667, 0.7058823529411765}
\definecolor{colorcbet8}{rgb}{1.0, 1.0, 0.0}
\definecolor{colorcbet9}{rgb}{1.0, 0.0, 1.0}
\definecolor{colorcbet10}{rgb}{0.5803921568627451, 0.403921568627451, 0.7411764705882353}
\definecolor{colorcbet11}{rgb}{1.0, 1.0, 0.0}
\definecolor{colorcbet13}{rgb}{1.0, 0.0, 1.0}

\definecolor{colora0}{rgb}{0.00392156862745098, 0.45098039215686275, 0.6980392156862745}
\definecolor{colora1}{rgb}{0.8705882352941177, 0.5607843137254902, 0.0196078431372549}
\definecolor{colora2}{rgb}{0.00784313725490196, 0.6196078431372549, 0.45098039215686275}
\definecolor{colora3}{rgb}{0.8352941176470589, 0.3686274509803922, 0.0}
\definecolor{colora4}{rgb}{0.8, 0.47058823529411764, 0.7372549019607844}
\definecolor{colora5}{rgb}{0.792156862745098, 0.5686274509803921, 0.3803921568627451}
\definecolor{colora6}{rgb}{0.984313725490196, 0.6862745098039216, 0.8941176470588236}

\usepackage{xifthen}
\usepackage{amsmath}
\usepackage{amsfonts}

\renewcommand{\vec}[1]{{\boldsymbol{#1}}}

\usepackage{color}
\newcommand{\textsub}[1]{\textsc{\tiny #1}} 

\usepackage{pifont}
\makeatletter
\newcommand{\colorlabel}[1]{%
	\global\tag@true%
	\nonumber%
	\refstepcounter{equation}%
	\gdef\df@tag{\maketag@@@{{\color{#1}(\theequation)}}\def\@currentlabel{\theequation}}}
\makeatother

\newcommand{\bigmid}{\;\ifnum\currentgrouptype=16 \middle\fi|\;} 
\newcommand{\mytilde}[0]{\mathds{\raise.17ex\hbox{$\scriptstyle\sim$}}} 

\newcommand{\action}{a}
\newcommand{\state}{s}
\newcommand{\params}{\theta}

\newcommand{\context}{c}

\providecommand{\rwd}{\mathcal{R}}
\providecommand{\prob}{\mathcal{P}}

\newcommand{\rmodel}[1][]{%
	\ifthenelse{\equal{#1}{}}{\rwd\left(\state,\action\right)}{\rwd\left(\state^{[#1]},\action^{[#1]}\right)}%
}
\newcommand{\probmodel}[1][]{%
	\ifthenelse{\equal{#1}{}}{\prob\left(\state'|\state,\action\right)}{\prob\left(\state^{[#1+1]}|\state^{[#1]},\action^{[#1]}\right)}%
}

\newcommand{\rmodelctx}[1][]{%
	\ifthenelse{\equal{#1}{}}{\rwd\left(\context,\params\right)}{\rwd\left(\context^{[#1]},\params^{[#1]}\right)}%
}
\newcommand{\berror}[1][]{%
	\ifthenelse{\equal{#1}{}}{\delta\left(\state,\action\right)}{\delta_#1\left(\state,\action\right)}%
}

\newcommand{\vecrmodel}[1][]{%
	\ifthenelse{\equal{#1}{}}{\vec\rwd\left(\state,\action\right)}{\rwd\left(\state^{[#1]},\action^{[#1]}\right)}%
}

\usepackage{cellspace}

\title{Interesting Object, Curious Agent: \\Learning Task-Agnostic Exploration}

\newcommand*\samethanks[1][\value{footnote}]{\footnotemark[#1]}

\author{%
  Simone Parisi\normalfont{\textsuperscript{1}}\thanks{Equal contribution. Contacts: \texttt{sparisi@fb.com} and  \texttt{vdean@cmu.edu}}
  \space\space\space\space\space\space\space\space\space\space
  \textbf{Victoria Dean}\normalfont{\textsuperscript{2}}\samethanks
  \space\space\space\space\space\space\space\space\space\space
  \textbf{Deepak Pathak}\normalfont{\textsuperscript{2}}
  \space\space\space\space\space\space\space\space\space\space
  \textbf{Abhinav Gupta}\normalfont{\textsuperscript{1}}\\
  \normalfont{
  \textsuperscript{1}Facebook AI Research
  \space\space\space\space\space\space\space\space\space\space
  \textsuperscript{2}Carnegie Mellon University
  }\\
  \space\space\space\space\space\space\space\space\space\space
}

\begin{document}

\maketitle

\begin{abstract}
Common approaches for task-agnostic exploration learn tabula-rasa --the agent assumes isolated environments and no prior knowledge or experience. However, in the real world, agents learn in many environments and always come with prior experiences as they explore new ones. Exploration is a lifelong process. In this paper, we propose a paradigm change in the formulation and evaluation of task-agnostic exploration. In this setup, the agent first \textit{learns to explore} across many environments without any extrinsic goal in a task-agnostic manner.
Later on, the agent effectively transfers the learned \textit{exploration policy} to better explore new environments when solving tasks. In this context, we evaluate several baseline exploration strategies and present a simple yet effective approach to learning task-agnostic exploration policies. Our key idea is that there are two components of exploration: (1) an agent-centric component encouraging exploration of unseen parts of the environment based on an agent’s belief; (2) an environment-centric component encouraging exploration of inherently interesting objects. We show that our formulation is effective and provides the most consistent exploration across several training-testing environment pairs. We also introduce benchmarks and metrics for evaluating task-agnostic exploration strategies. The source code is available at \url{https://github.com/sparisi/cbet/}.
\end{abstract}

\section{Introduction}
\label{sec:intro}
Exploration is one of the key unsolved problems in building intelligent agents capable of behaving like humans. In reinforcement learning (RL), exploration is usually studied under two different settings. The first is task-driven exploration, where the reward is well-defined and the agent's goal is to explore in order to maximize long-term rewards. However, in real life, external rewards are either sparse or unknown altogether. In this setting, exploration is task-agnostic: given a new environment, the agent has to explore it in absence of any external reward. Common approaches to encourage task-agnostic exploration use intrinsically motivated rewards such as prediction curiosity~\citep{schmidhuber1991possibility,pathak2017curiosity}, empowerment~\citep{rezende2015variational}, or visitation counts~\citep{bellemare2016unifying,ostrovski2017count}. 
But does this setup represent how humans explore?

We argue that the commonly-used task-agnostic exploration setup is unrealistic, both from practical and academic viewpoints. This setup assumes environments in isolation and agents exploring tabula-rasa, i.e., with no prior knowledge or experience. By contrast, we as humans do not learn from one environment in isolation and we do not throw away our past knowledge every time we encounter a new environment~\citep{dubey2018investigating}.
Exploration is rather a lifelong process: every time we encounter new environments, we use our prior knowledge and experience to develop new efficient exploration strategies. In this paper, we view the exploration problem from a continual learning lens. More specifically, in this setup, the learning agent interacts with one or many environments without any extrinsic goal. At this time, the agent \textit{learns to explore} the environments. Later on, the agent effectively transfers the learned \textit{exploration policy} to explore new environments, rather than exploring the new environment tabula-rasa. 

\begin{wrapfigure}{l}{0.45\textwidth}
\vspace{-0.5em}
\begin{center}
\includegraphics[width=0.98\linewidth]{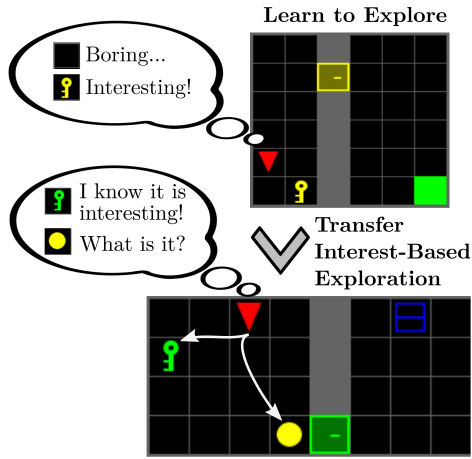}
\end{center}
\vspace{-0.7em}
\caption{\label{fig:teaser}\textbf{Change-Based Exploration Transfer (C-BET)} trains task-agnostic exploration agents that transfer to new environments. Here the agent learns that keys are interesting, as they allow further interaction with the environment (opening doors). Later, when tasked with reaching a box behind a door, the agent starts by picking up the key.}
\vspace{-1.5em}
\end{wrapfigure}

A key question in learning how to explore is what to learn and how to transfer prior knowledge from one environment to another. Most existing task-agnostic exploration approaches, such as visitation counts, curiosity, or empowerment, define intrinsic rewards in an \textit{agent-centric} manner: they encourage exploration of unseen parts of the environment based on the agent's own belief. In these approaches, exploration is driven by what the agent knows about the world.
However, most do not make a distinction between what the agent believes it is interested in and states that would make any agent interested.
For example, if the agent uses a visitation count model and has seen many objects of one kind in one environment, it would not explore the same type of objects again in a new environment. This seems to be in stark contrast to how humans
explore. Consider a switch with a bell sign. Even though we might have pressed hundreds of doorbell switches (and even this instance), we are still attracted to press it. Some objects in the world just demand curiosity. We argue that apart from an `agent-centric' component, there is an `environment-centric' component to exploration, which can be learned from prior knowledge and experiences.

In this paper, we propose a paradigm change to move away from stand-alone isolated task-agnostic environment exploration to a more realistic multi-environment transfer-exploration setup\footnote{While it can be argued that the real world has no explicit distinction between training and testing, we use this dichotomy only for the purpose of evaluation.}. We show how to learn exploration policies both from single- and multi-environment interaction, and how to transfer them to unseen environments. This transfer-exploration setup allows agents to use prior experiences for learning task-agnostic exploration. 
Notably, classic stand-alone task-agnostic approaches were designed for tabula-rasa exploration and hence only explore in an agent-centric manner. They fail to capture the inherent interestingness of some environment components. With this insight, we propose \textit{Change-Based Exploration Transfer (C-BET)}, a simple yet effective approach learning joint agent-centric and environment-centric exploration. The key idea is for an agent to seek out both surprises (unseen areas) and high-impact (interesting) components of the environment. The experiments show that C-BET (a) learns more effectively when placed in a multi-environment setup, and (b) either outperforms or performs competitively with prior methods across several unseen testing environments.
We hope this paper will inspire exploration research to focus more on learning from multiple environments and transferring experiences rather than tabula-rasa exploration.

\section{Preliminaries and Related Work}
\label{sec:pre}
We consider environments governed by Markov Decision Processes (MDPs). In MDPs, an agent observes the state of the environment $s$ and selects actions $a$ according to a policy $\pi(a|s)$. In turn, the environment changes, providing a new observation $s'$ and a reward $r$. Through environment interaction, the agent collects episodes, i.e., sequences of states, actions and rewards $(\state_t, \action_t, r_t)_{t=1\ldots T}$.
The goal of RL is to learn a policy maximizing the sum of rewards during episodes, i.e., {the return}.
%
\\
In this setting, exploration poses many questions. If the environment provides 
no rewards, what should the agent look for? When should it act greedily with respect to the rewards it has found and stop looking for more? In the history of RL, many approaches have been proposed to tackle these questions. On one hand, classic single-environment approaches range from intrinsic motivation with visitation counts~\citep{auer2002finite,strehl2008analysis,bellemare2016unifying,jin2018iq,dong2020q}, optimism~\citep{lai1985asymptotically,kearns2002near,brafman2002r,auer2007logarithmic,jaksch2010near}, or curiosity~\citep{ryan2000intrinsic,stadie2015incentivizing,houthooft2016vime,pathak2017curiosity,burda2018exploration,schultheis2020receding}, to bootstrapping~\citep{osband2019deep,deramo2019exploiting} or empowerment~\citep{klyubin2005all,rezende2015variational}. 
On the other hand, we find approaches to incrementally learn tasks, such as transfer learning~\citep{weiss2016survey}, continual learning~\citep{kirkpatrick2017overcoming}, curriculum learning~\citep{narvekar2020curriculum}, and meta learning~\citep{rakelly2019efficient}.
Below, we review approaches closely related to ours.

\paragraph{Intrinsic motivation.}
Exploration strategies relying on {intrinsic rewards} date back to~\citet{schmidhuber1991possibility}, who proposed to encourage exploration by visiting hard-to-predict states.
More recently, the idea of auxiliary rewards to make up for the lack of external rewards has been extensively studied in RL, supported by evidence from psychology and neuroscience~\citep{gottlieb2013information}. Several intrinsic rewards have been proposed, ranging from visitation count bonuses~\citep{strehl2008analysis,bellemare2016unifying} to bonuses based on prediction error of some quantity. For example, the agent may learn a dynamics model and try to predict the next state \citep{stadie2015incentivizing,houthooft2016vime,pathak2017curiosity,schmidhuber2006developmental}. By giving a bonus proportional to the prediction error, the agent is incentivized to explore unpredictable states. 
%
\citet{schultheis2020receding}, instead, proposed to learn intrinsic rewards function by maximizing extrinsic rewards by meta-gradient.
\\[0.2em]
However, in these approaches exploration is agent-centric, i.e., based on an agent’s belief such as the forward model error. 
In contrast, with this work we propose additionally learning \textit{environment-centric} exploration policies. C-BET neither requires a model nor knowledge of extrinsic rewards. Instead, it encourages the agent to perform actions causing \textit{interesting changes} to the environment.
We should note that while~\citet{raileanu2020ride} proposed a similar approach, their exploration policy lacks the transfer component and also requires to learn models.
\paragraph{Transfer learning.}
The idea of agents capable of incrementally learning tasks is well-known in the field of machine learning, with the first approaches dating back to the 90s'~\citep{ring1994continual,thrun1995lifelong,ring1998child}.
In RL, recent methods have focused on policy and feature transfer. In the former, a pre-trained agent (teacher) is used to transfer behaviors to a new agent (student). Examples include policy distillation, where the student is trained to minimize the Kullback-Leibler divergence to the teacher~\citep{rusu2015policy} or to multiple teachers at the same time~\citep{teh2017distral}. Alternative approaches, instead, directly reuse policies from source tasks to build the student policy~\citep{hailu1999amount,fernandez2006probabilistic,barreto2016successor}.
In feature transfer, a pre-learned state representation is used to encourage exploration when tasks are presented to the agent~\citep{hansen2020fast,yarats2021reinforcement}.
Similar to transfer RL, continual RL studies how learning on one or more tasks can help accelerate learning on different tasks, and how to prevent catastrophic forgetting~\citep{kirkpatrick2017overcoming, schwarz2018progress, rolnick2019experience}. 
Meta RL, instead, tries to exploit underlying common structures between tasks to learn new tasks more
quickly~\citep{finn2017model,rakelly2019efficient}. 
\\[0.2em]
However, the setup in these approaches is not task-agnostic, i.e., task-specific policies are transferred rather than exploration policies.
For example, after learning a policy maximizing the rewards of one task, the agent starts exploring guided by the same policy as a second task is given. 
Transfer is task-centric rather than task-agnostic and environment-centric.
Consequently, if tasks are too dissimilar information cannot be reused, even if the environments are similar.
By contrast, in this work we propose learning task-agnostic exploration from one or many environments and show transfer to unseen environments. We should note that while~\citet{pathak2017curiosity} did demonstrate fine-tuning on different maze maps, their focus and large-scale evaluations remain on tabula-rasa exploration.

\begin{figure*}[t]
\centering
\vspace*{-2em}
\begin{minipage}[t]{.49\linewidth}
    \centering
    \textbf{\small Exploration Training}\\[0.3em]
    \includegraphics[scale=.63]{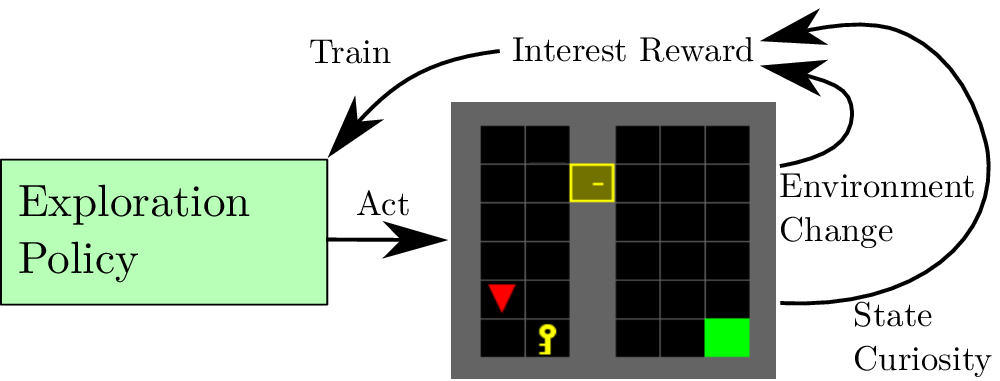}
    \caption{\label{fig:exp_training}
    \textbf{C-BET pre-training.} Our agent interacts with environments and learns using intrinsic rewards computed from state and change counts.}
\end{minipage}\hfill
\begin{minipage}[t]{.49\textwidth}
    \centering
    \textbf{\small Exploration Transfer}\\[0.3em]
    \includegraphics[scale=0.63]{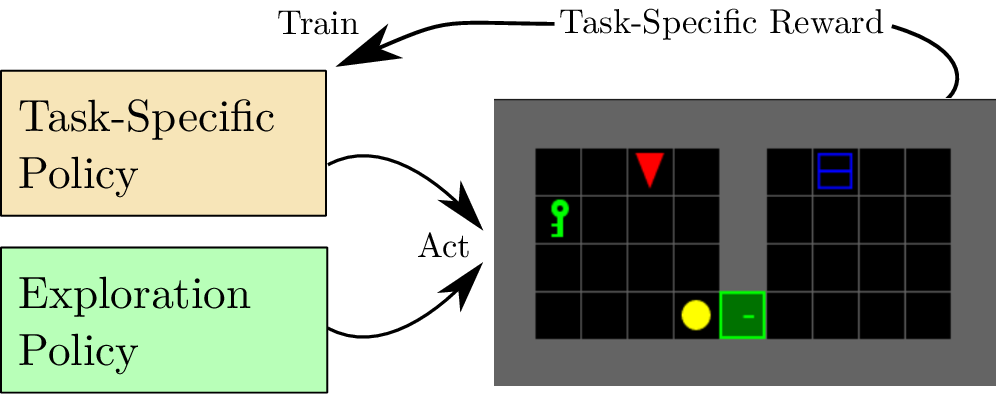}
    \vspace*{-2pt}
    \caption{\label{fig:exp_transfer}
    \textbf{C-BET transfer.} The pre-trained exploration policy is fixed and guides task-specific policy learning in new environments.}
\end{minipage}
\end{figure*}

\section{Learning to Explore}
\label{sec:method}
Our goal is to decouple the environment-centric nature of exploration from its agent-centric component. Contrary to prior work, we propose to first learn an environment-centric exploration policy and then to transfer it to unseen environments. The policy is driven by the inherent interestingness of states and is learned over time via interaction.
First, during a pre-training phase, the agent interacts with many environments without any tasks and learns an exploration policy. Then, when new environments and tasks are presented, the agent uses the previously learned policy to explore more efficiently and learn task-specific policies. C-BET's key components are (1) a novel intrinsic reward and the learning of a policy to disentangle exploration from exploitation, and (2) the use of such policy to help exploration for new tasks. Figures~\ref{fig:exp_training} and~\ref{fig:exp_transfer} summarize our framework.

We should note that \citet{rajendran2020should} also proposed a transfer framework based on intrinsic rewards. In their work, the agent switches between practice episodes --where the agent receives only intrinsic rewards-- and match episodes --giving only extrinsic rewards. However, practice episodes are simpler variations of match episodes (e.g., in Atari Pong the agent practices against itself) rather than different tasks as in C-BET. Furthermore, the intrinsic reward used in practice episodes is given by a function trained with meta-gradients to improve the extrinsic-reward return. That is, exploration is not task-agnostic as in C-BET, and extrinsic rewards are the main drive of the agent.

\subsection{Interestingness of State-Action Pairs}
\label{subsec:interest}
\begin{figure}[t]
\begin{minipage}[c]{0.55\textwidth}
\vspace{-1.5em}
\textbf{\tiny \hspace{6.7em} Turn Left \hspace{2.7em} Turn Right \hspace{1.9em} Move Forward \hspace{2.2em} Pick Up}
\vspace{-0.5em}
\begin{center}
	\raisebox{2.2em}{\small 
	$\frac{1}{\sqrt{N(s')}}$}\hfill
	\begin{subfigure}[b]{.86\linewidth} 
		\centering 
		\includegraphics[width=\linewidth]{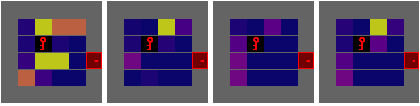}
	\end{subfigure}
	\raisebox{2.2em}{\small $\frac{||c||^2}{\sqrt{N(s')}}$}\hfill
	\begin{subfigure}[b]{.86\linewidth} 
		\centering 
		\includegraphics[width=\linewidth]{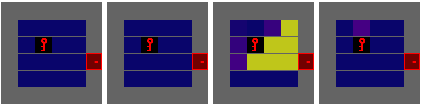}
	\end{subfigure} 
	\raisebox{2em}{\small
	Eq.~\eqref{eq:interest}
	}\hfill
	\begin{subfigure}[b]{.86\linewidth} 
		\centering 
		\includegraphics[width=\linewidth]{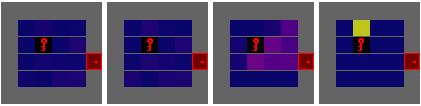}
	\end{subfigure} 
\end{center}
\vspace*{-0.5em}
\textbf{\tiny Gridworld with a key and a door. Observations encode cells depending on their content \\[-1em](e.g., 5 for the key, 10 for the agent). 
In each cell the agent is facing downward, and \\[-1em]can pick up the key only from the cell above it.
Samples have been collected randomly.
}
\end{minipage}
\hfill
\begin{minipage}[c]{0.43\textwidth}
\caption{\label{fig:interest_rwd}\textbf{Visualization of intrinsic rewards (row) for the agent's actions (column).}
Brighter color denotes higher reward. 
Rewarding only state counts (top) does not provide useful feedback, and going to the corners is valued more than picking up the key. 
With the L2 norm of state changes (middle), the agent is biased in favor of moving, because its position is encoded with the highest value in the observation space. The resulting policy will prefer to navigate without picking up the key. 
By contrast, C-BET (bottom) gives picking up the key the highest reward.}
\end{minipage}
\end{figure}

The natural world is filled with states or scenarios that are inherently interesting and our goal is to capture this inherent interestingness via intrinsic rewards. In this paper, we propose adding an environment-centric component of interestingness to the existing agent-centric component of surprise. Specifically, we hypothesize that the environment can \textit{change} on interaction, and the changes that are \textit{rare} are inherently interesting. That is, we penalize actions not affecting the environment, and favor actions producing rare changes. For instance, moving around, bumping into walls, or trying to open locked doors without keys all result in no change and thus will be of low interest. 
\\[0em]
We also want to keep the agent-centric component in exploration --that is, the exploration policy should look for surprises or unseen states. Thus, we further reward actions leading to less-visited states. By combining these two components, the resulting C-BET interest-based reward is
\begin{equation}
    r_i(s,a,s') = 1 / (N(s') + N(c)), \label{eq:interest}
\end{equation}
where $c(s,s')$ defines the \textit{environment change} of a transition $(s,a,s')$, and $N$ denotes (pseudo)counts of changes and states.
Figure~\ref{fig:interest_rwd} empirically shows its effectiveness. 
In Section~\ref{sec:eval} we discuss change encodings used in our experiments. 


\subsection{Exploration Learning}
\label{subsec:pretrain}
In this phase, we want to learn task-agnostic exploration policies from interaction with many environments. The agent has no goal, but states where it can `die' are still terminal.
In this setting, we would like to treat the problem of learning exploration as an MDP with intrinsic-rewards only, and train the agent to maximize discounted intrinsic-returns averaged over episodes. 
\\
Formally, the agent explores many environments $\mathcal{E}_\textsub{exp} = \{E_1, E_2, \ldots, E_N\}$, each governed by MDP $\langle S_n, A, P, r_i,\gamma_i\rangle$. That is, each environment has its own states but the action space is the same, and all environments obey the same dynamics $P$ and the same intrinsic reward function $r_i$. 
The agent's goal is to learn an exploration policy maximizing the sum of discounted intrinsic rewards, i.e., $\pi_\textsub{exp}(s,a) = \arg\max_\pi \mathbb{E}_{\mathcal{E},\pi}[\sum_t \gamma_i^t r_i(s_t,a_t)]$.
To approximate the average, after a maximum number of steps 
the environment is reset and a new episode starts, as typically done in RL.
\\
However, both common~\citep{pathak2017curiosity,burda2018exploration,raileanu2020ride} and Eq.~\eqref{eq:interest} intrinsic rewards decrease over time as the agent explores, to the point that they vanish to zero given enough samples. For instance, counts will grow to infinity, or prediction models error will go to zero. While this is not an issue in the tabula-rasa setup where the agent also gets extrinsic rewards, it can be problematic in the proposed task-agnostic exploration framework. Any policy, indeed, would be optimal if all rewards are zero.

To prevent 
Eq.~\eqref{eq:interest} from vanishing, we randomly reset counts any given time step.
To explain why resets need to be random, we start by considering `episodic counts' proposed by~\citet{raileanu2020ride}. 
These counts are reset at the beginning of every episode to ensure that the agent does not go back and forth between a sequence of states with high rewards.
While this work fine when extrinsic rewards are also given, it can be a problem if we learn only on intrinsic rewards.
When counts are reset, the agent `forgets' past trajectories and thinks that every state and change is new. 
If resets always happen at the end of an episode, then initial states will always get higher reward. Moreover, starting always with zero-counts may favor some trajectories and penalize others.
\begin{wrapfigure}{l}{0.345\textwidth}
\vspace{-1.3em}
\begin{center}
\includegraphics[width=0.99\linewidth]{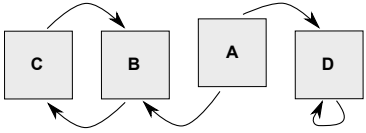}
\end{center}
\vspace{-0.9em}
\caption{\label{fig:chain}This chainworld illustrates that if counts are resets at the beginning of every episode, the learned policy will never visit D.}
\vspace{-0.9em}
\end{wrapfigure} 
Consider for example the chainworld in Figure~\ref{fig:chain}. The agent always starts in A, from where it can go to B or D. From B, it loops between B and C. From D, it cannot go anywhere else. The optimal exploration policy should visit all states uniformly, by randomly going to B and D. However, if we reset counts at every episode the agent forgets that it has already visited B and C. Thus, the intrinsic rewards for B and C are high again, and trajectory ABCBCBC... gives higher intrinsic return than ADDD... Consequently, the optimal policy with respect to episodic counts will always prefer to visit B rather than D.

The optimal exploration policy, instead, should have some randomness to visit the environment uniformly, while prioritizing interesting states.
For this reason, we propose to reset counts at any given step with probability $p$. When a new episode starts, counts may not be reset yet so the agent remembers what it has visited before. As the agent explores, on average common states and changes will have higher count more often, and the agent will correctly prefer rarer ones.
%
In this paper, we propose $p \leq 1 - \gamma_i$ where $\gamma_i$ is the intrinsic reward discount factor. This is a fitting choice because in an MDP the sum of discounted rewards can be interpreted as the expected sum of undiscounted rewards if every time step had a $1 - \gamma_i$ probability of ending. Intuitively, this means that $\gamma_i$ implies a `life expectancy' of $1 / (1 - \gamma_i)$ steps, and thus resets should not happen more frequently than that.

The resulting MDP with Eq.~\eqref{eq:interest} rewards and random count resets can be solved by any RL algorithm. However, we should note that this MDP is non-stationary, because the agent may receive different rewards for the same state, depending on how many times the state has been visited in the past.
Nonetheless, classic intrinsic rewards --even in tabula-rasa exploration-- either based on prediction errors~\citep{pathak2017curiosity,raileanu2020ride} or counts~\citep{bellemare2016unifying} 
also introduce non-stationarity because they change over time as well.
In practice, this non-stationarity is not an issue because intrinsic rewards change slowly over time.

\subsection{Exploration Transfer}
\label{subsec:transfer}
Now, the agent is presented with new environments and asked to solve tasks. Formally, each environment is governed by the standard MDP $\langle S, A, P, r,\gamma\rangle$ and the agent's goal is to learn a policy maximizing the sum of extrinsic rewards, i.e., $\pi_\textsub{task}(s,a) = \arg\max_\pi \mathbb{E}_\pi[\sum_t \gamma^t r(s_t,a_t)]$. Note that while during pre-training the policy was learned across all environments (one exploration policy for all environments), at transfer we learn one task-specific policy for each environment.
\\
In this phase, the interest-policy learned earlier drives exploration as tasks and environments are presented to the agent. 
In order not to forget interestingness over time, the exploration policy is added as a fixed bias to the task-specific policy, similarly to what~\citet{hailu1999amount} proposed.
Thanks to the decoupling of the interest-policy (based on the intrinsic reward) from the task-policy (based on the extrinsic reward), the latter can be also learned independently via any RL algorithm. 
\\
In our experiments, we use IMPALA~\citep{espeholt2018impala} to learn both $\pi_\textsub{exp}$ and $\pi_\textsub{task}$. IMPALA learns policies of the form $\pi(s,a) = \sigma(f(s,a))$, where $\sigma$ is the softmax function. The policy is trained to maximize a function representing the value of states $V(s)$, trained on the given rewards. In our framework, we combine the two policies as follows.
\begin{itemize}[nosep,leftmargin=*,before=\vspace{-4pt}]
    \item During pre-training, by using intrinsic rewards we learn $V_i(s)$ and $\pi_\textsub{exp}(s,a) = \sigma(f_{i}(s,a))$.
    \item At transfer, we learn $V_e(s)$ on extrinsic rewards. The policy is $\pi_\textsub{task}(s,a) = \sigma(f_{e}(s,a) + f_{i}(s,a))$. The interestingness $f_i$ is transferred but not trained, i.e., it acts as fixed bias to encourage interaction.
    Initially the policy follows $f_{i}$ since $f_{e}$ is initialized randomly. As it finds extrinsic rewards, the sum $f_{e} \!+\! f_{i}$ becomes greedier with respect to extrinsic rewards, and $f_{e}$ slowly overtakes $f_{i}$\footnote{If exploration and the task goals are misaligned, we can decay exploration, e.g., $\pi_\textsub{task}(s,a) = \sigma(\alpha f_{i}(s,a) + f_{e}(s,a))$, where $\alpha$ decays over time, similarly to common $\epsilon$-greedy policies.}.
\end{itemize}
Note that we transfer only $f_i$ (the policy) and not $V_i$ (the state value). We could think of transferring $V_i$ as fixed bias as well, i.e., by having $V_e(s) = V(s) + V_i(s)$. The policy would be trained on $V_e$ --the states value with respect to the given task-- where $V_i$ is fixed and only $V$ is updated. 
However, we believe it is more beneficial to isolate the exploratory component within the policy, in order to keep the task-specific value function targeted on extrinsic rewards. 
By not transferring $V_i$, $V_e$ can be accurately trained on extrinsic rewards --that the agent will see often thanks to $f_i$ from the pre-trained policy. $V_e$, in turn, can make $\pi_\textsub{task}$ greedy with respect to extrinsic rewards as $V_e$ is learned.

\begin{figure}[t]
\begin{minipage}[c]{0.51\textwidth}
\vspace*{-1em}
\includegraphics[width=\linewidth]{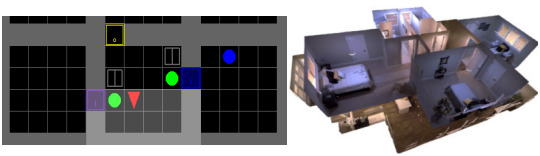}
\end{minipage}
\hfill
\begin{minipage}[c]{0.47\textwidth}
\caption{\label{fig:environments}\textbf{Examples of the environments used in our experiments.} In MiniGrid (left), the agent navigates through a grid and interacts with objects (keys, doors, boxes, and balls) to fulfill a task. In Habitat (right), the agent navigates through visually realistic rooms.}
\end{minipage}
\vspace*{-1.5em}
\end{figure}

\section{Experiments}
\label{sec:eval}
Our experiments are designed to highlight the benefits of disentangling the environment-centric nature of exploration from agent-centric behavior by learning a separate exploration policy and then transferring it to new environments. We stress that for learning task-agnostic exploration there are no standard benchmark environments, experimental setups, well-defined evaluation metrics, or even baselines to compare against. One of our contributions is to provide an exhaustive evaluation framework for the transfer exploration paradigm.
\\[0.3em]
\textbf{Environments.} The experiments are divided into two main sections.
The first is about MiniGrid~\citep{gym_minigrid} (Section~\ref{sec:minigrid}), a set of procedurally-generated environments where the agent can interact with many objects. The second is about Habitat~\citep{habitat19iccv} (Section~\ref{sec:habitat}), a navigation simulator showcasing the generality of our MiniGrid experiments to a visually realistic domain.
\\[0.3em]
\textbf{Change encoding.} In both MiniGrid and Habitat the agent partially observes the environment, since it cannot see through corners, closed door, or inside boxes, and has a limited field of view. Rather than egocentric views (i.e., what the agent sees in front of itself), we use $360^{\circ}$ panoramic views to count environment changes, as this is a rotation-invariant representation of the observed state. Similar to \citet{chaplot2020neural}, we concatenate four egocentric views taken from $0^{\circ}$, $90^{\circ}$, $180^{\circ}$, and $270^{\circ}$ with respect to the North. 
Then, the change of a transition is the difference between panoramic views $\mathsf{pano}(s)$, i.e., $c(s,s') := \mathsf{pano}(s') - \mathsf{pano}(s)$.
\\[0.3em]
\textbf{Baselines.} We evaluate against the following algorithms. For more details, refer to Appendix~\ref{supp:rewards}.
\begin{itemize}[nosep,leftmargin=*,before=\vspace{-4pt}]
\item \textit{Count}~\citep{bellemare2016unifying}. The intrinsic reward is inversely proportional to the next state visitation count. 
\item \textit{Random Network Distillation (RND)}~\citep{burda2018exploration}. The intrinsic reward is the prediction error of states' random features between a trained network and a fixed one. This can be interpreted as similar to using state counts because the prediction improves states are seen more often.
\item \textit{Rewarding Impact-Driven Exploration (RIDE)} \citep{raileanu2020ride}. The intrinsic reward is the prediction error between consecutive embedded states, normalized by episodic state counts.
\item \textit{Curiosity}~\citep{pathak2017curiosity}. The intrinsic reward is the prediction error between consecutive states. 
\end{itemize}
In Appendix~\ref{supp:ablation} we investigate the importance of count resets, panoramic changes, and different count-based rewards.
The source code is available at \url{https://github.com/sparisi/cbet/}.


\subsection{MiniGrid Experiments}
\label{sec:minigrid}
MiniGrid environments \citep{gym_minigrid} are procedurally-generated gridworlds where the agent can interact with objects, such as keys, doors, and boxes (Figure~\ref{fig:environments}). 
Exploration is challenging because rewards are sparse, observations are partial, and specific actions are needed to visit all states (e.g., pickup key to open door).
With MiniGrid, we can generate several pairs of train and test environments that are related but still different in many ways. These pairs enable evaluation of both the learning and transfer abilities of an exploration method and its ability to deal with unseen components.
\\[0.3em]
\textbf{Implementation details.} All environments give a 7$\times$7$\times$3 partial observation encoding the content of the 7$\times$7 tiles in front of the agent (including the agent's tile). The agent cannot see through walls, closed doors, or inside boxes. 
The action space is discrete: left, right, forward, pick up, drop, toggle, and done. 
For a complete description of the environments, we refer to Appendix~\ref{supp:minigrid}. 
\\[0.3em]
\textbf{Setups.} We present three setups, to study different exploration transfers against tabula-rasa.
\begin{itemize}[nosep,leftmargin=*,before=\vspace{-4pt}]
\item \textit{MultiEnv (many-to-many transfer).} The agent loops over three environments episode by episode, and learns the exploration policy using intrinsic rewards only. There is one state count and one change count for all three environments rather than separate counts for each. The environments are: KeyCorridorS3R3, BlockedUnlockPickup, and MultiRoom-N4-S5, and have been chosen for size and interaction variety: the first has both a locked and an unlocked door, a key, and a ball; the second adds a box; the third has more rooms. Note that even if these environments have all object types, the agent cannot experience all kinds of interactions. For example, it will not know that keys can be hidden in boxes, as in the ObstructedMazes. The policy is then transferred to ten environments, seven of which are new. A good intrinsic reward should help learn better exploration faster from multiple environments, thanks to sharing experience from diverse interaction.
\item \textit{SingleEnv (one-to-many transfer).} The policy is pre-trained on a single environment. DoorKey and KeyCorridor are used for pre-training because they have some --but not all-- objects.
\item \textit{Tabula-rasa (no pre-training / transfer).} A task-specific policy is learned as in classic intrinsic motivation by summing intrinsic and extrinsic rewards. While it is a non-realistic setup, it is the most common RL exploration approach, and thus serves as baseline against our transfer framework.
\end{itemize}
\textbf{Evaluation metrics.} Our goal is to learn exploration policies that encourage interaction with the environments
and transfer well to new environments, i.e., that can further be trained to solve extrinsic tasks faster. Therefore, we evaluate policies according to the following criteria.
\begin{itemize}[nosep,leftmargin=*,before=\vspace{-4pt}]
\item Unique interactions over 100 episodes at transfer to new environments, after intrinsic-reward pre-training (no extrinsic-reward training yet). Unique interactions are picks/drops/toggles resulting in new environment changes. For instance, repeatedly picking and dropping the same key in the same cell results in only two interactions.
\item Tasks success rate over 100 episodes at transfer to new environments, after intrinsic-reward pre-training (no extrinsic-reward training yet). The task success rate denotes in how many episodes the exploration policy visits goal states --thus, would have already solved the environment task.
\item Extrinsic return during extrinsic-reward training, after intrinsic-reward training. 
\end{itemize}

\subsubsection{MiniGrid Pre-Training Results} 
\label{subsec:minigrid_res}
Figure~\ref{fig:minigrid_pretrain} shows results after pre-training in MultiEnv. In Appendix~\ref{supp:mini_extra} we report results for the two SingleEnv setups as well.
C-BET policy both interacts with the environment and find goal states more often than all baselines. As we will see in the next section, this will result in faster extrinsic-reward learning.
Furthermore, C-BET's policy transfers well to all environments, even the ones with unknown dynamics (e.g., boxes in ObstructedMazes needs to be toggled to reveal keys). Of the baselines, only Count scores high average interactions and success rate, but it does not generalize as well as C-BET. Indeed, most of Count's success comes from environments visited at pre-training (the first five), but most of its interactions are in environments with unseen dynamics (ObstructedMazes). 
That is, Count's policy can explore familiar environments prioritizing state coverage (high success rate and few interactions), but not unfamiliar ones (low success rate yet high interactions).

\begin{figure*}[h]
\begin{center}
\includegraphics[width=\linewidth]{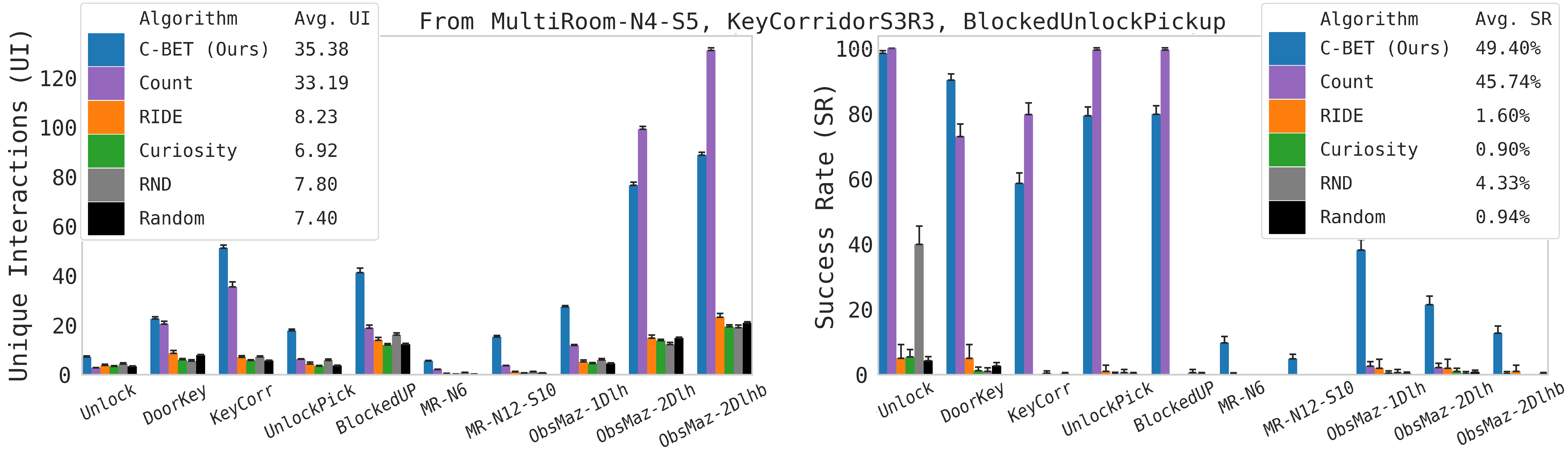}
\vspace{-15pt}
\caption{\label{fig:minigrid_pretrain}\textbf{Unique interactions and success rate} at the beginning of transfer of policies pre-trained in MultiEnv. 
Not only C-BET interacts the most and achieves the highest success rate, but also interacts and succeeds in \textbf{all} environments. 
Naturally, it interacts more in environment with many keys/balls/boxes to pick (KeyCorridor, BlockedUnblockPickup, ObstructedMazes), and less if there is nothing to pick (MultiRooms).
On the contrary, Count overfits to the training environments and performs well only on the first five. Other baselines perform poorly, almost as a random policy.}
\end{center}
\end{figure*}

\clearpage

Finally, RIDE, Curiosity, and RND baselines perform poorly.
This is unsurprising if we consider that they rely on predictive models and that MiniGrid dynamics are deterministic and simple. Dynamics and embeddings models are learned quickly, without giving the policy time to explore. 
In Appendix~\ref{supp:mini_decay} we support this claim by showing the intrinsic reward trend during pre-training.

\subsubsection{MiniGrid Transfer Results}

\begin{figure*}[t]
\vspace{-1.5em}
\begin{center}
\includegraphics[width=0.99\linewidth]{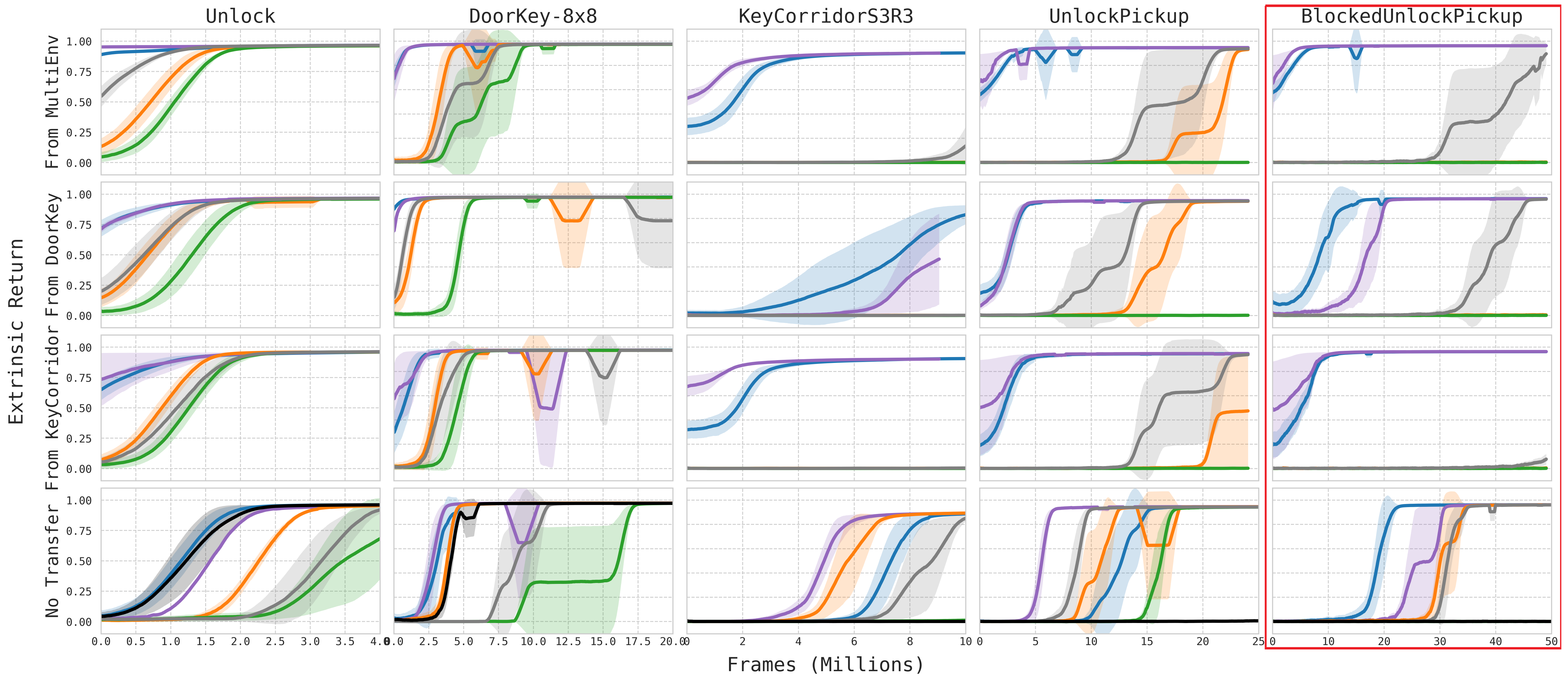}
\includegraphics[width=0.99\linewidth]{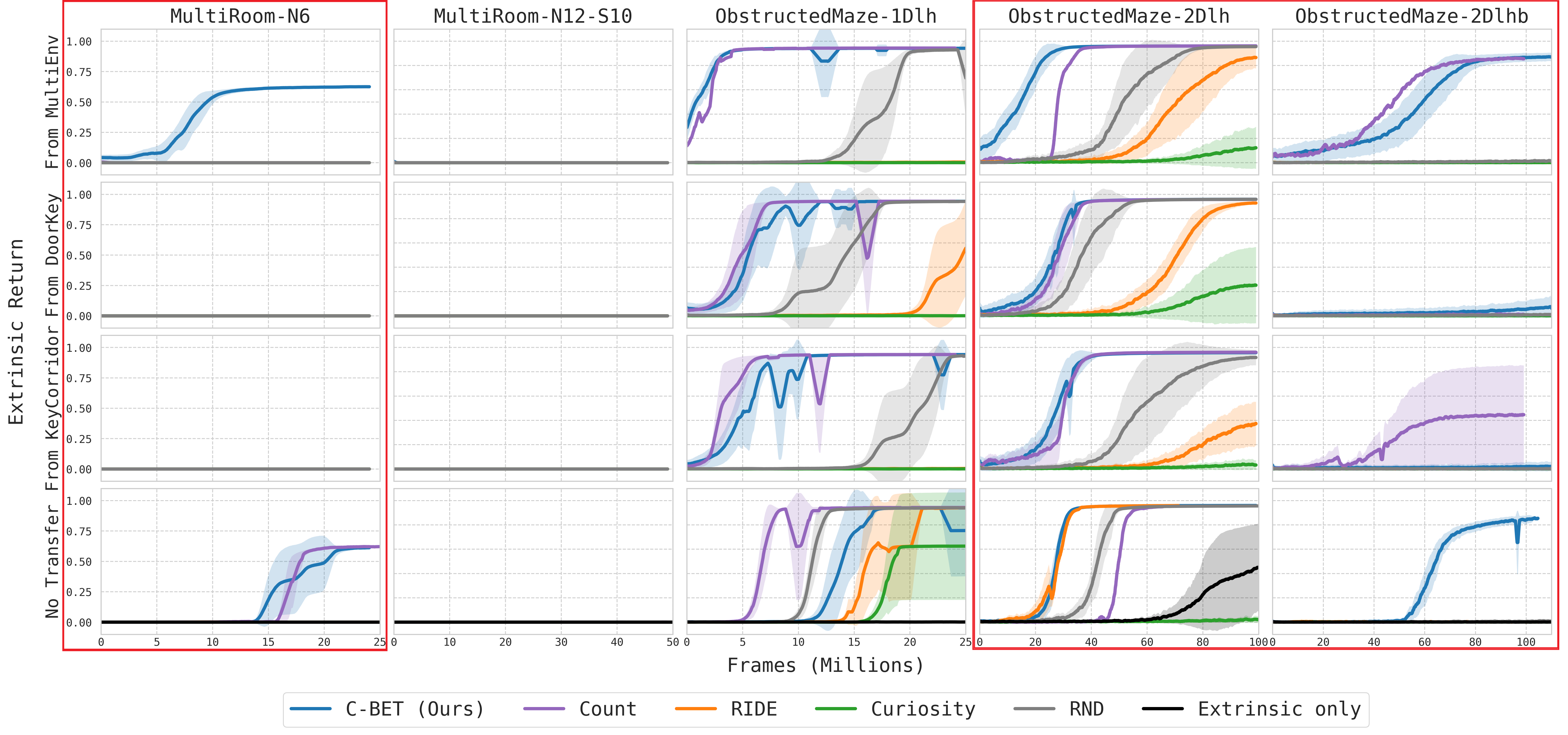}
\vspace{-1em}
\caption{\label{fig:results_minigrid}\textbf{MiniGrid task learning, for both transfer and tabula-rasa exploration.} The hardest tasks are outlined in red. 
C-BET (blue) from MultiEnv (top row under each environment) performs the best, starting with nearly optimal policies in most environments. This demonstrates the effectiveness of pre-training on multiple environments using the C-BET intrinsic reward.}
\vspace{-1.em}
\end{center}
\end{figure*}
We transfer the exploration policies learned in Figure~\ref{fig:minigrid_pretrain} as discussed in Section~\ref{subsec:transfer}. Figure~\ref{fig:results_minigrid} shows how transfer setups (many-to-many and one-to-many) perform against tabula-rasa exploration. 
\\
The first takeaway is that policies pre-trained with the C-BET intrinsic reward outperform baselines in both transfer and tabula-rasa. In MultiEnv transfer, C-BET performs the best, especially on the hardest environments (outlined in red). In particular, only C-BET is able to transfer to MultiRoom-N6. On the contary, Count --that can solve it in tabula-rasa-- fails at transfer.
C-BET is also the only solving ObstructedMaze-2Dlhb --the hardest environment among the ten-- even in tabula-rasa. 
\\
The second takeaway is that baselines relying on models are not suited to the transfer framework. RIDE, Curiosity and RND perform better in the tabula-rasa setup (last row), except for the easiest environments (Unlock and DoorKey), meaning that transfer is actually harmful. These results are in line with Figure~\ref{fig:minigrid_pretrain}, where only C-BET and Count show success at offline transfer. 
Furthermore, RIDE, Curiosity and RND perform worst when transfer is from MultiEnv, highlighting that their intrinsic rewards are not suited for a multi-environment setup.
\\
Finally, no algorithm learns MultiRoom-N12-10, not even C-BET despite showing some success in Figure~\ref{fig:minigrid_pretrain}. This is due to the randomly-initialized $f_e$ of the task-specific policy, hindering the pre-trained exploration policy success. 

\clearpage

\begin{figure*}[h]
\centering
\begin{minipage}[t]{.38\linewidth}
    \centering
    \includegraphics[width=\linewidth]{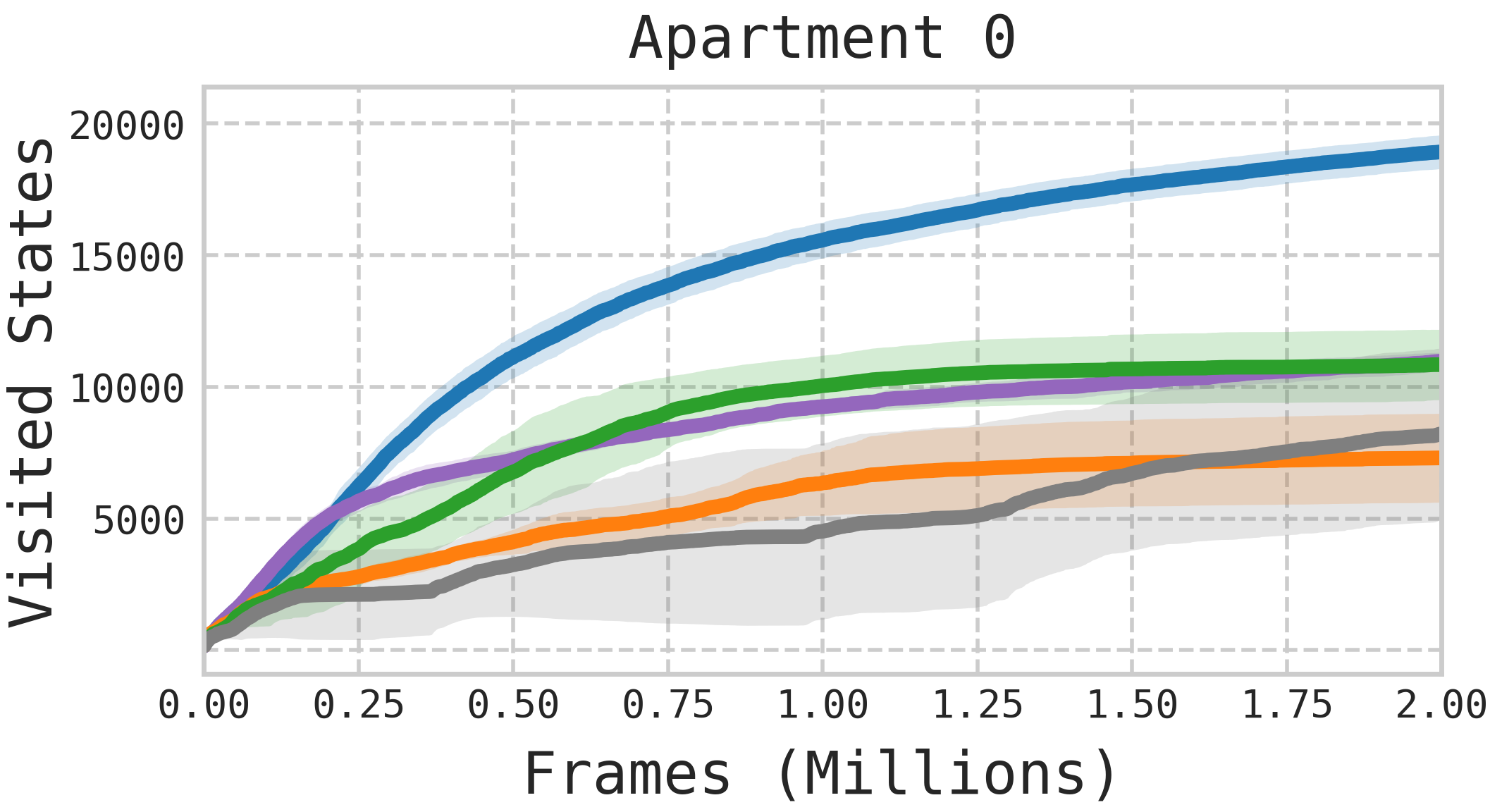}
    \vspace{-1.5em}
    \caption{\label{fig:habitat_pretrain}\textbf{Habitat pre-training.} C-BET explores the scene faster and scores the highest unique state count.}
\end{minipage}\hfill
\begin{minipage}[t]{.6\textwidth}
    \centering
    \includegraphics[width=\linewidth]{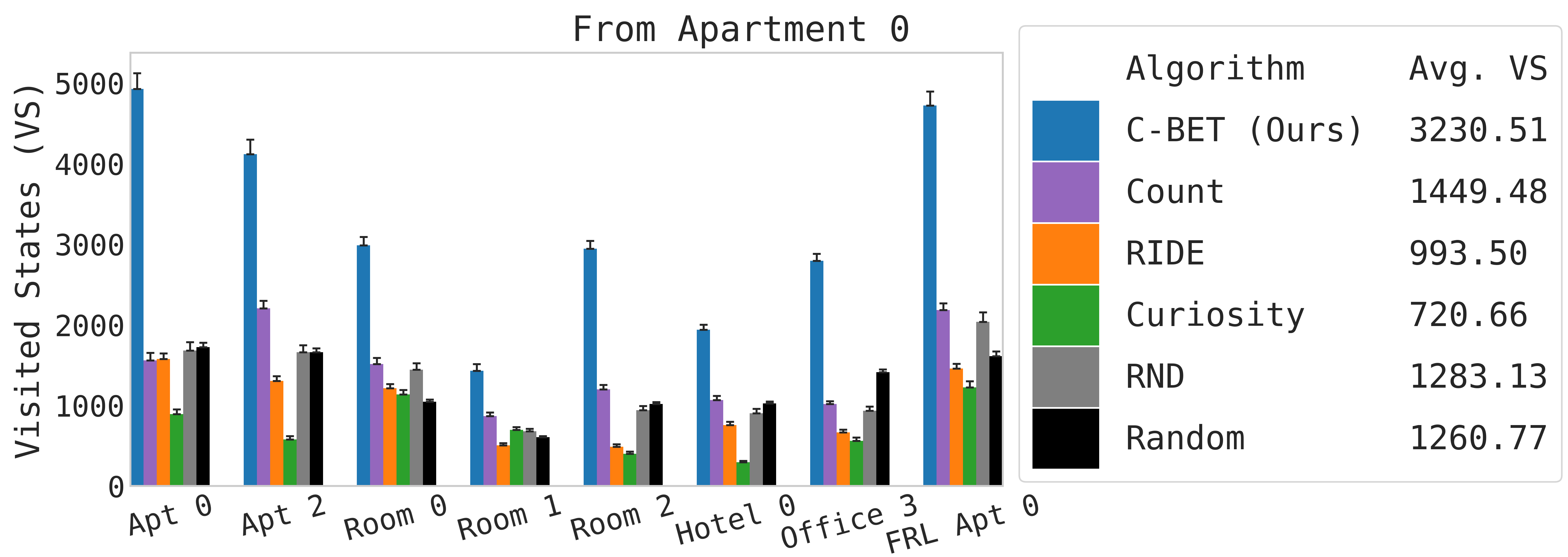}
    \vspace{-1.5em}
    \caption{\label{fig:habitat_transfer}\textbf{Habitat offline transfer.} Bars denote the unique state count in an new scene during one episode.
    C-BET visits more than twice as many states than all baselines.}
\end{minipage}
\vspace*{-1em}
\end{figure*}
\begin{figure*}[h]
\centering
\textbf{\small \hspace*{2.7em} C-BET \hfill RIDE \hfill Count \hfill Curiosity \hfill RND \hspace*{2.7em}}
\\[0.em]
\includegraphics[width=0.19\linewidth]{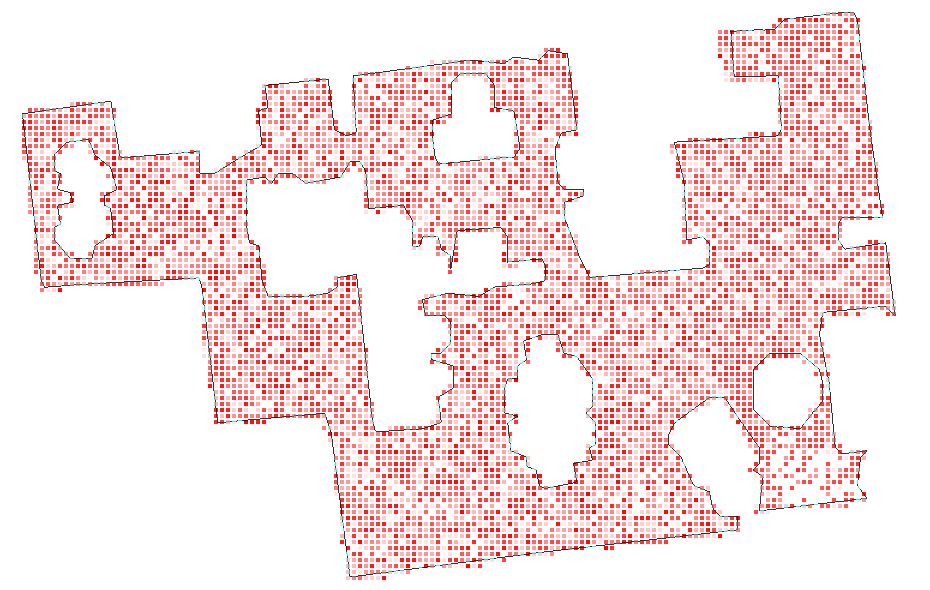}
\hfill
\includegraphics[width=0.19\linewidth]{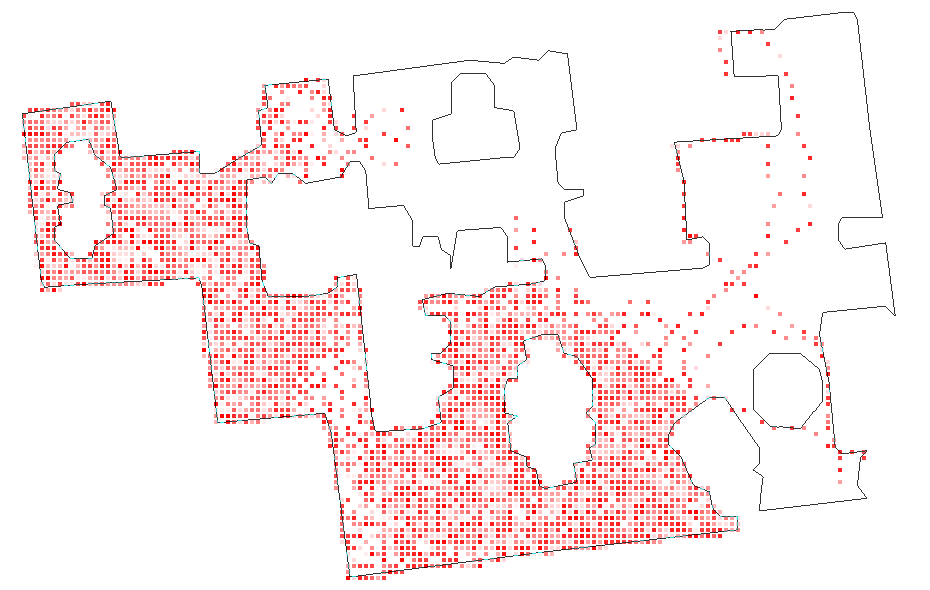}
\hfill
\includegraphics[width=0.19\linewidth]{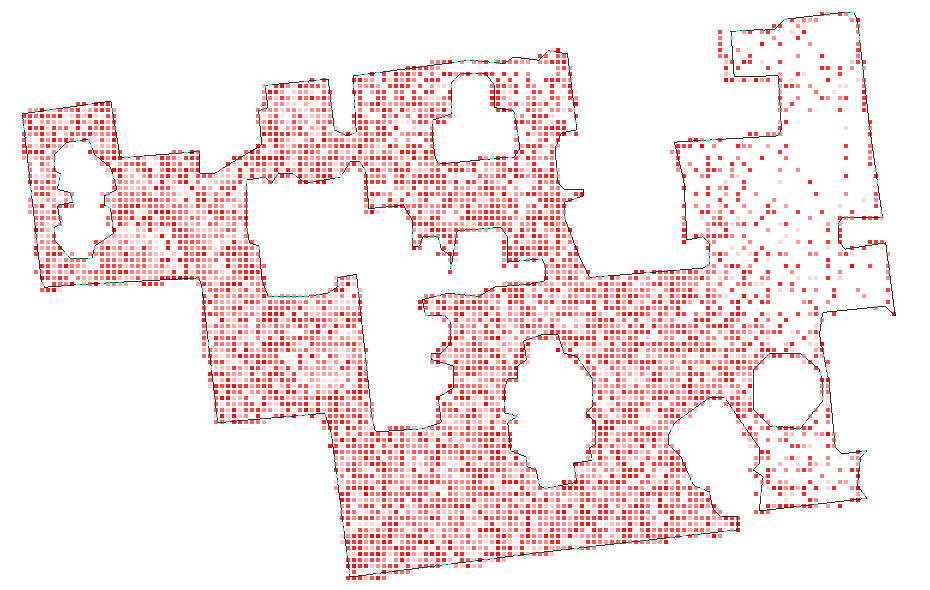}
\hfill
\includegraphics[width=0.19\linewidth]{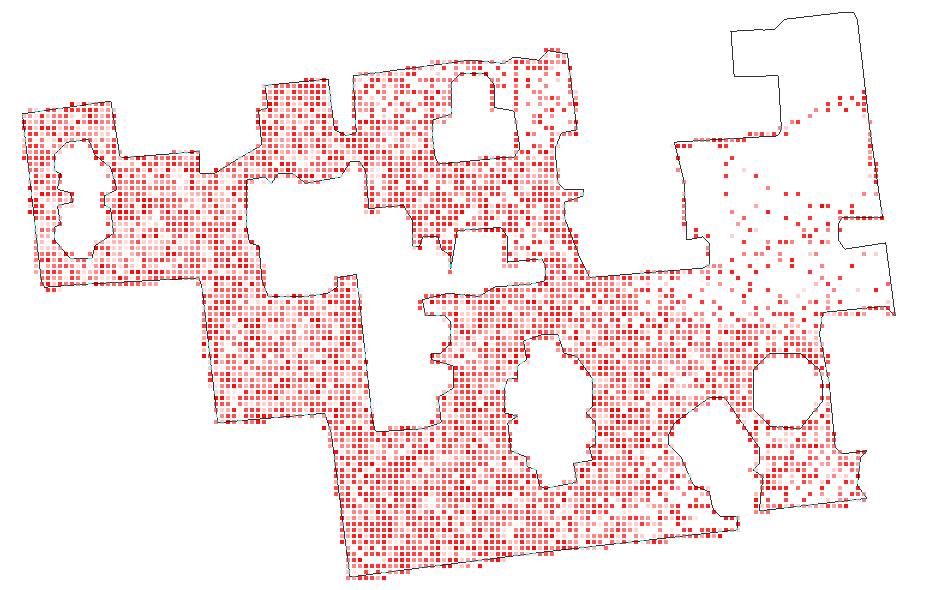}
\hfill
\includegraphics[width=0.19\linewidth]{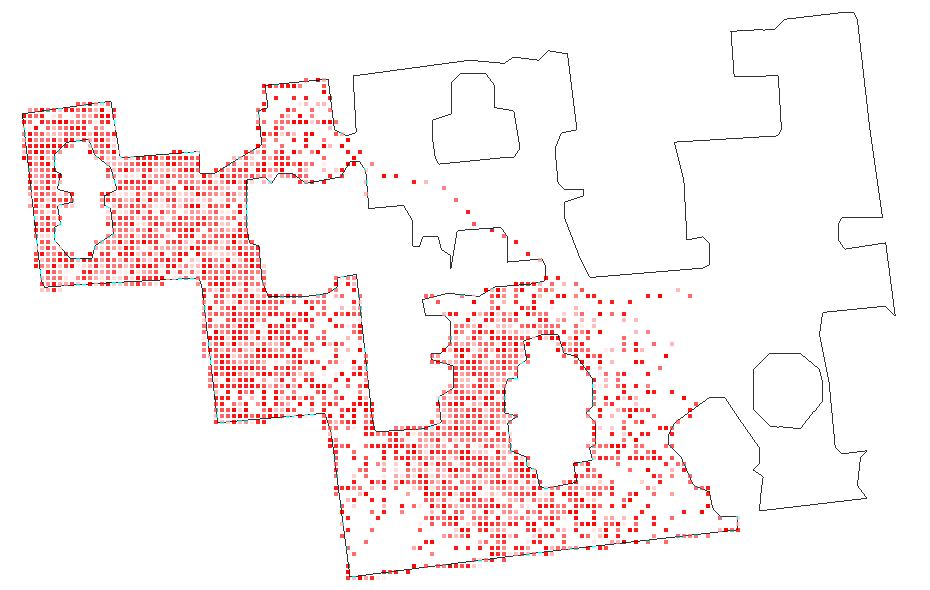}
\caption{\label{fig:habitat_heatmap_pre}
\textbf{Scene coverage} of exploration policies during pre-training (2M steps) in Apartment 0. Darker red cells denote higher visitation rates. Only C-BET visits all of the scene uniformly.}
\end{figure*}

\subsection{Habitat Experiments}
\label{sec:habitat}
To demonstrate that C-BET's efficacy extends to realistic settings with visual inputs, we perform experiments on Habitat~\citep{habitat19iccv} with Replica scenes \citep{replica19arxiv}.
\\[0.3em]
\textbf{Implementation details.}
Egocentric views have resolution 64$\times$64$\times$3. The action space is discrete: forward 0.25 meter, turn $10^{\circ}$ left, and turn $10^{\circ}$ right.
To ease computational demands, we use \#Exploration~\citep{tang2017exploration} with static hashing to map both egocentric and panoramic views to hash codes and count their occurrences with a hash table. More details in Appendix~\ref{supp:habitat}.
\\[0.3em]
\textbf{Setups.} We evaluate Habitat on the \textit{one-to-many transfer}. First, we pre-train exploration policies with only intrinsic rewards in one scene. Then, we evaluate them on new scenes without further learning. Given a fixed amount of steps, better policies will visit more of the new scenes.
\\[0.3em]
\textbf{Evaluation metrics.}
Unlike MiniGrid, we use no extrinsic rewards in Habitat.
Since the agent has to navigate through rooms and spaces, we evaluate exploration policies using scene coverage measured by the agent's true state in Cartesian coordinates (not accessible by the agent)\footnote{To ease memory usage, we round states to $0.05$ precision, e.g., $1.26$ is rounded to $1.25$, and $1.28$ to $1.30$.}. 
Faster, larger and more uniform coverage corresponds to better exploration. Plots show mean and confidence interval over seven random seeds per method with no smoothing.

\subsubsection{Habitat Pre-Training Results}
We pre-train exploration policies on Apartment 0 (Figure~\ref{fig:environments}), one of the largest Replica scene in the dataset.
Figures~\ref{fig:habitat_pretrain} and~\ref{fig:habitat_heatmap_pre} show state coverage throughout and at the end of pre-training, respectively. C-BET explores more efficiently, covering twice as much of the scene than all baselines. In particular, at the end of pre-training it has explored almost all Apartment 0 uniformly. In Appendix~\ref{supp:hab_cbet_ego} we also report C-BET results when environment changes are encoded with egocentric views rather than panoramic views.

\subsubsection{Habitat Transfer Results}
Here, we evaluate scene coverage of pre-trained policies in seven unseen scenes for episodes of fixed steps.
A better exploration policy will exhibit generalization by covering a larger portion of all scenes as evenly as possible, an impressive feat given the visual complexity of the observations. Indeed, generalization is harder than MiniGrid because the lighting, colors, objects, and layout can be very different between scenes (see Figure~\ref{fig:replica_scenes} in the Appendix).  
Figures~\ref{fig:habitat_transfer} and~\ref{fig:habitat_heatmap} show that, once again, C-BET clearly outperforms all baselines. Its exploration policy transfer well to all scenes, as it uniformly discovers more states. 
No baseline comes closer to its results. Actually, in many scenes baselines perform worse than a random policy.

\clearpage

\begin{figure*}[h]
\centering
\textbf{\small \hspace*{2.9em} C-BET \hfill RIDE \hfill Count \hfill Curiosity \hfill RND \hfill Random \hspace*{1.5em}}
\\[0.em]
\raisebox{24pt}{\rotatebox[origin=t]{90}{\textbf{Room 0}}}
\includegraphics[width=0.155\linewidth,height=2cm]{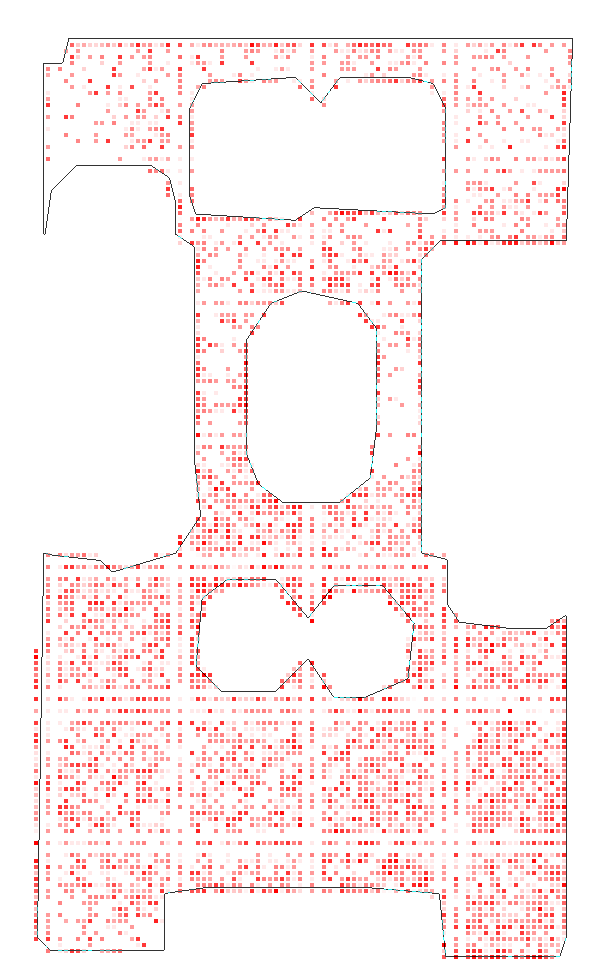}
\hfill
\includegraphics[width=0.155\linewidth,height=2cm]{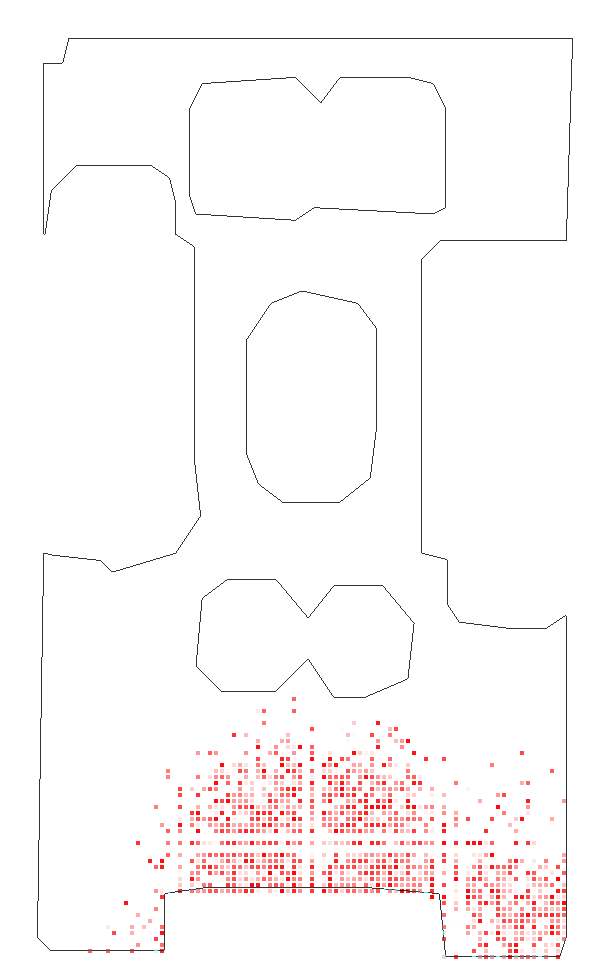}
\hfill
\includegraphics[width=0.155\linewidth,height=2cm]{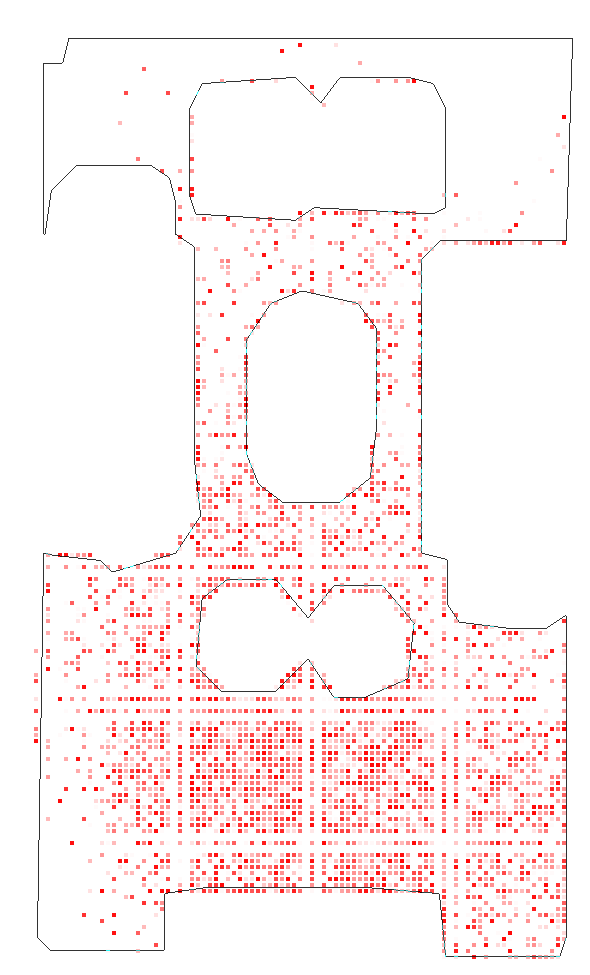}
\hfill
\includegraphics[width=0.155\linewidth,height=2cm]{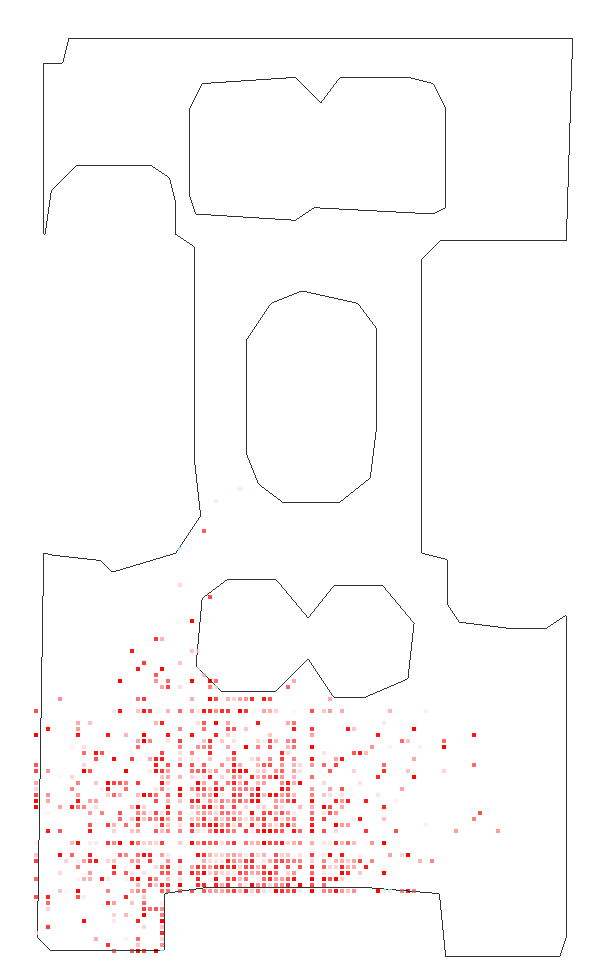}
\hfill
\includegraphics[width=0.155\linewidth,height=2cm]{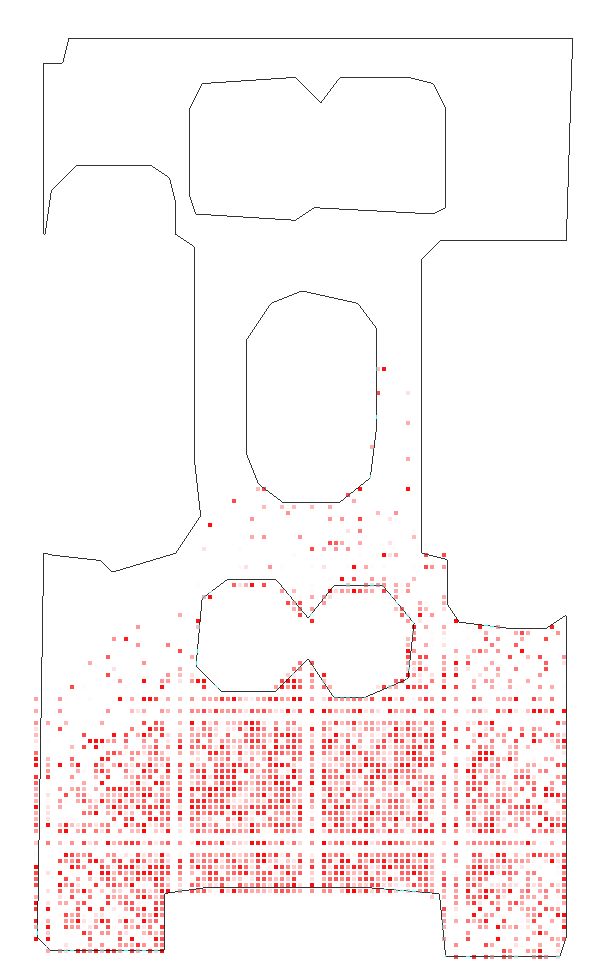}
\hfill
\includegraphics[width=0.155\linewidth,height=2cm]{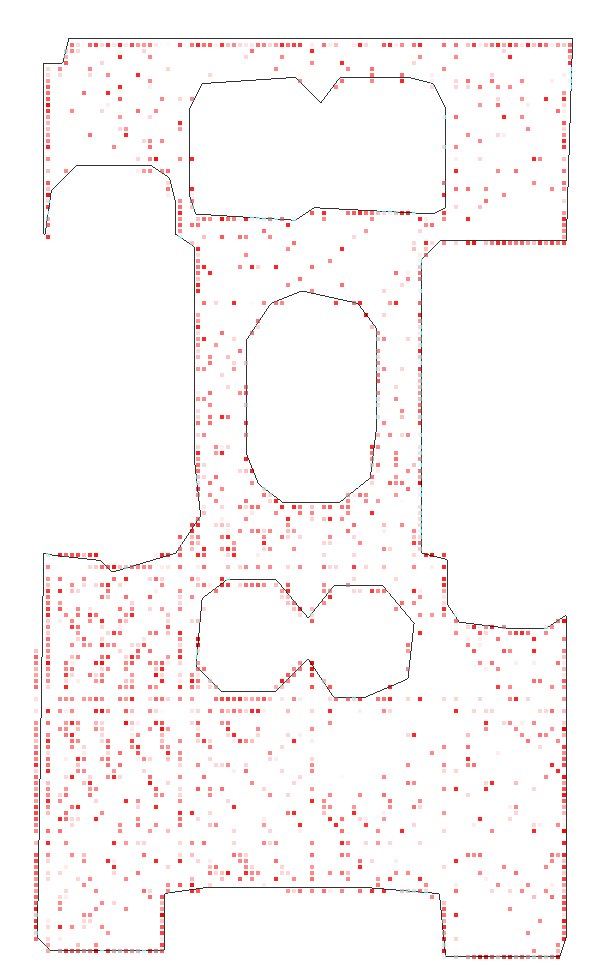}
\vspace*{-0.8em}
\caption{\label{fig:habitat_heatmap}
\textbf{Scene coverage} of exploration policies after 100 episodes (50,000 total steps) at offline transfer to Room 0. C-BET outperforms baselines and exhibits great transfer by visiting all of the scene uniformly. In Appendix~\ref{supp:scenes}, we show heatmaps for all transfer scenes.}
\end{figure*}

\section{Discussion}
\label{sec:outro}
In this paper, we proposed a paradigm change in task-agnostic exploration. Instead of studying task-agnostic exploration in isolated environments, we proposed to (1) learn task-agnostic exploration policies from one or multiple environments, and (2) transfer learned exploration policies to unseen environments at testing time. In our setup, the agent interacts with the environment without any extrinsic goal and learns to explore environments in a task-agnostic manner. To this end, we proposed a novel intrinsic reward to encourage interaction with the environment and the visitation of unseen states. Subsequently, our agent effectively transfers its exploration policy to unseen environments.
\\[0.2em]
\textbf{Advantages.}
The proposed two-phase framework achieves two important features, making it fundamentally different from prior work. First, we account for \textit{environment interestingness} without relying on additional models. 
Instead, we use a data-driven approach, estimating the rarity of states and environment changes. Rare changes are considered more interesting, actions causing them receive higher intrinsic rewards, and the agent is encouraged to perform them again.
For instance, when navigating through rooms, opening doors will be more interesting due to rarity: the agent must navigate to the corresponding key, collect it, navigate to the door, and finally open it. Thus opening a door is rarer than picking up a key, in turn rarer than simple navigation movements.
Furthermore, relying on environment-centric intrinsic rewards rather than task-centric extrinsic rewards facilitates learning from multiple environments at the same time.
\\[0.2em]
Second, contrary to prior transfer and continual learning algorithms we transfer policies learned on \textit{interestingness of the environment} rather than task-specific policies.
In the interest-based pre-training phase, we learn through interaction with the environment in a task-agnostic fashion, i.e., the agent freely explores the environment without any extrinsic task.
\\[0.2em]
\textbf{Limitations.}
\label{subsec:limitations}
In this paper, we assumed that interacting with the environment while looking for rare changes helps find better extrinsic rewards faster. However, exploration and the task goals may be misaligned, thus a highly exploratory policy may slow down the discovery of extrinsic rewards. For instance, the environment may have dangerous states or harmful objects that the agent should avoid, even though they would make it curious during pre-training.
Furthermore, C-BET is currently tied to (pseudo)counts to compute the rarity of states and changes. While extensions to continuous spaces exist, count-based metrics are more suited for discrete spaces. 
\\[0.2em]
\textbf{Impact.}
\label{subsec:impact}
RL can positively impact real-world problems, e.g., healthcare \citep{gottesman2019guidelines}, assistive robotics \citep{erickson2020assistive}, and climate change \citep{rolnick2019tackling}. Yet, RL may have negative impacts, e.g., in autonomous weapons or workforce displacement~\citep{brynjolfsson2017can}. 
Our work focuses on exploration in RL. 
Better understanding of what is interesting to do or visit helps exploration in unseen environments, as the agent will not waste time with random actions. Similarly, transferring policies learned in a related setting --as we do-- can help narrow the range of the agent's expected behavior.
Conversely, in many real-world scenarios exploration by curiosity and interestingness is unacceptable. For instance, autonomous cars cannot run over pedestrians just for the sake of curiosity.
At present, our work is far from these impacts, but we hope to direct research to focus more on learning from multiple environments and transferring experiences,
while at the same time ensuring the safety and reliability of autonomous agents.


%

\begin{ack}
The authors would like to thank Davide Tateo for his thoughtful discussions, Roberta Raileanu for providing support for RIDE's codebase, and Sudeep Dasari for helping run experiments. VD and DP were supported in part by NSF Fellowship and DARPA Machine Common Sense grant, respectively.
\end{ack}

\clearpage

\bibliography{rl_bib, cl_bib}

\begin{thebibliography}{59}
\providecommand{\natexlab}[1]{#1}
\providecommand{\url}[1]{\texttt{#1}}
\expandafter\ifx\csname urlstyle\endcsname\relax
  \providecommand{\doi}[1]{doi: #1}\else
  \providecommand{\doi}{doi: \begingroup \urlstyle{rm}\Url}\fi

\bibitem[Auer and Ortner(2007)]{auer2007logarithmic}
P.~Auer and R.~Ortner.
\newblock Logarithmic online regret bounds for undiscounted reinforcement
  learning.
\newblock In \emph{Advances in Neural Information Processing Systems (NIPS)},
  pages 49--56, 2007.

\bibitem[Auer et~al.(2002)Auer, {Cesa-Bianchi}, and Fischer]{auer2002finite}
P.~Auer, N.~{Cesa-Bianchi}, and P.~Fischer.
\newblock {Finite-time} analysis of the multiarmed bandit problem.
\newblock \emph{Machine Learning}, 47\penalty0 (2-3):\penalty0 235--256, 2002.

\bibitem[Barreto et~al.(2017)Barreto, Dabney, Munos, Hunt, Schaul, Van~Hasselt,
  and Silver]{barreto2016successor}
A.~Barreto, W.~Dabney, R.~Munos, J.~J. Hunt, T.~Schaul, H.~Van~Hasselt, and
  D.~Silver.
\newblock Successor features for transfer in reinforcement learning.
\newblock In \emph{International Conference on Neural Information Processing
  Systems (NeurIPS)}, 2017.

\bibitem[Bellemare et~al.(2016)Bellemare, Srinivasan, Ostrovski, Schaul,
  Saxton, and Munos]{bellemare2016unifying}
M.~G. Bellemare, S.~Srinivasan, G.~Ostrovski, T.~Schaul, D.~Saxton, and
  R.~Munos.
\newblock Unifying {count-based} exploration and intrinsic motivation.
\newblock In \emph{Advances in Neural Information Processing Systems (NIPS)},
  2016.

\bibitem[Brafman and Tennenholtz(2002)]{brafman2002r}
R.~I. Brafman and M.~Tennenholtz.
\newblock {{R-MAX} - A} general polynomial time algorithm for {near-optimal}
  reinforcement learning.
\newblock \emph{Journal of Machine Learning Research (JMLR)}, 3\penalty0
  (Oct):\penalty0 213--231, 2002.

\bibitem[Brynjolfsson and Mitchell(2017)]{brynjolfsson2017can}
E.~Brynjolfsson and T.~Mitchell.
\newblock What can machine learning do? workforce implications.
\newblock \emph{Science}, 358\penalty0 (6370), 2017.

\bibitem[Burda et~al.(2019)Burda, Edwards, Storkey, and
  Klimov]{burda2018exploration}
Y.~Burda, H.~Edwards, A.~Storkey, and O.~Klimov.
\newblock Exploration by random network distillation.
\newblock In \emph{International Conference on Learning Representations
  (ICLR)}, 2019.

\bibitem[Chaplot et~al.(2020)Chaplot, Salakhutdinov, Gupta, and
  Gupta]{chaplot2020neural}
D.~S. Chaplot, R.~Salakhutdinov, A.~Gupta, and S.~Gupta.
\newblock Neural topological slam for visual navigation.
\newblock In \emph{Proceedings of the IEEE/CVF Conference on Computer Vision
  and Pattern Recognition}, pages 12875--12884, 2020.

\bibitem[Charikar(2002)]{charikar2002similarity}
M.~S. Charikar.
\newblock Similarity estimation techniques from rounding algorithms.
\newblock In \emph{Symposium on Theory of Computing}, 2002.

\bibitem[{Chevalier-Boisvert} et~al.(2018){Chevalier-Boisvert}, Willems, and
  Pal]{gym_minigrid}
M.~{Chevalier-Boisvert}, L.~Willems, and S.~Pal.
\newblock {Minimalistic Gridworld Environment for OpenAI Gym}, 2018.
\newblock URL \url{https://github.com/maximecb/gym-minigrid}.

\bibitem[Clevert et~al.(2015)Clevert, Unterthiner, and
  Hochreiter]{clevert2015fast}
D.-A. Clevert, T.~Unterthiner, and S.~Hochreiter.
\newblock Fast and accurate deep network learning by exponential linear units
  ({ELU}s), 2015.

\bibitem[{D'Eramo} et~al.(2019){D'Eramo}, {Cini}, and
  {Restelli}]{deramo2019exploiting}
C.~{D'Eramo}, A.~{Cini}, and M.~{Restelli}.
\newblock Exploiting {Action-Value} uncertainty to drive exploration in
  reinforcement learning.
\newblock In \emph{International Joint Conference on Neural Networks (IJCNN)},
  2019.

\bibitem[Dong et~al.(2020)Dong, Wang, Chen, and Wang]{dong2020q}
K.~Dong, Y.~Wang, X.~Chen, and L.~Wang.
\newblock {Q-learning} with {UCB} exploration is sample efficient for
  {Infinite-Horizon} {MDP}.
\newblock In \emph{International Conference on Learning Representation (ICLR)},
  2020.

\bibitem[Dubey et~al.(2018)Dubey, Agrawal, Pathak, Griffiths, and
  Efros]{dubey2018investigating}
R.~Dubey, P.~Agrawal, D.~Pathak, T.~L. Griffiths, and A.~A. Efros.
\newblock Investigating human priors for playing video games.
\newblock In \emph{International Conference on Machine Learning (ICML)}, 2018.

\bibitem[Erickson et~al.(2020)Erickson, Gangaram, Kapusta, Liu, and
  Kemp]{erickson2020assistive}
Z.~Erickson, V.~Gangaram, A.~Kapusta, C.~K. Liu, and C.~C. Kemp.
\newblock Assistive gym: A physics simulation framework for assistive robotics.
\newblock In \emph{2020 IEEE International Conference on Robotics and
  Automation (ICRA)}, pages 10169--10176. IEEE, 2020.

\bibitem[Espeholt et~al.(2018)Espeholt, Soyer, Munos, Simonyan, Mnih, Ward,
  Doron, Firoiu, Harley, Dunning, et~al.]{espeholt2018impala}
L.~Espeholt, H.~Soyer, R.~Munos, K.~Simonyan, V.~Mnih, T.~Ward, Y.~Doron,
  V.~Firoiu, T.~Harley, I.~Dunning, et~al.
\newblock Impala: Scalable distributed deep-rl with importance weighted
  actor-learner architectures.
\newblock In \emph{International Conference on Machine Learning}, pages
  1407--1416. PMLR, 2018.

\bibitem[Fern{\'a}ndez and Veloso(2006)]{fernandez2006probabilistic}
F.~Fern{\'a}ndez and M.~Veloso.
\newblock Probabilistic policy reuse in a reinforcement learning agent.
\newblock In \emph{International Joint Conference on Autonomous Agents and
  Multiagent Systems (AAMAS)}, 2006.

\bibitem[Finn et~al.(2017)Finn, Abbeel, and Levine]{finn2017model}
C.~Finn, P.~Abbeel, and S.~Levine.
\newblock {Model-agnostic} {meta-learning} for fast adaptation of deep
  networks.
\newblock In \emph{International Conference on Machine Learning (ICML)}, 2017.

\bibitem[Gottesman et~al.(2019)Gottesman, Johansson, Komorowski, Faisal,
  Sontag, Doshi-Velez, and Celi]{gottesman2019guidelines}
O.~Gottesman, F.~Johansson, M.~Komorowski, A.~Faisal, D.~Sontag,
  F.~Doshi-Velez, and L.~A. Celi.
\newblock Guidelines for reinforcement learning in healthcare.
\newblock \emph{Nature medicine}, 25\penalty0 (1):\penalty0 16--18, 2019.

\bibitem[Gottlieb et~al.(2013)Gottlieb, Oudeyer, Lopes, and
  Baranes]{gottlieb2013information}
J.~Gottlieb, P.~Oudeyer, M.~Lopes, and A.~Baranes.
\newblock {Information-seeking}, curiosity, and attention: computational and
  neural mechanisms.
\newblock \emph{Trends in Cognitive Sciences}, 17\penalty0 (11):\penalty0
  585--593, 2013.

\bibitem[Hailu and Sommer(1999)]{hailu1999amount}
G.~Hailu and G.~Sommer.
\newblock On amount and quality of bias in reinforcement learning.
\newblock In \emph{International Conference on Systems, Man, and Cybernetics
  (SMC)}, 1999.

\bibitem[Hansen et~al.(2020)Hansen, Dabney, Barreto, Van~de Wiele,
  {Warde-Farley}, and Mnih]{hansen2020fast}
S.~Hansen, W.~Dabney, A.~Barreto, T.~Van~de Wiele, D.~{Warde-Farley}, and
  V.~Mnih.
\newblock Fast task inference with variational intrinsic successor features.
\newblock In \emph{International Conference on Learning Representations
  (ICLR)}, 2020.

\bibitem[Hochreiter and Schmidhuber(1997)]{hochreiter1997long}
S.~Hochreiter and J.~Schmidhuber.
\newblock Long short-term memory.
\newblock \emph{Neural Computation}, 9\penalty0 (8):\penalty0 1735--1780, 1997.

\bibitem[Houthooft et~al.(2016)Houthooft, Chen, Duan, Schulman, De~Turck, and
  Abbeel]{houthooft2016vime}
R.~Houthooft, X.~Chen, Y.~Duan, J.~Schulman, F.~De~Turck, and P.~Abbeel.
\newblock {VIME}: Variational information maximizing exploration.
\newblock In \emph{Advances in Neural Information Processing Systems (NIPS)},
  2016.

\bibitem[Jaksch et~al.(2010)Jaksch, Ortner, and Auer]{jaksch2010near}
T.~Jaksch, R.~Ortner, and P.~Auer.
\newblock {Near-optimal} regret bounds for reinforcement learning.
\newblock \emph{Journal of Machine Learning Research (JMLR)}, 11\penalty0
  (Apr):\penalty0 1563--1600, 2010.

\bibitem[Jin et~al.(2018)Jin, {Allen-Zhu}, Bubeck, and Jordan]{jin2018iq}
C.~Jin, Z.~{Allen-Zhu}, S.~Bubeck, and M.~I. Jordan.
\newblock Is {Q-learning} provably efficient?
\newblock In \emph{Advances in Neural Information Processing Systems (NIPS)},
  2018.

\bibitem[Kearns and Singh(2002)]{kearns2002near}
M.~Kearns and S.~Singh.
\newblock {Near-optimal} reinforcement learning in polynomial time.
\newblock \emph{Machine Learning}, 49\penalty0 (2-3):\penalty0 209--232, 2002.

\bibitem[Kirkpatrick et~al.(2017)Kirkpatrick, Pascanu, Rabinowitz, Veness,
  Desjardins, Rusu, Milan, Quan, Ramalho, {Grabska-Barwinska}, Hassabis,
  Clopath, Kumaran, and Hadsell]{kirkpatrick2017overcoming}
J.~Kirkpatrick, R.~Pascanu, N.~Rabinowitz, J.~Veness, G.~Desjardins, A.~A.
  Rusu, K.~Milan, J.~Quan, T.~Ramalho, A.~{Grabska-Barwinska}, D.~Hassabis,
  C.~Clopath, D.~Kumaran, and R.~Hadsell.
\newblock Overcoming catastrophic forgetting in neural networks.
\newblock \emph{National Academy of Sciences}, 114\penalty0 (13):\penalty0
  3521--3526, 2017.

\bibitem[Klyubin et~al.(2005)Klyubin, Polani, and Nehaniv]{klyubin2005all}
A.~S. Klyubin, D.~Polani, and C.~L. Nehaniv.
\newblock All else being equal be empowered.
\newblock In \emph{European Conference on Artificial Life}, 2005.

\bibitem[K\"{u}ttler et~al.(2019)K\"{u}ttler, Nardelli, Lavril, Selvatici,
  Sivakumar, Rockt\"{a}schel, and Grefenstette]{torchbeast2019}
H.~K\"{u}ttler, N.~Nardelli, T.~Lavril, M.~Selvatici, V.~Sivakumar,
  T.~Rockt\"{a}schel, and E.~Grefenstette.
\newblock {TorchBeast: A PyTorch Platform for Distributed RL}, 2019.
\newblock URL \url{https://github.com/facebookresearch/torchbeast}.

\bibitem[Lai and Robbins(1985)]{lai1985asymptotically}
T.~Lai and H.~Robbins.
\newblock Asymptotically efficient adaptive allocation rules.
\newblock \emph{Adv. Appl. Math.}, 6\penalty0 (1):\penalty0 4--22, Mar 1985.

\bibitem[Narvekar et~al.(2020)Narvekar, Peng, Leonetti, Sinapov, Taylor, and
  Stone]{narvekar2020curriculum}
S.~Narvekar, B.~Peng, M.~Leonetti, J.~Sinapov, M.~E. Taylor, and P.~Stone.
\newblock Curriculum learning for reinforcement learning domains: A framework
  and survey.
\newblock \emph{Journal of Machine Learning Research}, 21\penalty0
  (181):\penalty0 1--50, 2020.

\bibitem[Osband et~al.(2019)Osband, Roy, Russo, and Wen]{osband2019deep}
I.~Osband, B.~V. Roy, D.~J. Russo, and Z.~Wen.
\newblock Deep exploration via randomized value functions.
\newblock \emph{Journal of Machine Learning Research (JMLR)}, 20\penalty0
  (124):\penalty0 1--62, 2019.

\bibitem[Ostrovski et~al.(2017)Ostrovski, Bellemare, van~den Oord, and
  Munos]{ostrovski2017count}
G.~Ostrovski, M.~G. Bellemare, A.~van~den Oord, and R.~Munos.
\newblock {Count-based} exploration with neural density models.
\newblock In \emph{International Conference on Machine Learning (ICML)}, 2017.

\bibitem[Pathak et~al.(2017)Pathak, Agrawal, Efros, and
  Darrell]{pathak2017curiosity}
D.~Pathak, P.~Agrawal, A.~A. Efros, and T.~Darrell.
\newblock {Curiosity-driven} exploration by self-supervised prediction.
\newblock In \emph{International Conference on Machine Learning (ICML)}, 2017.

\bibitem[Raileanu and Rockt{\"{a}}schel(2020)]{raileanu2020ride}
R.~Raileanu and T.~Rockt{\"{a}}schel.
\newblock {RIDE: Rewarding Impact-Driven Exploration for Procedurally-Generated
  Environments}.
\newblock In \emph{International Conference on Learning Representations
  (ICLR)}, 2020.

\bibitem[Rajendran et~al.(2020)Rajendran, Lewis, Veeriah, Lee, and
  Singh]{rajendran2020should}
J.~Rajendran, R.~Lewis, V.~Veeriah, H.~Lee, and S.~Singh.
\newblock How should an agent practice?
\newblock In \emph{Conference on Artificial Intelligence (AAAI)}, 2020.

\bibitem[Rakelly et~al.(2019)Rakelly, Zhou, Finn, Levine, and
  Quillen]{rakelly2019efficient}
K.~Rakelly, A.~Zhou, C.~Finn, S.~Levine, and D.~Quillen.
\newblock Efficient {off-policy} {meta-reinforcement} learning via
  probabilistic context variables.
\newblock In \emph{International Conference on Machine Learning (ICML)}, 2019.

\bibitem[Rezende and Mohamed(2015)]{rezende2015variational}
D.~Rezende and S.~Mohamed.
\newblock Variational inference with normalizing flows.
\newblock In \emph{International Conference on Machine Learning (ICML)}, 2015.

\bibitem[Ring(1994)]{ring1994continual}
M.~B. Ring.
\newblock \emph{Continual learning in reinforcement environments}.
\newblock PhD thesis, University of Texas at Austin Austin, Texas 78712, 1994.

\bibitem[Ring(1998)]{ring1998child}
M.~B. Ring.
\newblock {CHILD}: A first step towards continual learning.
\newblock In \emph{Learning to learn}, pages 261--292. Springer, 1998.

\bibitem[Rolnick et~al.(2019{\natexlab{a}})Rolnick, Ahuja, Schwarz, Lillicrap,
  and Wayne]{rolnick2019experience}
D.~Rolnick, A.~Ahuja, J.~Schwarz, T.~Lillicrap, and G.~Wayne.
\newblock Experience replay for continual learning.
\newblock In \emph{Advances in Neural Information Processing Systems
  (NeurIPS)}, 2019{\natexlab{a}}.

\bibitem[Rolnick et~al.(2019{\natexlab{b}})Rolnick, Donti, Kaack, Kochanski,
  Lacoste, Sankaran, Ross, Milojevic-Dupont, Jaques, Waldman-Brown,
  et~al.]{rolnick2019tackling}
D.~Rolnick, P.~L. Donti, L.~H. Kaack, K.~Kochanski, A.~Lacoste, K.~Sankaran,
  A.~S. Ross, N.~Milojevic-Dupont, N.~Jaques, A.~Waldman-Brown, et~al.
\newblock Tackling climate change with machine learning.
\newblock \emph{arXiv preprint arXiv:1906.05433}, 2019{\natexlab{b}}.

\bibitem[Rusu et~al.(2015)Rusu, Colmenarejo, Gulcehre, Desjardins, Kirkpatrick,
  Pascanu, Mnih, Kavukcuoglu, and Hadsell]{rusu2015policy}
A.~A. Rusu, S.~G. Colmenarejo, C.~Gulcehre, G.~Desjardins, J.~Kirkpatrick,
  R.~Pascanu, V.~Mnih, K.~Kavukcuoglu, and R.~Hadsell.
\newblock Policy distillation.
\newblock In \emph{International Conference on Learning Representations
  (ICLR)}, 2015.

\bibitem[Ryan and Deci(2000)]{ryan2000intrinsic}
R.~M. Ryan and E.~L. Deci.
\newblock Intrinsic and extrinsic motivations: Classic definitions and new
  directions.
\newblock \emph{Contemporary Educational Psychology}, 25\penalty0 (1):\penalty0
  54--67, 2000.

\bibitem[Savva et~al.(2019)Savva, Kadian, Maksymets, Zhao, Wijmans, Jain,
  Straub, Liu, Koltun, Malik, Parikh, and Batra]{habitat19iccv}
M.~Savva, A.~Kadian, O.~Maksymets, Y.~Zhao, E.~Wijmans, B.~Jain, J.~Straub,
  J.~Liu, V.~Koltun, J.~Malik, D.~Parikh, and D.~Batra.
\newblock Habitat: {A} {P}latform for {E}mbodied {AI} {R}esearch.
\newblock In \emph{International Conference on Computer Vision (ICCV)}, 2019.

\bibitem[Schmidhuber(1991)]{schmidhuber1991possibility}
J.~Schmidhuber.
\newblock A possibility for lmplementing curiosity and boredom in
  {Model-Building} neural controllers.
\newblock In \emph{International Conference on Simulation of Adaptive Behavior
  (SAB)}, 1991.

\bibitem[Schmidhuber(2006)]{schmidhuber2006developmental}
J.~Schmidhuber.
\newblock Developmental robotics, optimal artificial curiosity, creativity,
  music, and the fine arts.
\newblock \emph{Connection Science}, 18\penalty0 (2):\penalty0 173--187, 2006.

\bibitem[Schultheis et~al.(2019)Schultheis, Belousov, Abdulsamad, and
  Peters]{schultheis2020receding}
M.~Schultheis, B.~Belousov, H.~Abdulsamad, and J.~Peters.
\newblock Receding horizon curiosity.
\newblock In \emph{Conference on Robot Learning (CoRL)}, 2019.

\bibitem[Schwarz et~al.(2018)Schwarz, Luketina, Czarnecki, {Grabska-Barwinska},
  Teh, Pascanu, and Hadsell]{schwarz2018progress}
J.~Schwarz, J.~Luketina, W.~M. Czarnecki, A.~{Grabska-Barwinska}, Y.~W. Teh,
  R.~Pascanu, and R.~Hadsell.
\newblock Progress \& compress: A scalable framework for continual learning.
\newblock In \emph{International Conference on Machine learning (ICML)}, 2018.

\bibitem[Stadie et~al.(2015)Stadie, Levine, and
  Abbeel]{stadie2015incentivizing}
B.~C. Stadie, S.~Levine, and P.~Abbeel.
\newblock Incentivizing exploration in reinforcement learning with deep
  predictive models.
\newblock In \emph{NIPS Workshop on Deep Reinforcement Learning}, 2015.

\bibitem[Straub et~al.(2019)Straub, Whelan, Ma, Chen, Wijmans, Green, Engel,
  Mur-Artal, Ren, Verma, Clarkson, Yan, Budge, Yan, Pan, Yon, Zou, Leon,
  Carter, Briales, Gillingham, Mueggler, Pesqueira, Savva, Batra, Strasdat,
  Nardi, Goesele, Lovegrove, and Newcombe]{replica19arxiv}
J.~Straub, T.~Whelan, L.~Ma, Y.~Chen, E.~Wijmans, S.~Green, J.~J. Engel,
  R.~Mur-Artal, C.~Ren, S.~Verma, A.~Clarkson, M.~Yan, B.~Budge, Y.~Yan,
  X.~Pan, J.~Yon, Y.~Zou, K.~Leon, N.~Carter, J.~Briales, T.~Gillingham,
  E.~Mueggler, L.~Pesqueira, M.~Savva, D.~Batra, H.~M. Strasdat, R.~D. Nardi,
  M.~Goesele, S.~Lovegrove, and R.~Newcombe.
\newblock The {R}eplica dataset: A digital replica of indoor spaces, 2019.

\bibitem[Strehl and Littman(2008)]{strehl2008analysis}
A.~L. Strehl and M.~L. Littman.
\newblock An analysis of {model-based} interval estimation for {M}arkov
  decision processes.
\newblock \emph{Journal of Computer and System Sciences (JCSS)}, 74\penalty0
  (8):\penalty0 1309--1331, 2008.

\bibitem[Tang et~al.(2017)Tang, Houthooft, Foote, Stooke, Chen, Duan, Schulman,
  DeTurck, and Abbeel]{tang2017exploration}
H.~Tang, R.~Houthooft, D.~Foote, A.~Stooke, O.~X. Chen, Y.~Duan, J.~Schulman,
  F.~DeTurck, and P.~Abbeel.
\newblock {\#Exploration}: A study of {count-based} exploration for deep
  reinforcement learning.
\newblock In \emph{Advances in Neural Information Processing Systems (NIPS)},
  2017.

\bibitem[Teh et~al.(2017)Teh, Bapst, Czarnecki, Quan, Kirkpatrick, Hadsell,
  Heess, and Pascanu]{teh2017distral}
Y.~W. Teh, V.~Bapst, W.~M. Czarnecki, J.~Quan, J.~Kirkpatrick, R.~Hadsell,
  N.~Heess, and R.~Pascanu.
\newblock Distral: Robust multitask reinforcement learning.
\newblock In \emph{International Conference on Neural Information Processing
  Systems (NeurIPS)}, 2017.

\bibitem[Thrun and Mitchell(1995)]{thrun1995lifelong}
S.~Thrun and T.~M. Mitchell.
\newblock Lifelong robot learning.
\newblock \emph{Robotics and Autonomous Systems}, 15\penalty0 (1-2):\penalty0
  25--46, 1995.

\bibitem[Tieleman and Hinton(2017)]{tieleman2017divide}
T.~Tieleman and G.~Hinton.
\newblock Divide the gradient by a running average of its recent magnitude.
  coursera: Neural networks for machine learning.
\newblock \emph{Technical Report.}, 2017.

\bibitem[Weiss et~al.(2016)Weiss, Khoshgoftaar, and Wang]{weiss2016survey}
K.~Weiss, T.~M. Khoshgoftaar, and D.~Wang.
\newblock A survey of transfer learning.
\newblock \emph{Journal of Big data}, 3\penalty0 (1):\penalty0 9, 2016.

\bibitem[Yarats et~al.(2021)Yarats, Fergus, Lazaric, and
  Pinto]{yarats2021reinforcement}
D.~Yarats, R.~Fergus, A.~Lazaric, and L.~Pinto.
\newblock Reinforcement learning with prototypical representations.
\newblock In \emph{International Conference on Machine Learning (ICML)}, 2021.

\end{thebibliography}
\bibliographystyle{abbrvnat}

\clearpage

\appendix











\section{Supplemental Details}
\subsection{Algorithms Details}
\label{supp:rewards}
Here, we formally define all intrinsic rewards evaluated in the paper. 
Let's denote by $N(s)$ the state visitation (pseudo)count, by $N(c)$ the state-change visitation (pseudo)count, and by $\phi(s)$ some state embeddings parameterized by a neural network. Formally, the rewards $r_i(s,a,s')$ we evaluate are:
\begin{itemize}[nosep,leftmargin=*]
\item \textbf{C-BET (Ours)}: $r_i = 1 / {{(N(c) + N(s'))}}$. At pre-training, both counts are reset with probability $p \leq 1 - \gamma_i$ 
at every step. 
\item \textbf{Count}: $r_i = 1 / {\sqrt{N(s')}}$. $N(s')$ is never reset. 
\item \textbf{RND}: $r_i = ||\phi(s') - \hat{\phi}(s')||^2$, where $\hat{\phi}$ is a fixed random network, and $\phi$ is trained to minimize the same error.
\item \textbf{Curiosity}: $r_i = ||f(\phi(s),a) - \phi(s')||^2$, where $\phi$ is trained to minimize the prediction error of both an inverse and a forward model, and $f$ is the forward model.
\item \textbf{RIDE}: $r_i = {||\phi(s) - \phi(s')||^2} / {\sqrt{N(s')}}$, where $\phi$ is trained to minimize the prediction error of both an inverse and a forward model. $N(s')$ is reset at the beginning of every episode during both pre-training and transfer.
\end{itemize}

\subsection{Hyperparameter Details}
\label{supp:hyper}
%
Experiments are built on the codebase developed by \citet{raileanu2020ride}, which includes the Torchbeast implementation of IMPALA~\citep{torchbeast2019}. 
All algorithms use the same network architectures.
\begin{itemize}[nosep,leftmargin=*]
\item \textbf{Policy and value function}. The input is the environment partial observation, which is 7$\times$7$\times$3 for MiniGrid and 64$\times$64$\times$3 for Habitat. It is passed through three (for MiniGrid) or five (for Habitat) convolutional layers with 32 filters each, kernel size $3\!\times\!3$, stride 2, and padding 1. An exponential linear unit~\citep{clevert2015fast} is after each convolution layer. The output of the last convolution layer is fed into two layers with ReLU activation and 1,024 units, and then into an LSTM~\citep{hochreiter1997long} with 1,024 units. Finally, two separate fully connected layers of 1,024 units each are after the LSTM, and are used for the value function and the policy, respectively.
\item \textbf{State embeddings}. The partial observation is passed through five convolutional layers with 32 filters each, kernel size $3\!\times\!3$, stride 2, and padding 1.
\item \textbf{Inverse model}. The input is the concatenation of two successive state embeddings. It is fed into two layers with ReLU activation and 256 units.
\item \textbf{Forward model}. The input is the concatenation of the state embedding and the action. It is fed into two layers with ReLU activation and 256 and 128 units, respectively.
\end{itemize}

Observations and changes counts are based on egocentric and panoramic views, respectively. For MiniGrid, egocentric views are 147-dimensional while panoramic views are 588-dimensional. \\
For Habitat, views are encoded using HashSim~\citep{tang2017exploration,charikar2002similarity} to ease computational demands. 
Habitat views are 12,288- and 49,152-dimensional and are both hashed with 128 bits.

Value functions and policies are updated every 100 steps with IMPALA~\citep{espeholt2018impala}, using mini-batches of 32 samples and the RMSProp optimizer~\citep{tieleman2017divide} (learning rate 0.0001 linearly decaying to 0, momentum 0, and $\varepsilon = 0.00001$). Gradients are clipped to have maximum norm 40. 

Both the intrinsic and extrinsic rewards discount factors are $0.99$ as goals can be reached in less than 100 steps. This is also the default discount factor used by~\citet{raileanu2020ride}.
The counts reset probability is $p = 0.001$, meaning that the agent has an approximate `life expectancy' of 1,000 steps. This is a fitting choice because the largest time horizon for the environments is 640 steps (see Appendix~\ref{supp:minigrid}), and $p \leq 1 - \gamma_i$.

Intrinsic rewards and losses are further scaled down by a coefficient for numerical stability.
\begin{itemize}[nosep,leftmargin=*]
    \item Intrinsic reward: 0.1 (RIDE, RND, Curiosity), 0.005 (C-BET, Count).
    \item Policy entropy loss (all algorithms): 0.0005. 
    \item Value function loss (all algorithms): 0.5.
    \item Forward model loss (RIDE, Curiosity): 10. 
    \item Inverse model loss (RIDE, Curiosity): 0.1. 
    \item Random network loss (RND): 0.1.
\end{itemize}


\subsection{Compute Details}
\label{supp:compute}
Experiments were run on a SLURM-based cluster, using a NVIDIA Quadro GP100 GPU and 40 CPUs. 
For MiniGrid, IMPALA uses 40 actors.
One pre-training run takes \textasciitilde8 hours for 50M steps. 
Learning time after transfer ranges from \textasciitilde30 minutes (5M steps in \textit{Unlock}), to \textasciitilde55 hours (200M steps in \textit{ObstructedMaze2Dlhb}).
For Habitat, IMPALA uses 4 actors to prevent memory errors. One pre-training run takes \textasciitilde45 hours for 2M steps (because Habitat simulation is slower, network inputs --observations-- are larger, and true state counts are saved for visitation heatmaps).

\begin{figure*}[t]
\centering
\includegraphics[width=0.95\linewidth]{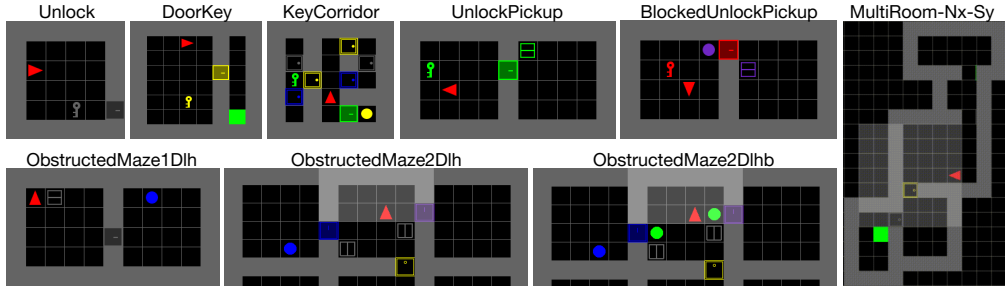}
\vspace*{-3pt}
\caption{\label{fig:minigrid_envs}\textbf{The MiniGrid environments.} The agent has to navigate through a grid and interact with different objects (keys, doors, boxes, balls) to fulfil a task. At each episode, the grids are procedurally generated, changing rooms layout, objects positioning and color. 
}
\end{figure*}

\subsection{MiniGrid Environments Details}
\label{supp:minigrid}
Figure~\ref{fig:minigrid_envs} shows the MiniGrid environments used in this paper. Below is a list of their respective tasks (the hardest are bolded). $T$ denotes the maximum number of steps per episode. Everything is as implemented by default in MiniGrid codebase.
\begin{itemize}[nosep,leftmargin=*]
\item \textit{Unlock}: pick up the key and unlock the door ($T = 288$).
\item \textit{DoorKey-8x8}: pick up the key, unlock the door, and go to green goal ($T = 640$).
\item \textit{KeyCorridorS3R3}: pick up the key, unlock the door, and pick up the ball (only the door before the ball is locked) ($T = 270$).
\item \textit{UnlockPickup}: pick up the key, unlock the door, and open the box ($T = 288$).
\item \textit{\textbf{BlockedUnlockPickup}}: pick up the ball in front of the door, drop it somewhere else, pick up the key, unlock the door, and open the box ($T = 576$).
\item \textit{ObstructedMaze1Dlh}: open the box to reveal the key, pick it up, unlock the door, and pick up the ball ($T = 288$).
\item \textit{{ObstructedMaze2Dlh}}: same as above, but with two doors to unlock ($T = 576$).
\item \textit{\textbf{ObstructedMaze2Dlhb}}: same as above, but with two balls in front of the doors (like in BlockedUnblockPickup) ($T = 576$).
\item \textit{\textbf{MultiRoom-N6}}: navigate through six rooms of maximum size ten and go to the green goal (all doors are already unlocked) ($T = 120$).
\item \textit{\textbf{MultiRoom-N12-S10}}: same as above, but with ten rooms of maximum size twelve ($T = 240$).
\item \textit{MultiRoom-N4-S5}: smaller MultiRoom environment (five rooms of maximum size five), used only for pre-training ($T = 100$).
\end{itemize}
In all tasks, the extrinsic reward is $r_t$ = 1 - 0.9 $(t / T)$.
This reward is given only for solving the task. 
\\
The action space is discrete with seven actions: left, right, forward, pick up, drop, toggle, and done. `Toggle' unlocks a door if the agent has the corresponding key, opens/closes a door if unlocked, and open boxes to reveal keys. `Done' is implemented by default in MiniGrid codebase and it is used for language-based tasks only, thus it does nothing in our tasks. 
\\
The grid is procedurally generated at each episode, and the agent's initial position is random within a fixed area far from the goal (e.g., in DoorKey the agent starts in the area with the key, or in MultiRooms it starts in the farthest room from the goal).
\\
BlockedUnblockPickup, ObstructedMaze2Dlh, and ObstructedMaze2Dlhb provide all types of interaction, but only ObstructedMazes have hidden keys. 
{MultiRooms} differ from the other environments because they have more rooms, all doors are already unlocked, and the goals are further away from the agent start position. 
They require more steps to be completed, and this explains their smallest extrinsic return compared to other environments in Figure~\ref{fig:results_minigrid}.

\clearpage

\subsection{MiniGrid Plot Smoothing Details}
\label{supp:smooth}
MiniGrid's plots show the mean and confidence interval over seven random seeds per method, smoothed over 300 epochs with a sliding window.
Each epoch consists of 3,200 samples (32 mini-batches of 100 steps). These samples are used for both updating the agent and computing evaluation metrics. This smoothing is necessary because we need to wait for the end of the episode to compute its return, but episodes are longer than 100 steps (see Appendix~\ref{supp:minigrid}). Thus, in some epochs we cannot compute the expected return and we end up with less than one evaluation data point per epoch. 
\\
On the contrary, for Habitat visitation counts we can always retrieve the number of visited states at every step without waiting for the end of the episode, therefore there is no need for smoothing.

\subsection{Habitat Environments Details}
\label{supp:habitat}
\begin{figure*}[hb]
\captionsetup[subfigure]{aboveskip=3pt,belowskip=-2pt}
\centering
\begin{subfigure}[t]{0.24\linewidth}
\includegraphics[width=\linewidth]{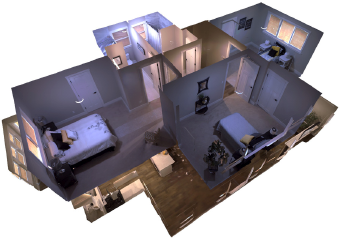}
\caption{\label{fig:habitat_apt0}Apartment 0}
\end{subfigure}
\hfill
\begin{subfigure}[t]{0.24\linewidth}
\includegraphics[width=\linewidth]{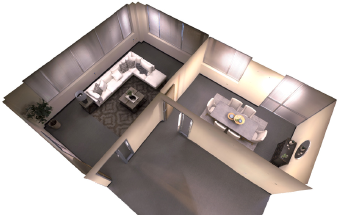}
\caption{\label{fig:habitat_apt1}Apartment 1}
\end{subfigure}
\hfill
\begin{subfigure}[t]{0.23\linewidth}
\includegraphics[width=\linewidth]{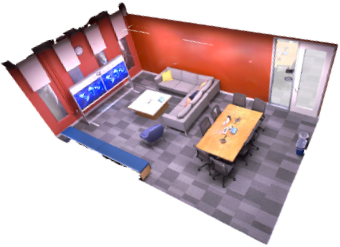}
\caption{\label{fig:habitat_apt2}Office 3}
\hfill
\end{subfigure}
\begin{subfigure}[t]{0.24\linewidth}
\includegraphics[width=\linewidth]{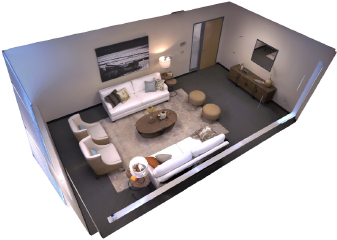}
\caption{\label{fig:habitat_room0}Room 0}
\end{subfigure}
\\[0.2em]
\begin{subfigure}[t]{0.24\linewidth}
\includegraphics[width=\linewidth]{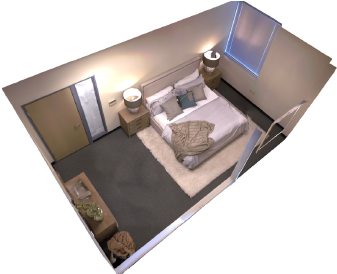}
\caption{\label{fig:habitat_room1}Room 1}
\end{subfigure}
\hfill
\begin{subfigure}[t]{0.24\linewidth}
\includegraphics[width=\linewidth]{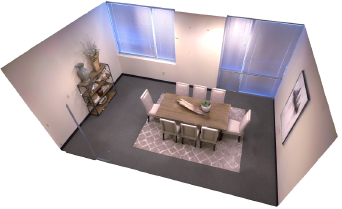}
\caption{\label{fig:habitat_room2}Room 2}
\end{subfigure}
\hfill
\begin{subfigure}[t]{0.24\linewidth}
\includegraphics[width=\linewidth]{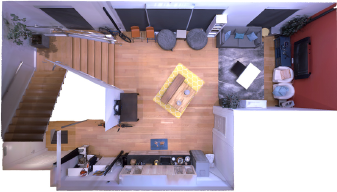}
\caption{\label{fig:habitat_frl_apt0}FRL Apartment 0}
\end{subfigure}
\hfill
\begin{subfigure}[t]{0.24\linewidth}
\includegraphics[width=\linewidth]{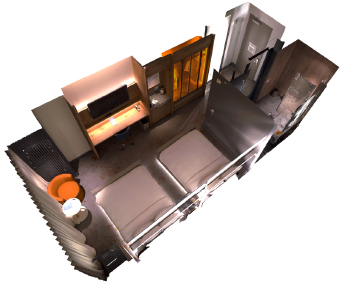}
\caption{\label{fig:habitat_hotel0}Hotel 0}
\end{subfigure}
\hfill
\caption{\label{fig:replica_scenes}\textbf{Real-world scenes from the Replica dataset used for our Habitat experiments.} The scenes have several rooms and obstacles that make exploration challenging. Apartment 0 is used for pre-training, while the remainder are used to evaluate transfer policies.} 
\end{figure*}

\begin{figure*}[hb]
\centering
\includegraphics[width=0.7\linewidth]{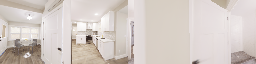}
\vspace*{-5pt}
\caption{\label{fig:habitat_obs}\textbf{Agent's $\boldsymbol{360^{\circ}}$ panoramic view of Apartment 0.} The view concatenates four egocentric images taken from $0^{\circ}$, $90^{\circ}$, $180^{\circ}$, and $270^{\circ}$ with respect to the North.}
\end{figure*}

Figure~\ref{fig:replica_scenes} shows the Replica scenes used in this paper. At each step, the agent can move forward by 0.25 meter or turn left/right by $10^{\circ}$. 
The episode length is $T = 500$ steps.
%
Egocentric views (used as policy input and to count states) have resolution 64$\times$64$\times$3. 
Figure~\ref{fig:habitat_obs} shows the agent's $360^{\circ}$ panoramic view, used to count environment changes.
For computational ease, we use the following static hashing to map both views to hash codes and count their occurrences with a hash table. 
\begin{equation}
\mathsf{cbet\_hash}(x) = 
\begin{cases}
1 & \text{if $g(x) > 0.5$} \\
- 1 & \text{if $g(x) < -0.5$}\\
0 & \text{otherwise,}
\end{cases}
\label{eq:hash}
\end{equation}
where $x$ is either an egocentric or panoramic view, and $g(x) = \mathsf{tanh}(Ax) + w$.
$A$ and $w$ are a projection matrix and a bias vector, respectively, both with i.i.d. entries drawn from the standard normal distribution. 
Our hashing is inspired by SimHash~\citep{charikar2002similarity} binary encoding $\mathsf{sim\_hash}(x) = \mathsf{sign}(Ag(x))$. 
In preliminary results, our ternary encoding performed better than SimHash.

\clearpage

\section{Ablation Study on MiniGrid}
\label{supp:ablation}
\begin{figure*}[h]
\begin{center}
\includegraphics[width=\linewidth]{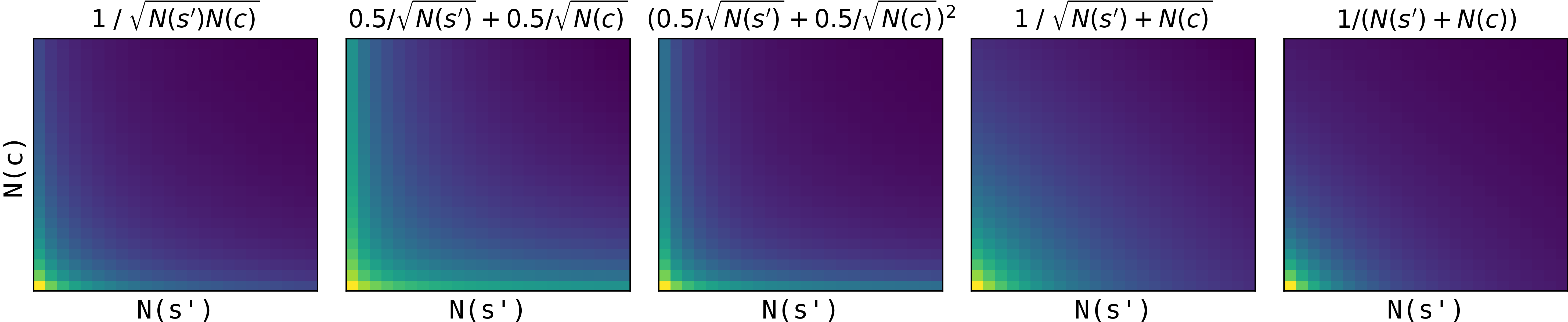}
\caption{\label{fig:reward_combo}Heatmaps of different intrinsic rewards on varying state and change counts. The brighter the color, the higher the reward. Considering the sum of the parts --either squared or not-- (second and third maps) still rewards transitions with either low state or change counts --not necessarily both. This is shown by the bright edges of the heatmap. The same happens with only the product (first map), albeit to a lesser extent.
If we sum the parts {inside} the square root (fourth and fifth map) we correctly reward only transitions with {both} low state and change counts. Between the latter two, squaring the reward further penalizes transitions with high counts, because the reward decreases faster as either the state or change count increases.
}
\end{center}
\end{figure*}

In this paper, we propose the following C-BET intrinsic reward:
\begin{align}
    r_i(s,a,s') &= \frac{1}{{\underbrace{N(c)}_{\text{change part}} + \underbrace{N(s')}_{\text{state part}}}}, \label{eq:cbet_parts}
\end{align}
where $c(s,s')$ is the {environment change} of a transition $(s,a,s')$, and $N$ denotes (pseudo)counts of changes and states. Both counts are randomly reset at any given time step, independently from each other.
We use this reward for two reasons. First, as shown in Figure~\ref{fig:reward_combo}, summing the two parts {\textbf{within}} the square root yields high rewards {\textbf{only}} to transitions with {\textbf{both}} low state and change counts. Second, the ablation study in Appendix~\ref{supp:abl_cbet_rewards} shows that squaring the reward performs best. 

\paragraph{Aim of the Study}\mbox{}\\[1em]
The key components of C-BET reward are count random resets and the change-based reward.
In this section, we investigate these components and answer the following fundamental questions:
\begin{itemize}[nosep,leftmargin=*]
    \item How important are count resets? 
    \item How important is how we encode changes? Are panoramic views crucial for C-BET? 
    \item How do different combinations of the state and change terms in Figure~\ref{fig:reward_combo} perform? Is Eq.~\ref{eq:cbet_parts} truly the best? 
    \item How do pre-training environments affect the quality of the exploration policy?
\end{itemize}
We start by investigating if panoramic views are good for counting states, and if squaring the reward can help state-only baseline as well (Appendix~\ref{supp:abl_pano_state}).
We continue by comparing the different types of rewards shown in Figure~\ref{fig:reward_combo} (Appendix~\ref{supp:abl_cbet_rewards}). Subsequently, we evaluate C-BET with egocentric views for counting changes (Appendix~\ref{supp:abl_ego_vs_pano}). Then, we further review interesting questions arising from such evaluation regarding change-based rewards (Appendix~\ref{supp:abl_change_only}). 
Finally, we study the importance of count resets.
Finally, we conclude by showing what really matters for C-BET (Appendix~\ref{supp:abl_resets}).

As evaluation metric, we consider the `average episode success' of pre-trained policies, in both the \textit{one-to-many} (SingleEnv) and \textit{many-to-many} (MultiEnv) setups discussed in Section~\ref{sec:eval}. 
After pre-training, policies are transferred to new environments and have to solve their respective task. Without further training, we evaluate their average success rate over 100 episodes.
The higher the success rate, the faster the policy is likely to solve the task if further extrinsic-reward training would be carried one, as the extrinsic reward will be seen more often.
\\
We present results through error bar plots and recap tables, reporting mean and confidence interval over seven seeds.
Error bar plots show the success rate for each new environment a policy is transferred to, and each setup (SingleEnv and MultiEnv) has a dedicated plot.
Figure legends average the rate over the ten environments. Tables further average it over the three setups.

\clearpage

\subsection{State-Only Counts: Egocentric vs Panoramic Views, Squaring vs Classic Reward}
\label{supp:abl_pano_state}

\begin{figure*}[h]
\begin{center}
\begin{tabular}{@{\extracolsep{\fill}}|>{\columncolor{colorcbet4!70}}c>{\columncolor{colorcbet8!70}}c>{\columncolor{colorcbet4!70}}c>{\columncolor{colorcbet8!70}}c|}
\hline
{\small$\dfrac{1}{\sqrt{N(s')}}$}, Ego & {\small$\dfrac{1}{{N(s')}}$}, Ego & {\small$\dfrac{1}{\sqrt{N(s')}}$}, Pano & {\small$\dfrac{1}{{N(s')}}$}, Pano \\
\hline
\textcolor{black}{\textbf{28.35\%}} & \textcolor{black}{\textbf{29.08\%}} & 24.90\% & 20.93\%\\
\hline
\end{tabular}
\\[3pt]
\includegraphics[width=0.99\linewidth]{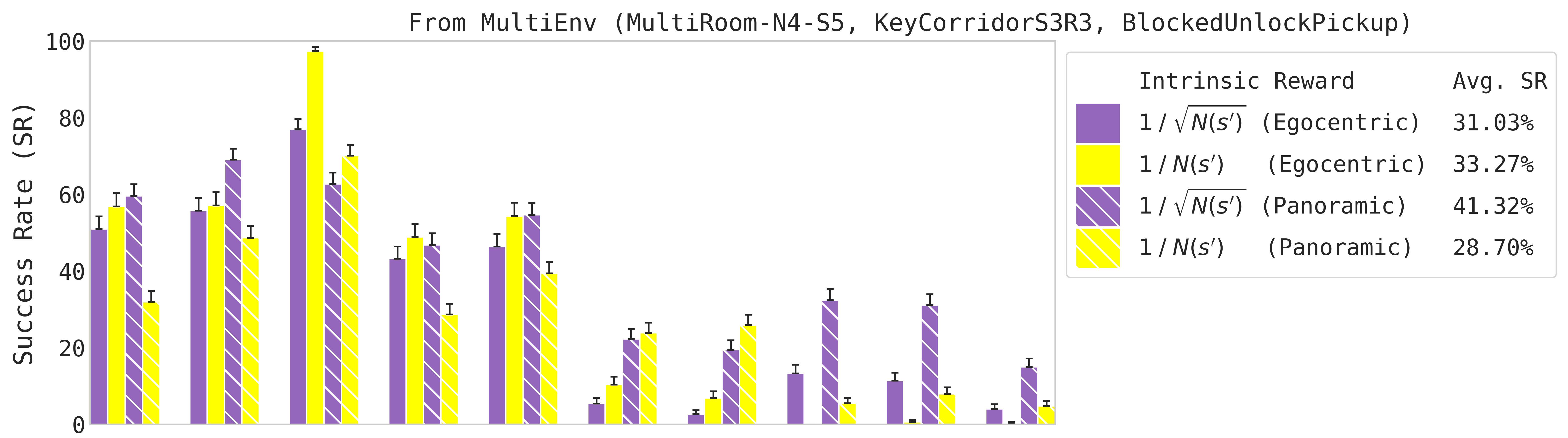}
\\[0em]
\includegraphics[width=0.99\linewidth]{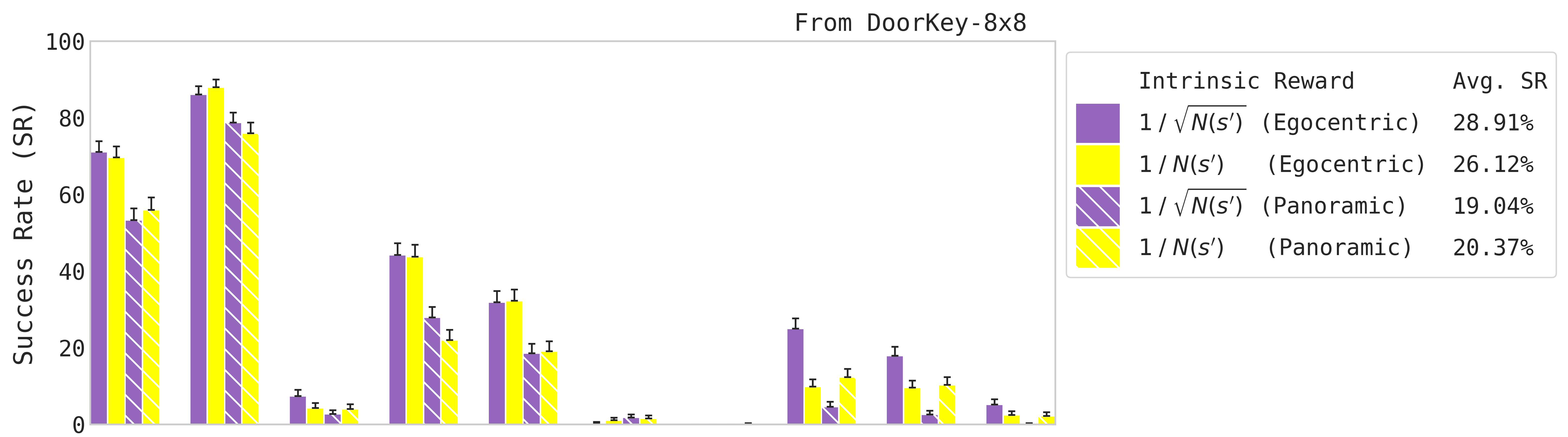}
\\[0em]
\includegraphics[width=0.99\linewidth]{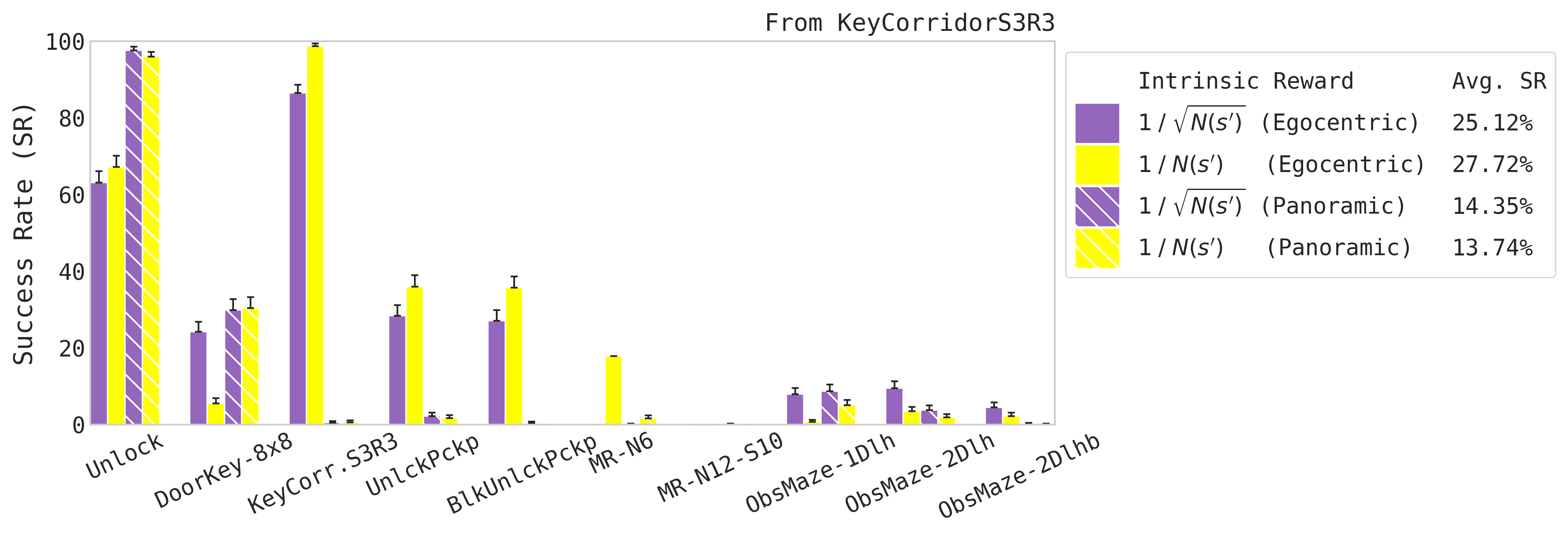}
\vspace{-7pt}
\caption{\label{fig:abl1}Success rate of policies pre-trained on different state-only intrinsic rewards with count resets. Squared rewards tend to overfit to the training environments. Panoramic views perform worse than egocentric views in SingleEnvs, but better in MultiEnv due to the diversity of visited states.}
\end{center}
\end{figure*}

\textbf{Questions:} Are panoramic views good for counting states? Does squaring the state-only baseline increase its performance?
\\
\textbf{Discussion:} Figure~\ref{fig:abl1} shows that egocentric views (plain patch) perform best, with an average success rate of 28.3\% over the three setups. Panoramic views (dashed patch) perform well in MultiEnv but poorly in SingleEnvs, for an average success rate of 24.9\%.
The reason is that panoramic views are rotation-invariant: when the agent turns left/right the panoramic state does not change, the state count increases and the intrinsic reward decreases. In the end, the agent will not be encourage to turn at all. This is not surprising, because panoramic views are designed to represent \textit{environment changes} rather than the agent's state\footnote{Note that panoramic views are used only for the intrinsic reward. The policy still receives egocentric views.}. 
Only in MultiEnv panoramic views perform well. Thanks to the diversity of visited states, counts do not increase too rapidly, thus turning around is not penalized too much. In Appendix~\ref{supp:abl_change_only} we further investigate this issue with `change-only counts` rewards.
\\
Figure~\ref{fig:abl1} also shows that squaring the reward (\textcolor{colorcbet8}{yellow}) performs slightly worse, since policies seem to overfit to the training environments.

\clearpage

\subsection{C-BET: Different Reward Combinations}
\label{supp:abl_cbet_rewards}
\begin{figure*}[h]
\begin{center}
\setlength{\tabcolsep}{2pt}
\begin{tabular}{@{\extracolsep{\fill}}|>{\columncolor{colorcbet1!70}}c>{\columncolor{colorcbet2!70}}c>{\columncolor{colorcbet6!70}}c>{\columncolor{colorcbet3!70}}c>{\columncolor{colorcbet7!70}}c>{\columncolor{colorcbet5!70}}c|}
    \hline
    {\small$\dfrac{1}{\sqrt{N(s')N(c)}}$} & {\small$\dfrac{0.5}{\sqrt{N(s')}} \!+\! \dfrac{0.5}{\sqrt{N(s')}}$} & {\small$\left(\vcenter{\hbox{$\dfrac{0.5}{\sqrt{N(s')}} \!+\! \dfrac{0.5}{\sqrt{N(s')}}$}}\right)^2$} & {\small$\dfrac{1}{\sqrt{N(c) + N(s')}}$} & {\small$\dfrac{1}{{N(c) + N(s')}}$} & {\small$\dfrac{1}{\sqrt{N(c)}}$} \\[9pt]
    \hline
    30.43\% & 30.40\% & \textcolor{black}{{31.92\%}} & 30.45\% & \textcolor{black}{\textbf{33.64\%}} & 18.15\% \\
    \hline
\end{tabular}
\\[3pt]
\includegraphics[width=0.99\linewidth]{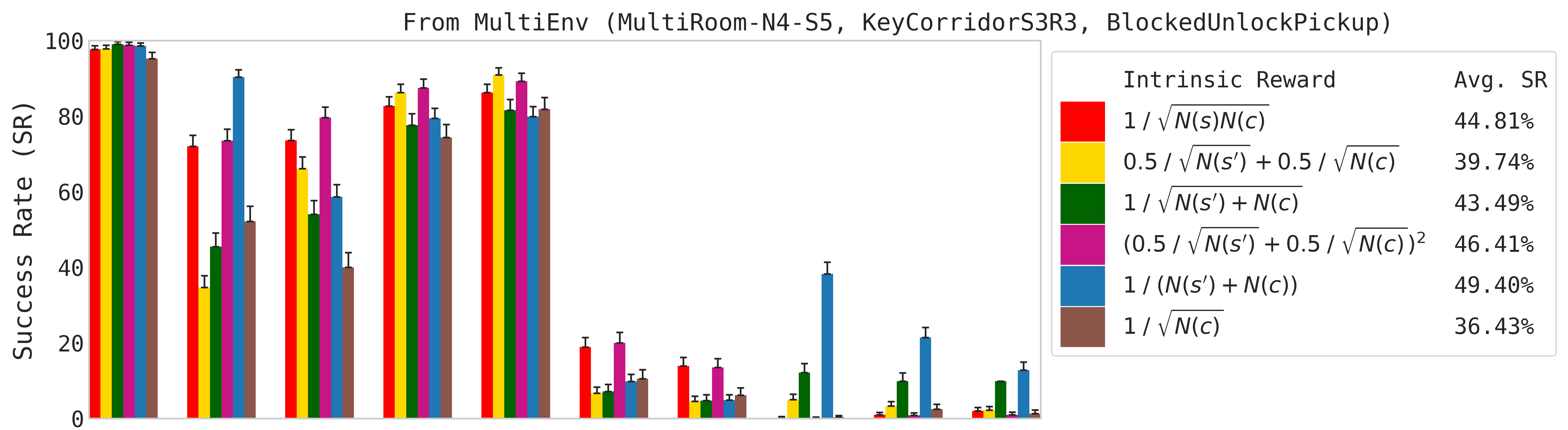}
\\[0em]
\includegraphics[width=0.99\linewidth]{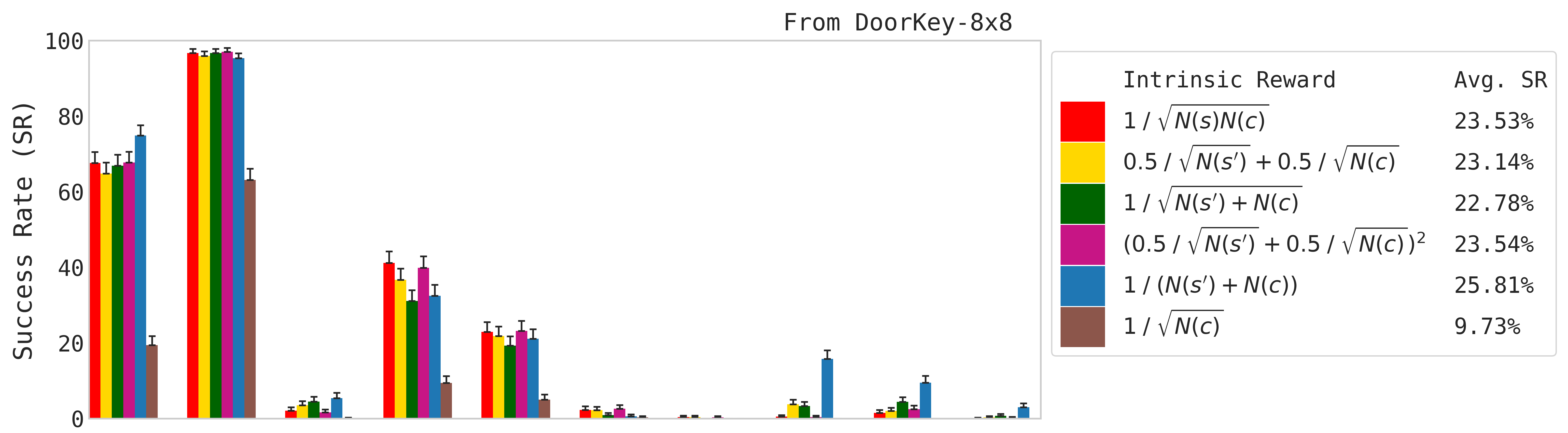}
\\[0em]
\includegraphics[width=0.99\linewidth]{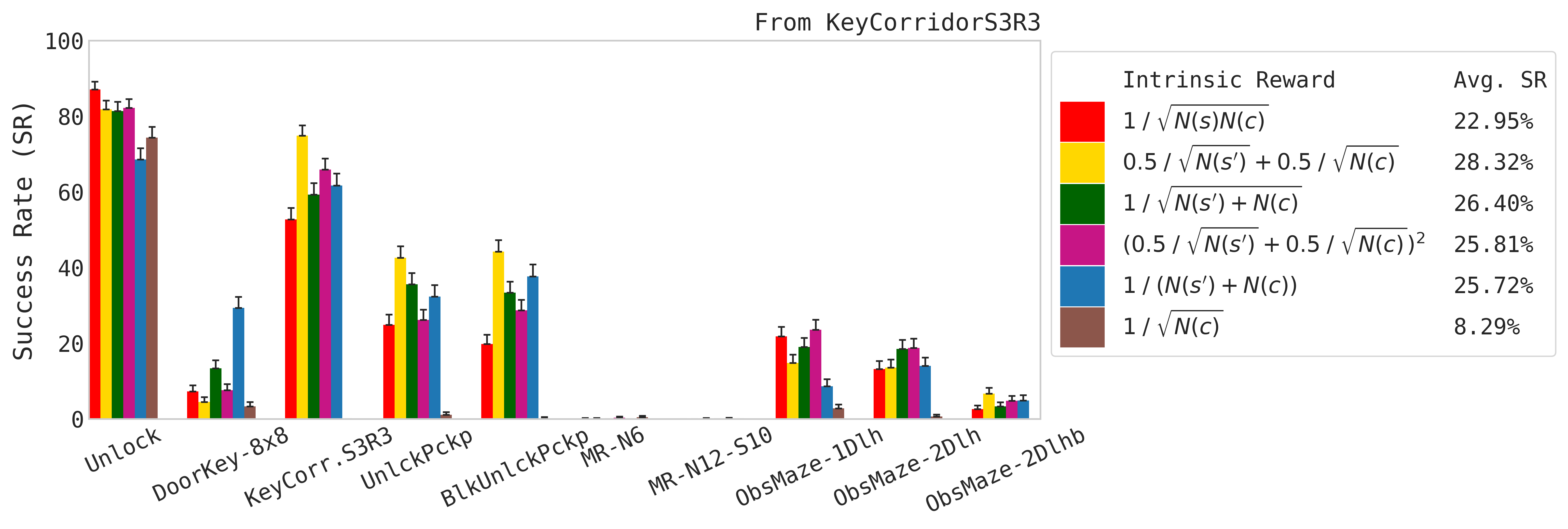}
\vspace{-7pt}
\caption{\label{fig:abl2a}Success rate of policies pre-trained on different combinations of state and change count rewards, with panoramic changes and random resets. States are always counted with egocentric views. C-BET (\textcolor{colorcbet7}{blue}) outperforms other rewards, achieving the highest overall success rate and being the only transferring well to ObstructedMazes.}
\end{center}
\end{figure*}

\textbf{Question:} How do different combinations of state and change rewards perform? Is Eq.~\eqref{eq:cbet_parts} better than other combinations?
\\
\textbf{Discussion:} Figures~\ref{fig:abl2a} shows that Eq.~\eqref{eq:cbet_parts} reward indeed performs best, achieving the highest success rate (\textcolor{colorcbet7}{blue}, 33.64\%). Furthermore, it also generalizes to the most environments, as it is the only showing good transfer to ObstructedMazes, especially when pre-trained on MultiEnv. Considering both the sum and the product of the parts (\textcolor{colorcbet6}{violet}, 31.9\%) comes in a close second, but does not transfer to ObstructedMazes when pre-trained on MultiEnv. Other reward combinations perform comparably (\textcolor{colorcbet1}{red}, \textcolor{colorcbet2}{gold}, \textcolor{colorcbet3}{green}, 30.4\%). Nonetheless, they all perform better than `state-only count' (\textcolor{colorcbet4}{purple}, 28.35\%, Figure~\ref{fig:abl1}). 
Interestingly, `change-only count' perform poorly (\textcolor{colorcbet5}{brown}, 18.15\%). We will come back to this in Appendix~\ref{supp:abl_change_only}.

\clearpage

\subsection{C-BET: Egocentric Views}
\label{supp:abl_ego_vs_pano}
\textbf{Question:} How does C-BET perform when environment changes are encoded with egocentric views?  
\\
\textbf{Discussion:} With egocentric views, the performance of all rewards decreases. This is not surprising, because panoramic views better encode environment changes, being rotation-invariant w.r.t. the orientation of the agent. 
The key to better exploration, thus, is the combination of agent-centric encoding of the state (egocentric views) and environment-centric encoding of the change (panoramic views). This conclusion is also strengthened by the following fact: neither `state-only count' (Figure~\ref{fig:abl1}, \textcolor{colorcbet4}{purple} and \textcolor{colorcbet8}{yellow}) nor `change-only count' (\textcolor{colorcbet5}{brown}) achieve the same success rate and transfer generalization as C-BET (\textcolor{colorcbet7}{blue}). \textit{C-BET does not perform better because of panoramic change counts, but because of how it combines them with egocentric state counts.}

\begin{figure*}[h]
\begin{center}
\setlength{\tabcolsep}{2pt}
\begin{tabular}{@{\extracolsep{\fill}}|>{\columncolor{colorcbet1!70}}c>{\columncolor{colorcbet2!70}}c>{\columncolor{colorcbet6!70}}c>{\columncolor{colorcbet3!70}}c>{\columncolor{colorcbet7!70}}c>{\columncolor{colorcbet5!70}}c|}
    \hline
    {\small$\dfrac{1}{\sqrt{N(s')N(c)}}$} & {\small$\dfrac{0.5}{\sqrt{N(s')}} \!+\! \dfrac{0.5}{\sqrt{N(s')}}$} & {\small$\left(\vcenter{\hbox{$\dfrac{0.5}{\sqrt{N(s')}} \!+\! \dfrac{0.5}{\sqrt{N(s')}}$}}\right)^2$} & {\small$\dfrac{1}{\sqrt{N(c) + N(s')}}$} & {\small$\dfrac{1}{{N(c) + N(s')}}$} & {\small$\dfrac{1}{\sqrt{N(c)}}$} \\[9pt]
    \hline
    \textbf{27.19\%} & 19.61\% & \textcolor{black}{\textbf{26.95\%}} & 17.33\% & \textcolor{black}{{21.69\%}} & 6.64\% \\
    \hline
\end{tabular}
\\[3pt]
\includegraphics[width=0.88\linewidth]{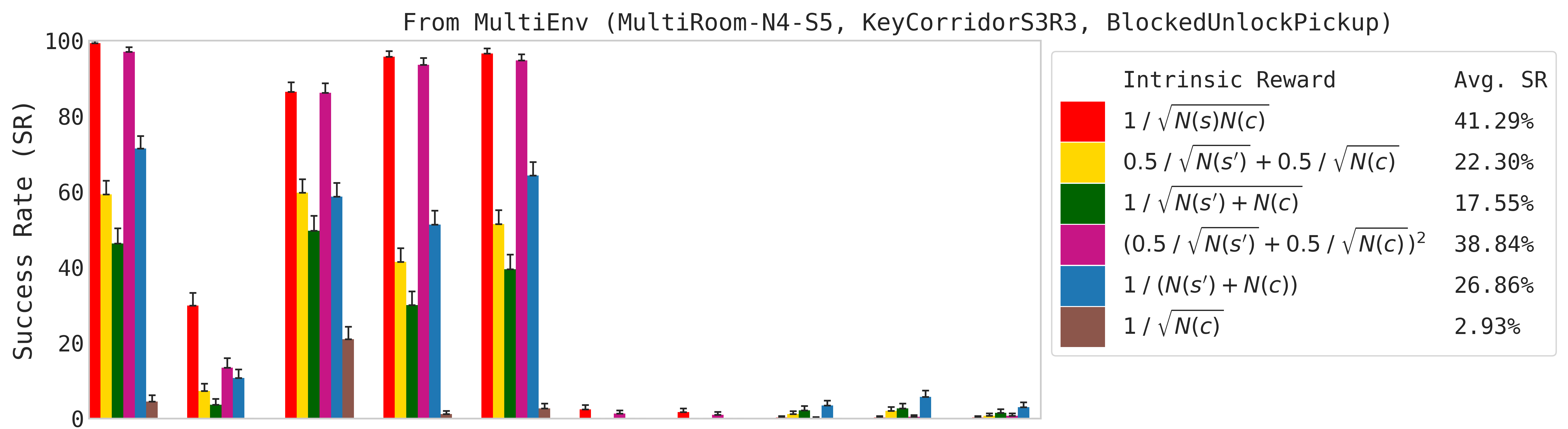}
\\[0em]
\includegraphics[width=0.88\linewidth]{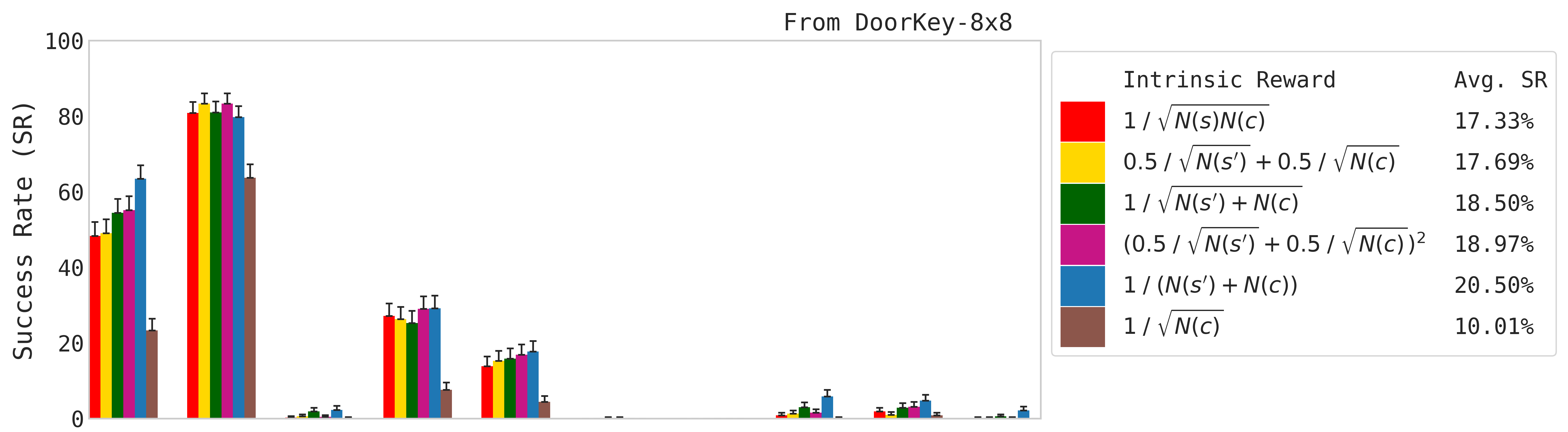}
\\[0em]
\includegraphics[width=0.88\linewidth]{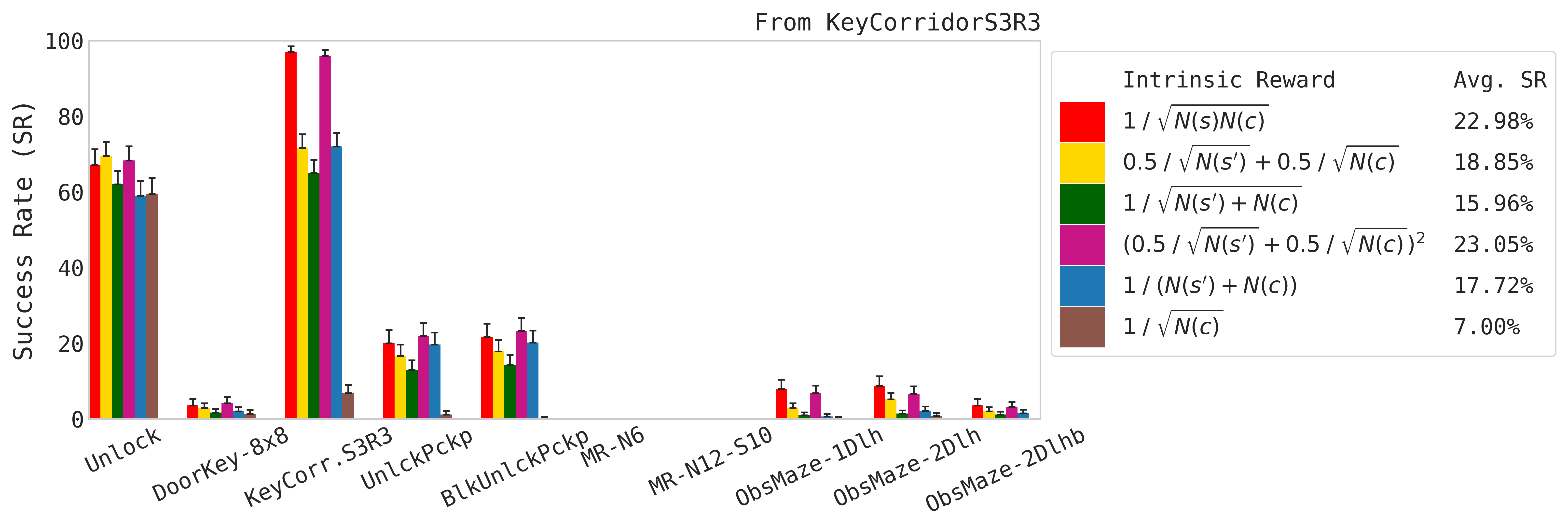}
\vspace{-7pt}
\caption{\label{fig:abl2b}Compared to Figure~\ref{fig:abl2a}, egocentric changes decrease the performance of all rewards, especially the one based only on change counts (\textcolor{colorcbet5}{brown}). Despite its lower success rate, Eq.~\eqref{eq:cbet_parts} still transfer well to many environments, including ObstructedMazes when pre-trained on MultiEnv.}
\end{center}
\end{figure*}

\subsection{C-BET: Why Do `Change-Only Counts' Perform Poorly?}
\label{supp:abl_change_only}
\textbf{Question:} In the previous sections, `change-only count' (\textcolor{colorcbet5}{brown}) performed poorly compared to other rewards. However, `state-count only' (\textcolor{colorcbet4}{purple}) does not. Why is that so? 
\\
\textbf{Discussion:} 
First, `panoramic change-only counts' performs poorly only in SingleEnvs and not in MultiEnv. This is similar to what we have seen in Figure~\ref{fig:abl1} with `state-only counts': in SingleEnvs panoramic change counts for turning left/right increase too fast and rewards decay too rapidly, while in MultiEnv the diversity of states prevents that. Thus, the resulting policy will learn not to turn.

\clearpage

\begin{figure*}[h]
\begin{center}
\vspace*{-3pt}
\begin{tabular}{@{\extracolsep{\fill}}|l|>{\columncolor{colora0!70}}c>{\columncolor{colora1!70}}c>{\columncolor{colora2!70}}c>{\columncolor{colora3!70}}c>{\columncolor{colora4!70}}c>{\columncolor{colora5!70}}c>{\columncolor{colora6!70}}c||c|}
    \hline
     & Left & Right & Forward & Pick & Drop & Toggle & Done & Entropy \\
    \hline
    C-BET & {10.7\%} & 11.1\% & \textcolor{black}{{51.5\%}} & 7.5\% & \textcolor{black}{{7.0\%}} & 9.3\% & 2.6\% & 0.78
    \\
    Pano Change-Only & {11.5\%} & 9.8\% & \textcolor{black}{{58.4\%}} & 5.5\% & \textcolor{black}{{5.1\%}} & 9.5\% & 0.05\% & 0.68
    \\
    Ego Change-Only & {34.9\%} & 35.0\% & \textcolor{black}{{24.7\%}} & 2.0\% & \textcolor{black}{{1.9\%}} & 1.2\% & 0.1\% & 0.66
    \\
    \hline
\end{tabular}
\\[3pt]
\includegraphics[width=0.75\linewidth]{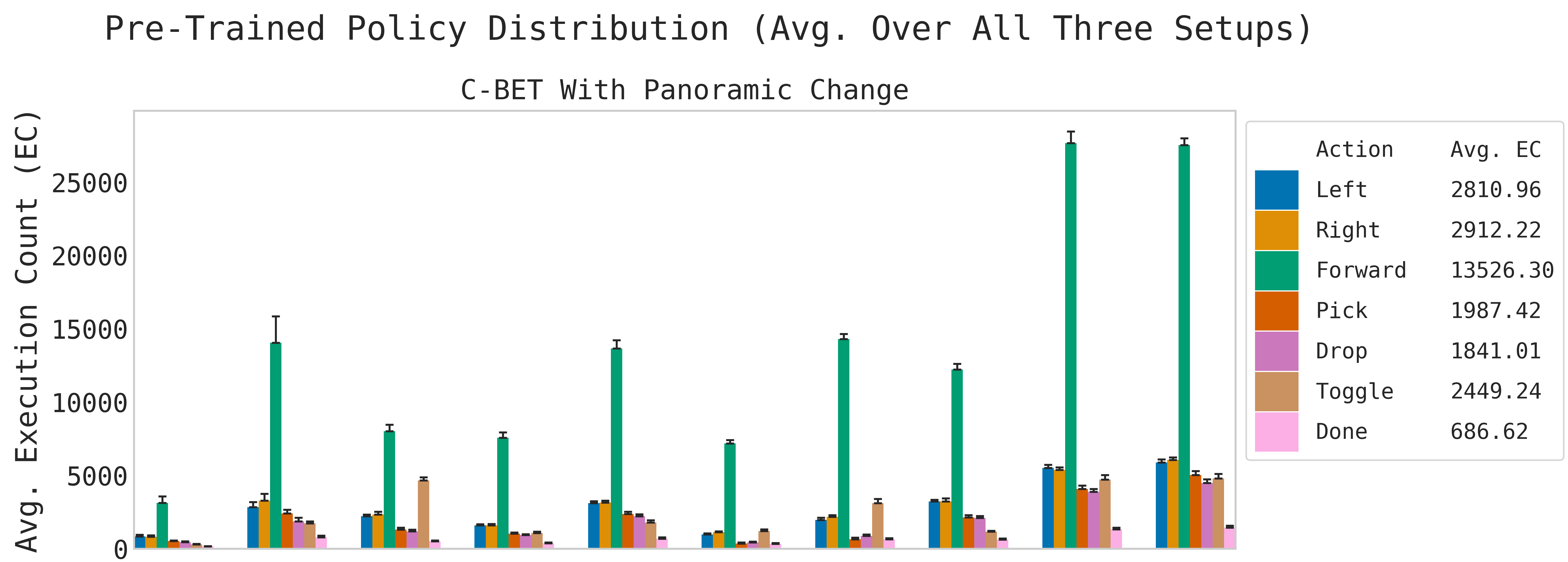}
\\[0em]
\includegraphics[width=0.75\linewidth]{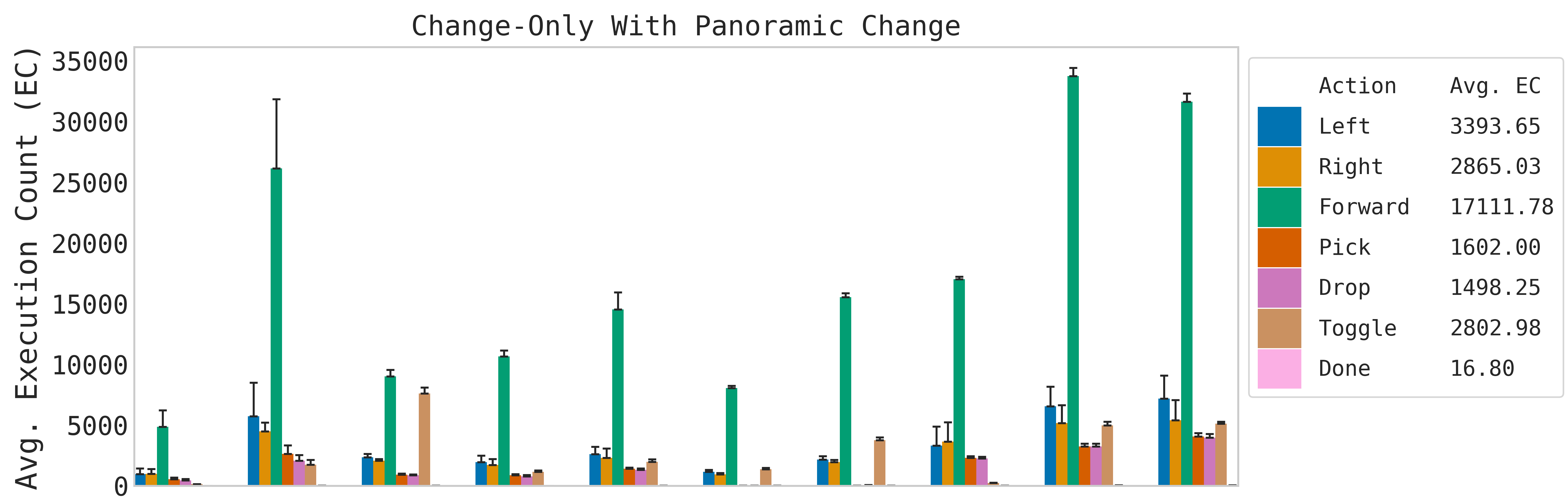}
\\[0em]
\includegraphics[width=0.75\linewidth]{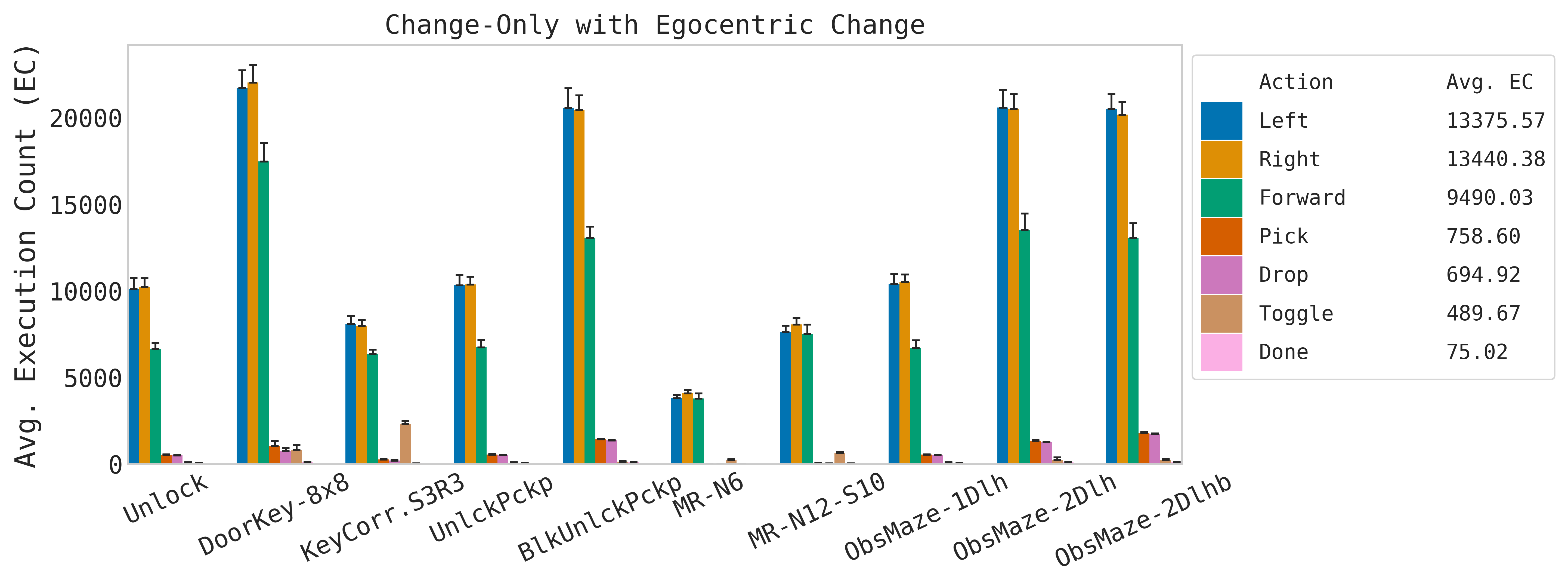}
\vspace{-8pt}
\caption{\label{fig:abl3}Plots show how many times pre-trained policies executed an action during 100 episodes when transferred to new environments. Counts are averages over both SingleEnvs and MultiEnv transfers. 
The table reports the resulting action probability. Last column shows the normalized entropy of the distribution (a random distribution has entropy 1).
`Panoramic change-only' overfits to moving forward, while `egocentric change-only' to turns.
On the contrary, C-BET moves less, interacts more, and overfit less to any action as shown by its higher entropy.
}
\end{center}
\end{figure*}

Then why does `change-only count' perform even worse with egocentric changes, especially in MultiEnv? In this case, the problem is the opposite. Turning left/right is the only action \textbf{always} resulting in some egocentric change, while other actions (including moving forward) can fail and thus produce no change. Furthermore, in visually rich environmnents, such as KeyCorridor, the egocentric change is likely to be unique. Therefore, the policy will overfit to turning left/right. \\
This is confirmed by Figure~\ref{fig:abl3}, showing the policy distribution at the end of pre-training. When trained with panoramic changes (second plot), `change-only counts' overfits to moving forward. When trained with egocentric views (third plot), the policy is drastically different and prefers to turn most of the time. 
On the contrary, C-BET policy (first plot) moves less and interacts more.
\\
Interestingly, C-BET also does nothing (`done') more than other policies, but this can be explained by recalling what we discussed in Section~\ref{subsec:pretrain}: the optimal exploration policy should keep some randomness to visit the environment uniformly.
In practice, C-BET does not have to always interact with the environment if the resulting change is not diverse. That is, C-BET intrinsic reward tries to keep uniform state/changes counts over \textbf{all} possible state/changes, including `no change' counts.
\\
Finally, Figure~\ref{fig:abl3} also shows how the action distribution changes depending on the environment.
In larger environments (DoorKey, BlockedUnblockPickup, MultiRooms and ObstructedMazes), the agent moves more.
In environments where doors are already unlocked (KeyCorridor and MultiRooms) the agent toggles more and picks/drops fewer times.

\clearpage

\subsection{Counts: No Resets vs Random Resets}
\label{supp:abl_resets}
\begin{figure*}[h]
\begin{center}
\includegraphics[width=0.95\linewidth]{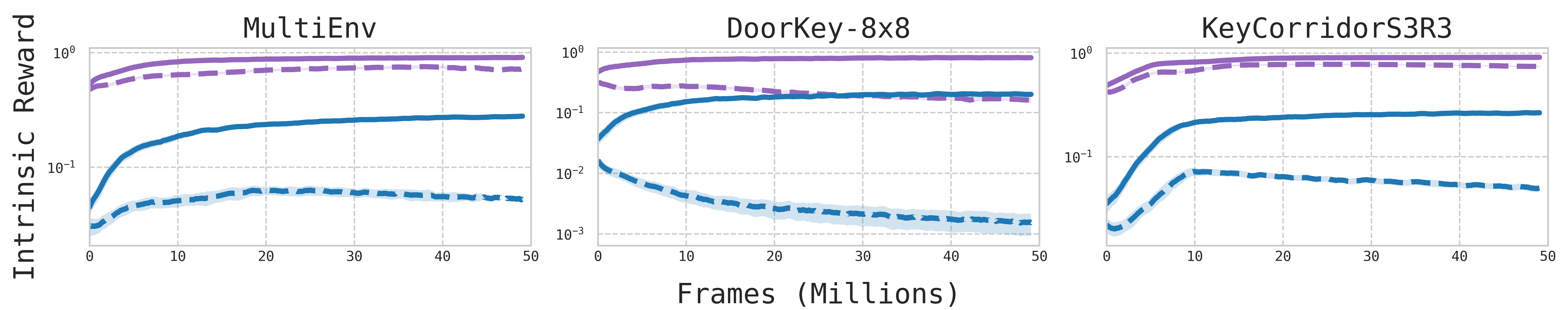}
\vspace{-7pt}
\caption{\label{fig:abl4_resets_intrinsic}Log scale intrinsic rewards trend at pre-training. Without resets, rewards decay preventing further learning. This is prominent in DoorKey, where states are similar and changes sparse.}
\vspace*{-5pt}
\end{center}
\end{figure*}

\begin{figure*}[h]
\begin{center}
\vspace*{-5pt}
\begin{tabular}{@{\extracolsep{\fill}}|>{\columncolor{colorcbet4!70}}c>{\columncolor{colorcbet7!70}}c>{\columncolor{colorcbet4!70}}c>{\columncolor{colorcbet7!70}}c|}
\hline
{\small$\dfrac{1}{\sqrt{N(s')}}$}, Resets & {\small$\dfrac{1}{{N(s') + N(c)}}$}, Resets & {\small$\dfrac{1}{\sqrt{N(s')}}$}, No Resets & {\small$\dfrac{1}{{N(s') + N(c)}}$}, No Resets \\
\hline
\textcolor{black}{{28.35\%}} & \textcolor{black}{\textbf{33.67\%}} & 27.21\% & 18.73\%\\
\hline
\end{tabular}
\\[3pt]
\includegraphics[width=0.93\linewidth]{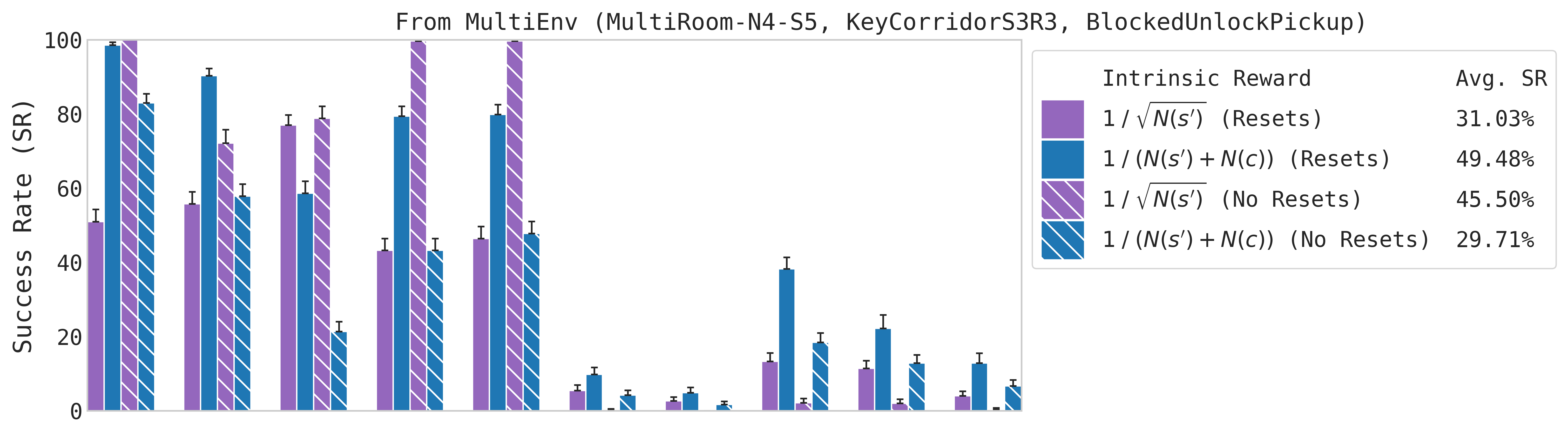}
\\[0em]
\includegraphics[width=0.93\linewidth]{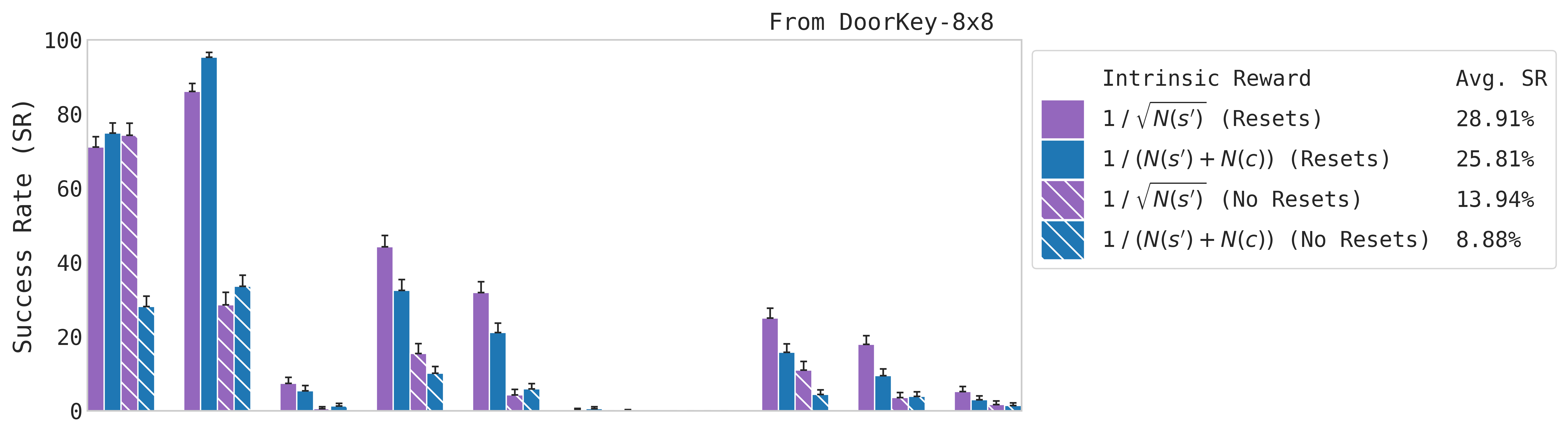}
\\[0em]
\includegraphics[width=0.93\linewidth]{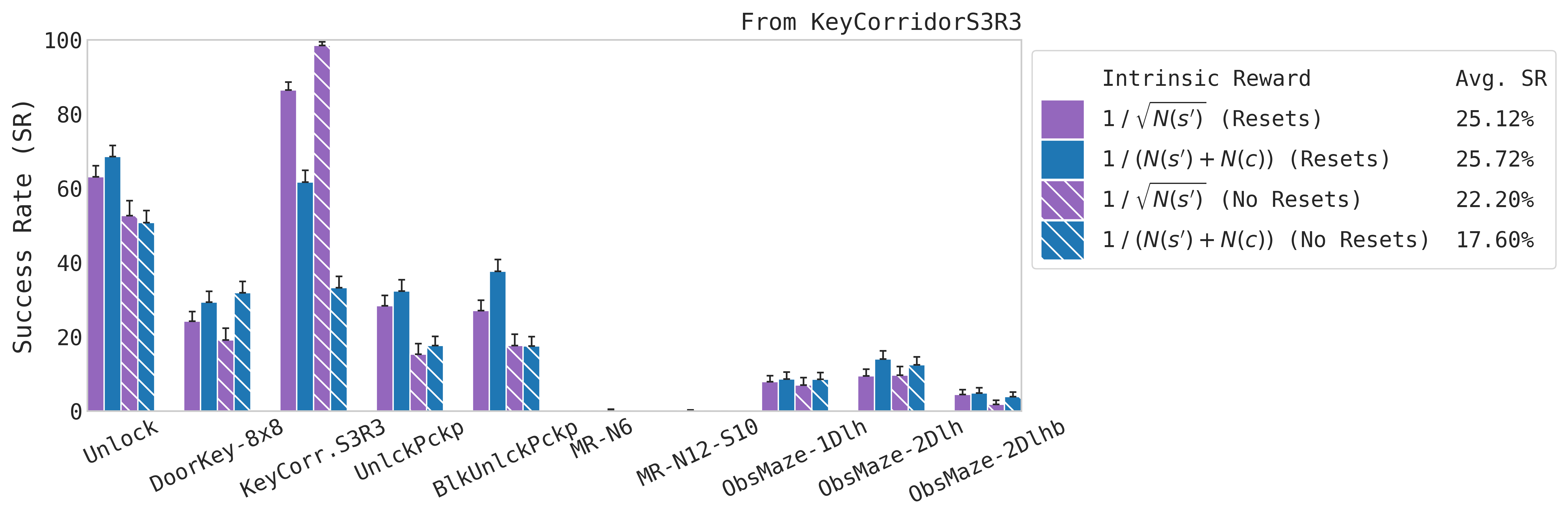}
\vspace{-7pt}
\caption{\label{fig:abl4_resets}Due to intrinsic rewards decay, policies cannot be properly trained, and do not transfer well. Yet, C-BET still transfers well to all environments when pre-trained in MultiEnv.}
\end{center}
\end{figure*}

\textbf{Questions:} Are count random resets necessary for intrinsic-only learning?
\\
\textbf{Discussion:} Figure~\ref{fig:abl4_resets} and its table show that not resetting counts achieves lower success rate. In particular, transfer is significantly worse when the agent is pre-trained on DoorKey.  
The reason for the worse performance --especially in DoorKey-- is shown in Figure~\ref{fig:abl4_resets_intrinsic}. As the agent explores, intrinsic rewards decay due to counts growth. The decay is more prominent in DoorKey because of its sparse changes and similar states, but it is noticeable also in KeyCorridor and MultiEnv.
\\
Interestingly, `state-count only' (\textcolor{colorcbet4}{purple}) achieves a better success rate without resets when pre-trained in MultiEnv. However, Figure~\ref{fig:abl4_resets} also shows that the policy overfits to training environments, not showing any transfer to MultiRooms and ObstructedMazes. 
On the contrary, despite the overall success rate, C-BET (\textcolor{colorcbet7}{blue}) still transfers well to all environments from MultiEnv.


\clearpage

\subsection{Conclusion of the Study}
\label{supp:abl_end}
The ablation study above has provided the answers to the fundamental questions we asked at the beginning.
\begin{itemize}[nosep,leftmargin=15pt]
    \item[1.] In Appendix~\ref{supp:abl_pano_state} and~\ref{supp:abl_change_only}, state-only and change-only counts with panoramic views did not perform well, thus \textit{panoramic views, alone, are not sufficient for learning good exploration policies}.
    \item[2.] In Appendix~\ref{supp:abl_cbet_rewards} and~\ref{supp:abl_ego_vs_pano}, combining egocentric views for counting states and panoramic views for counting changes performs best, thus \textit{the most important contribution of C-BET is the combination of agent-centric (egocentric states) and environment-centric (panoramic changes) exploration}. The design of the reward is also important, albeit to a lesser extent.
    \item[3.] In Appendix~\ref{supp:abl_resets}, resetting counts prevents the vanishing of intrinsic rewards, thus \textit{random resets are also a crucial contribution of C-BET}.
    \item[4.] In all experiments, policies transfers better when pre-trained in MultiEnv --especially C-BET-- thus \textit{pre-training by exploring multiple environments allows better generalization}. Similarly, transfer is harder when pre-training environments have similar states or sparse changes.
\end{itemize}

\clearpage

\section{MiniGrid Pre-Training Supplemental Results}
\label{supp:mini_extra}
\vspace*{-5pt}
In this paper, we argued that interacting with the environment while looking for rare changes helps finding extrinsic rewards faster. Thus, we evaluated policies on the number of interactions with the environment and on their success rate. Here we further elaborate the results presented in Section~\ref{sec:minigrid}.

\vspace*{-3pt}

\subsection{Unique Interactions per Episode at Pre-Training and at Offline Transfer}
\begin{figure*}[h]
\begin{center}
\vspace{-7pt}
\includegraphics[width=0.9\linewidth]{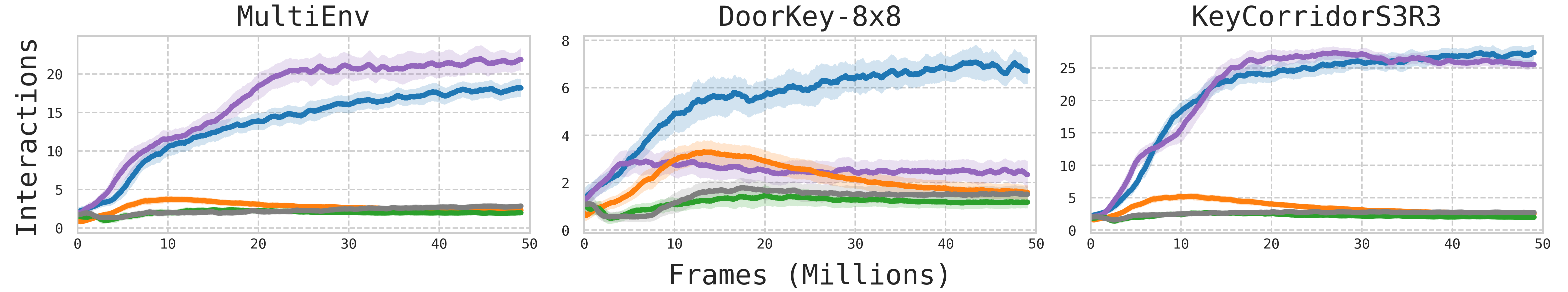}
\vspace{-8pt}
\caption{\label{fig:mini_extra_inter}Trend of unique interactions per episode, i.e., picks/drops/toggles resulting in unseen changes. For instance, repeatedly picking and dropping the same key in the same cell results in only two interactions.
Only C-BET interacts more with all pre-training environments as it explores.}
\vspace*{-5pt}
\end{center}
\end{figure*}

\begin{figure*}[h]
\begin{center}
\vspace*{-10pt}
\begin{tabular}{@{\extracolsep{\fill}}|>{\columncolor{colorcbet!70}}c>{\columncolor{colorcount!70}}c>{\columncolor{colorride!70}}c>{\columncolor{colorcuriosity!70}}c>{\columncolor{colorrnd!70}}c>{\columncolor{white}}c|}
    \hline
    {C-BET} & Count & RIDE & Curiosity & RND & Random \\
    \hline
    \textcolor{black}{\textbf{26.92}} & 24.72 & \textcolor{black}{{8.32}} & 7.18 & 7.34 & 7.40 \\
    \hline
\end{tabular}
\\[1pt]
\includegraphics[width=0.79\linewidth]{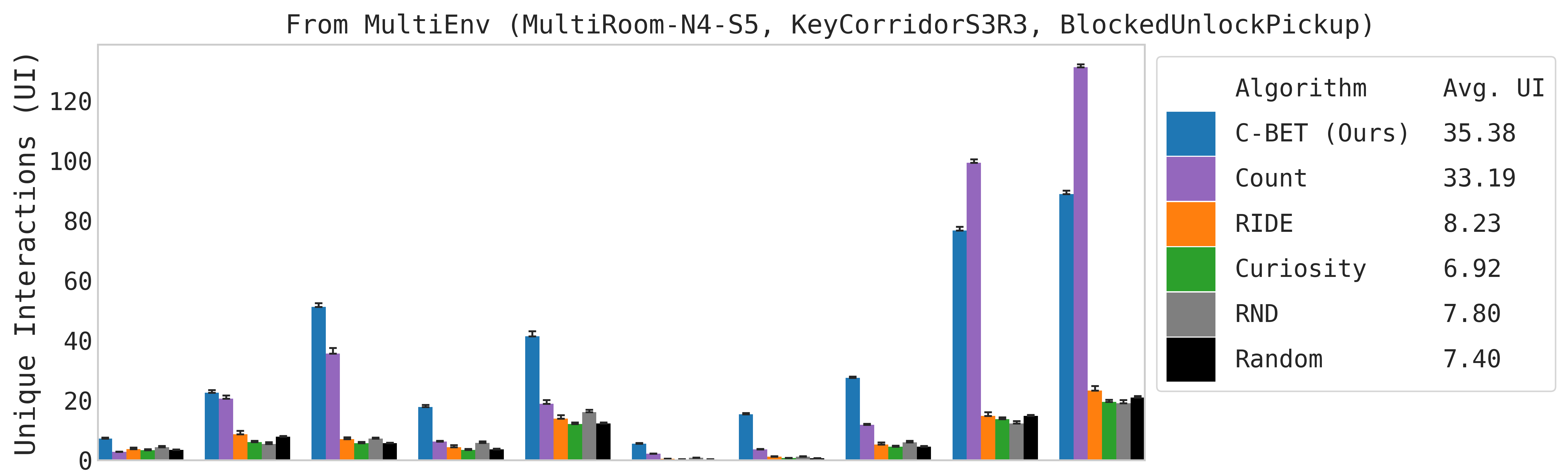}
\\[0em]
\includegraphics[width=0.79\linewidth]{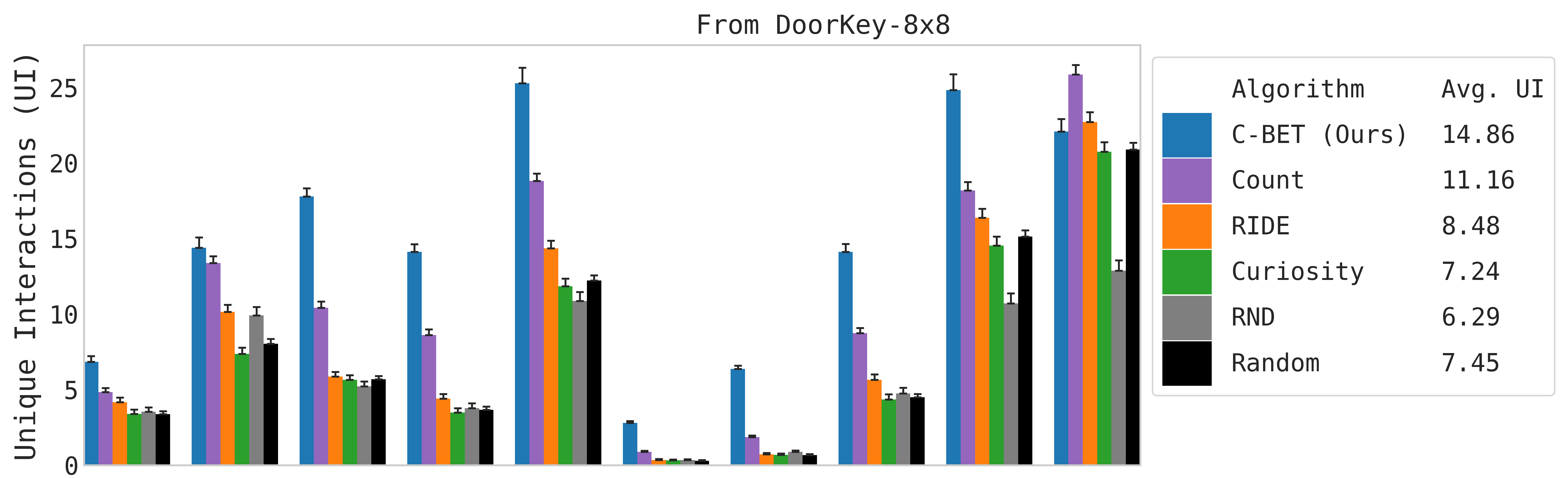}
\\[0em]
\includegraphics[width=0.78\linewidth]{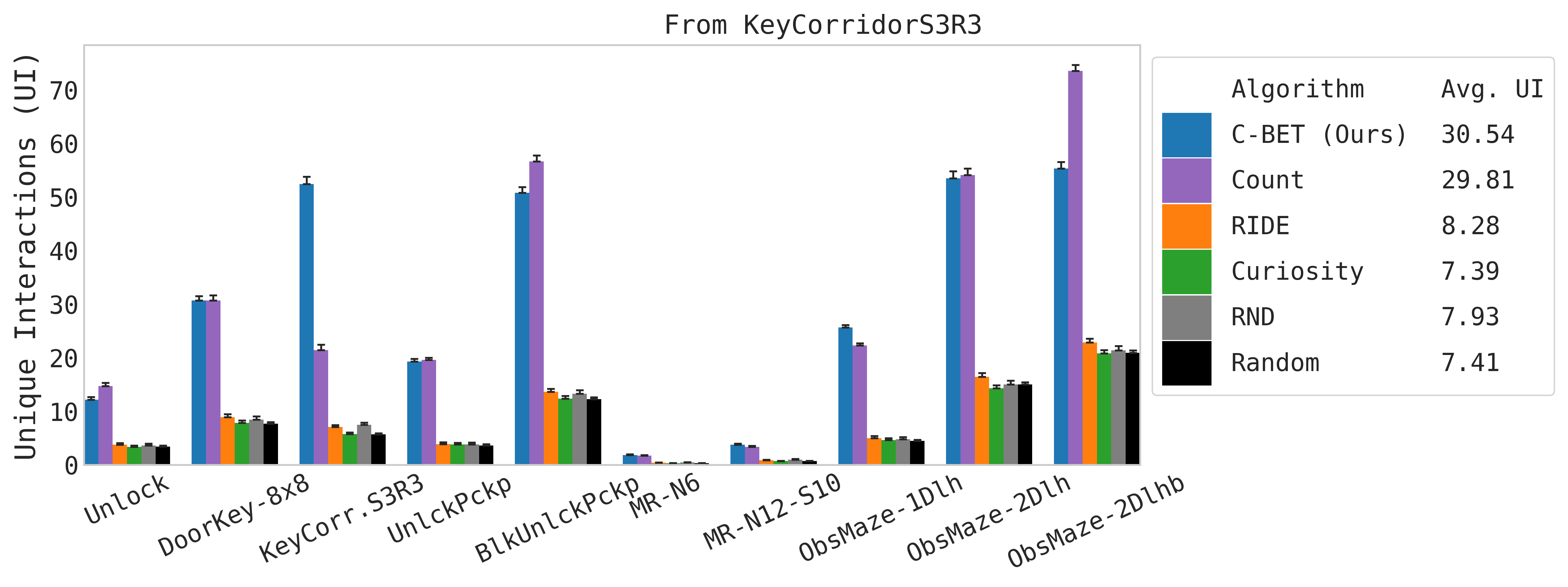}
\vspace{-8pt}
\caption{\label{fig:supp_pretrain_interactions} C-BET also outperforms baselines when we compare unique interactions at offline transfer. Not only it interacts the most as shown by the recap table, but also interacts in all environments. Clearly, it interacts more in environment with many keys/balls/boxes to pick (KeyCorridor, BlockedUnblockPickup, ObstructedMazes), and less if there is nothing to pick (MultiRooms).}
\end{center}
\end{figure*}

\vspace*{-3pt}
\label{supp:mini_inter}
Figures~\ref{fig:mini_extra_inter} and~\ref{fig:supp_pretrain_interactions} show that C-BET intrinsic reward encourages to interact with the environment more than all baselines. The more C-BET explores at pre-training, the more it interacts with it producing \textbf{unique} changes. At transfer, C-BET behavior transfers well to new environments, even the ones unknown dynamics (e.g., boxes in ObstructedMazes needs to be toggled to reveal keys). Of the baselines, only Count performs similarly to C-BET, but it does not generalize as well as C-BET. Most of its interactions, indeed, are in ObstructedMazes after training in MultiEnv.


\clearpage

\subsection{Extrinsic Return at Pre-Training and Success at Offline Transfer} 
\label{supp:mini_success}
\begin{figure*}[h]
\begin{center}
\includegraphics[width=0.99\linewidth]{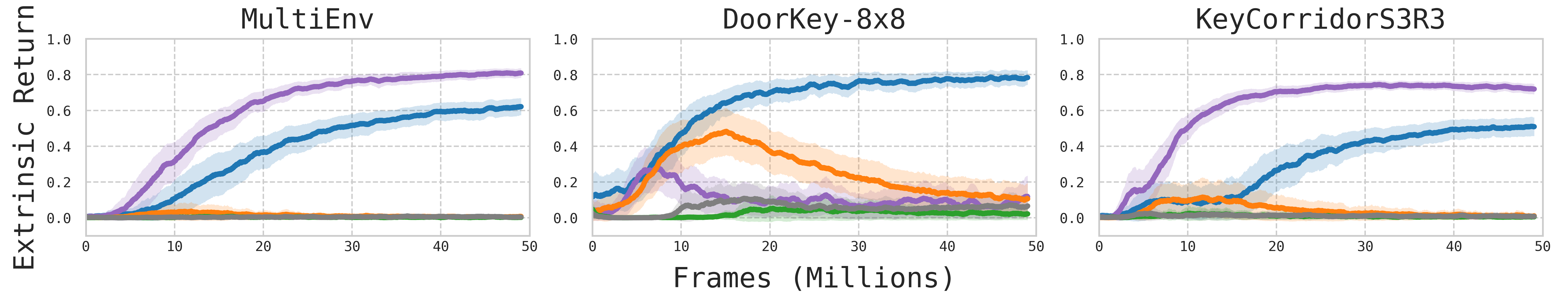}
\vspace{-7pt}
\caption{\label{fig:mini_extra_return}Trend of extrinsic return at pre-training. Only C-BET shows increasing return in all setups, showing that its better exploration brings it often to goal states.}
\vspace*{-5pt}
\end{center}
\end{figure*}

\begin{figure*}[h]
\begin{center}
\vspace*{-10pt}
\begin{tabular}{@{\extracolsep{\fill}}|>{\columncolor{colorcbet!70}}c>{\columncolor{colorcount!70}}c>{\columncolor{colorride!70}}c>{\columncolor{colorcuriosity!70}}c>{\columncolor{colorrnd!70}}c>{\columncolor{white}}c|}
    \hline
    {C-BET} & Count & RIDE & Curiosity & RND & Random \\
    \hline
    \textcolor{black}{\textbf{33.64\%}} & 27.21\% & \textcolor{black}{{1.8\%}} & 0.83\% & 2.52\% & 0.9\% \\
    \hline
\end{tabular}
\\[3pt]
\includegraphics[width=0.9\linewidth]{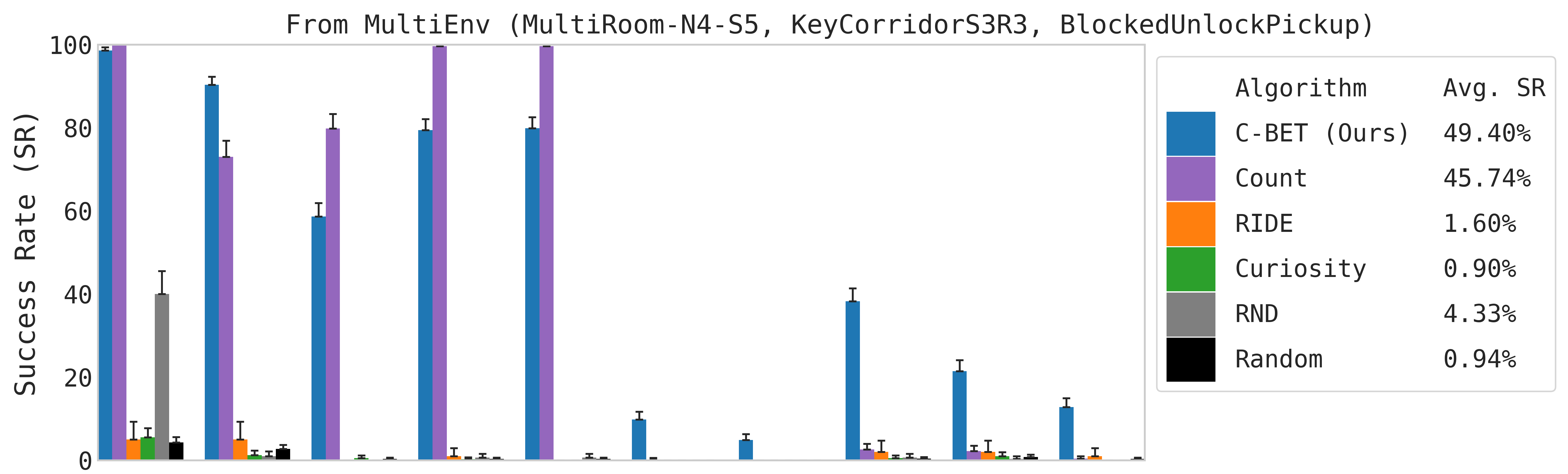}
\\[0em]
\includegraphics[width=0.9\linewidth]{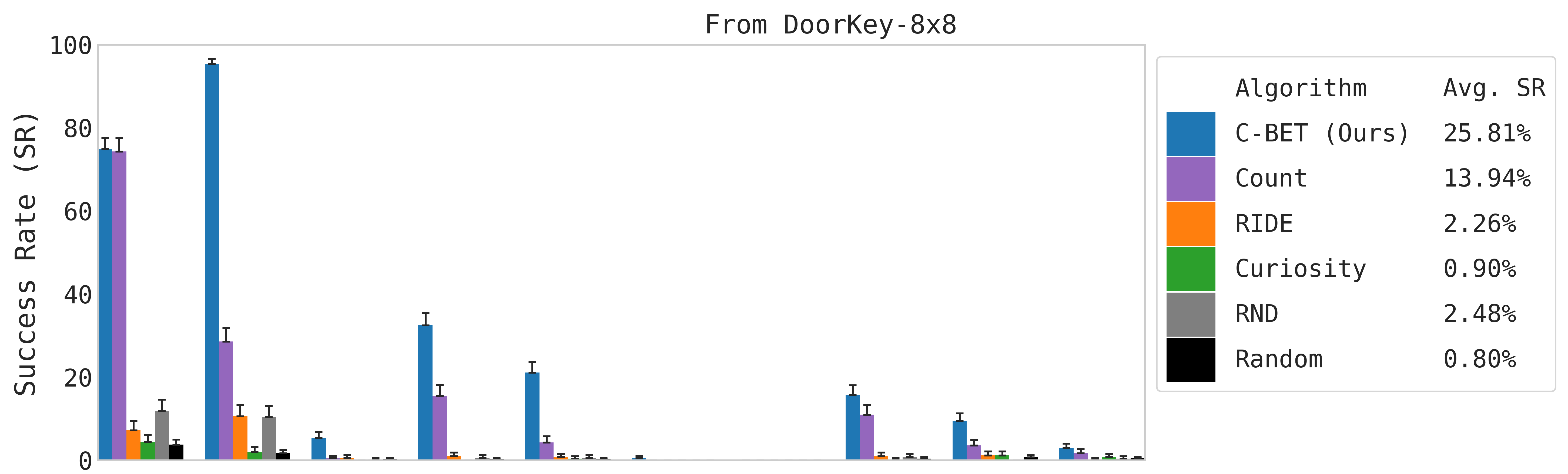}
\\[0em]
\includegraphics[width=0.9\linewidth]{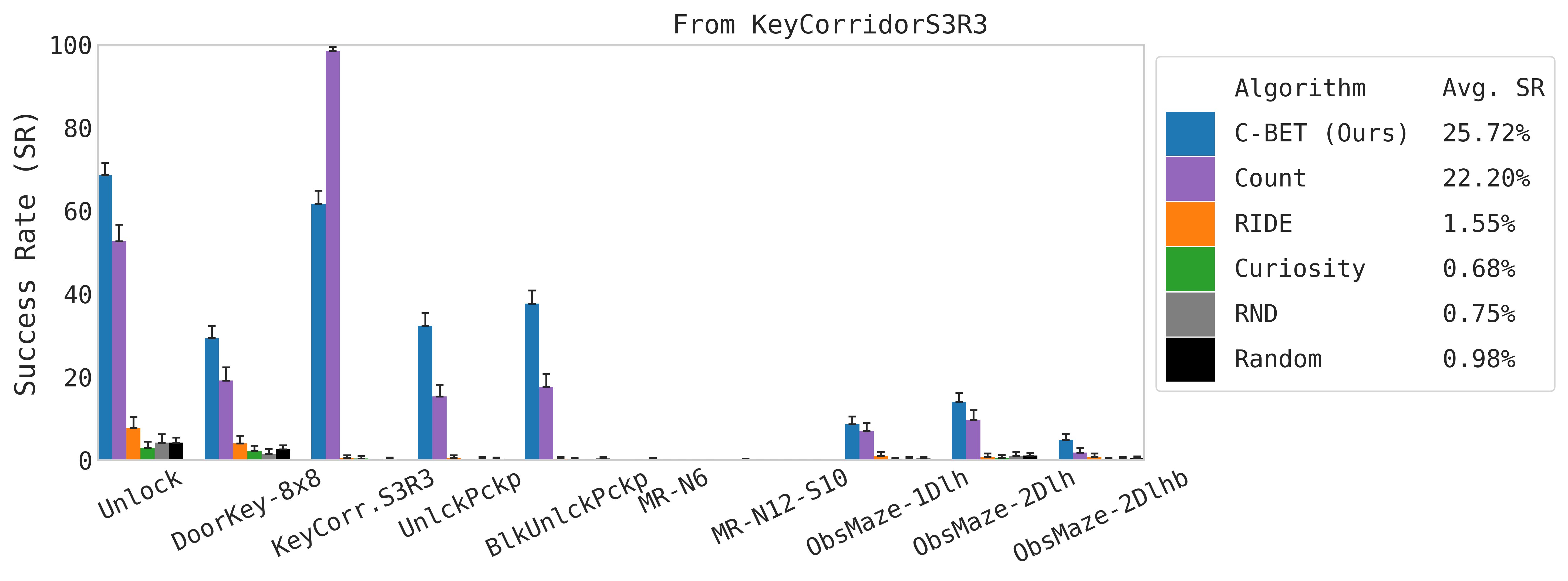}
\vspace{-7pt}
\caption{\label{fig:supp_pretrain_success}Success rate of pre-trained policies at the beginning of transfer.
Not only C-BET achieves the highest rate, but also generalizes to most environments, especially when pre-trained on MultiEnv.}
\end{center}
\end{figure*}

Already at pre-training, C-BET finds goal states thanks to its better exploration, as shown by the increasing trend of the extrinsic return\footnote{We recall that at pre-training agents do not get extrinsic rewards. We only record them as proxy for success.} (Figure~\ref{fig:mini_extra_return}). 
Similarly, at the beginning of transfer --i.e., without further training with extrinsic rewards-- C-BET already succeeds in many environments, especially when pre-trainined on MultiEnv (Figure~\ref{fig:supp_pretrain_success}). Among the baselines, only Count achieves success in some environments, but it clearly overfits to the pre-training environments and does not generalize nearly as well as C-BET. 


\clearpage

\subsection{The Problem of Vanishing Intrinsic Rewards at Pre-Training}
\label{supp:mini_decay}
\begin{figure*}[h]
\begin{center}
\includegraphics[width=0.95\linewidth]{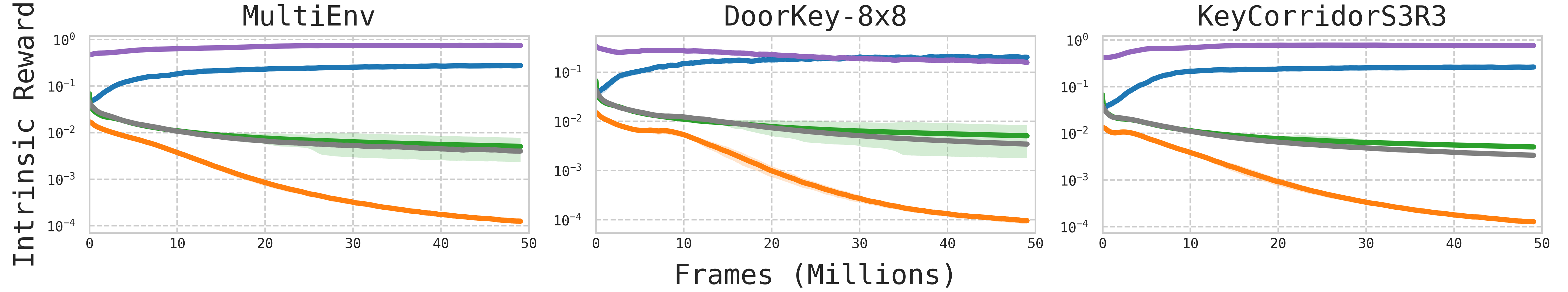}
\vspace{-7pt}
\caption{\label{fig:mini_extra_intrinsic}Log-scale trend of intrinsic rewards at pre-training. Baselines rewards decay over time, preventing learning. The problem is prominent in model-based rewards (RIDE, Curiosity, RND). On the contrary, C-BET rewards increase thanks to count resets.}
\vspace*{-5pt}
\end{center}
\end{figure*}

In the previous two sections, we have shown that C-BET outperforms baselines. Its agent interacts more with the environment while looking for rare and unique changes, and the resulting behavior allows it to discover extrinsic rewards more often.
But why do baselines --especially RIDE, Curiosity and RND-- perform poorly? 
As discussed in Section~\ref{subsec:pretrain}, classic intrinsic rewards decrease over time as the agent explores, to the point that they vanish to zero given enough samples, preventing further learning.
On the contrary, thanks to count resets C-BET reward never decays, and allows the agent to always get meaningful feedback.
\\
This is shown in Figure~\ref{fig:mini_extra_intrinsic}, where intrinsic rewards based on model errors (RIDE, Curiosity, RND) quickly decay as models are learned with ease. 
Even Count, that does not use any model, suffer from vanishing rewards, albeit to a lesser extent and mostly when trained in environments with similar states like DoorKey.
On the contrary, C-BET intrinsic reward increases over time the more agent learns to interact with the environment.
But does C-BET outperform Count only thanks to count resets? As we have shown in Appendix~\ref{supp:abl_resets}, even with resets Count does not generalize well to unseen environments, and overfits to the training environments. C-BET, instead, achieves success even in unseen environments.

\section{Noisy Environment Supplemental Results}
\label{supp:noisy}

\begin{wrapfigure}{l}{0.32\textwidth}
\vspace{-1.5em}
\begin{center}
\includegraphics[width=0.98\linewidth]{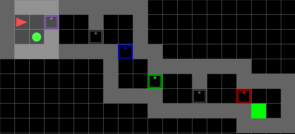}
\end{center}
\vspace{-0.7em}
\caption{\label{fig:noisytv}{Random instance of the MultiRoomNoisyTV-N7-S4 environment used in our experiments. It has seven rooms of maximum size four.}}
\vspace{-1.em}
\end{wrapfigure}
So far, we conduced experiments in environments with deterministic dynamics. It is known, that stochastic dynamics can be problematic for intrinsic rewards based on model prediction errors~\citep{pathak2017curiosity}, as the agents it attracted to sources of noise.
This may affect C-BET as well, especially if the noise influences the environment (e.g., the change) rather than agent (e.g., the state).
In this section, we test C-BET ability to deal with stochasticity. In particular, we want to see if either its change-based reward or its count resets may negatively affect learning. 
Therefore, for this evaluation we compare C-BET against the Count baseline --i.e., state-count reward only-- with and without count resets on two versions of the same environments (with and without noise)\footnote{We also tested RIDE, RND and Curiosity, but they showed the same poor performance of previous experiments. Despite the noise, their model error still decays quickly.}. 
The environment is MultiRoomNoisyTV, developed by~\citet{raileanu2020ride} and depicted in Figure~\ref{fig:noisytv}. 
In this environment, a ball is placed in the first room, where the agent spawns at the start of an episode. Anytime the agent does the `drop' action the ball randomly changes color, regardless of the agent position. Like any MultiRoom, there is nothing that can be picked, and the ball itself cannot be picked.

Table~\ref{tab:supp_noisy_success} shows that both C-BET and Count perform worse without resets. In particular, Table~\ref{tab:supp_noisy_actions} shows that their learned policies are essentially random. This further confirms what we showed in Appendix~\ref{supp:abl_resets}: without resets, intrinsic rewards vanish and any policy --even a random one-- is optimal. This problem is prominent in MultiRooms due to their scarce state diversity, as there are only empty rooms with already unlocked doors.


\begin{table*}[h]
\begin{center}
\caption{\label{tab:supp_noisy_success}\textbf{Policy success rate} averaged over all ten transfer environments in different pre-training setups. In both cases, not resetting counts perform worse. However, C-BET is negatively affected by the NoisyTV, albeit to a limited extend, while Count benefits from it.}
\vspace*{-3pt}
\begin{tabular}{@{\extracolsep{\fill}}|l|>{\columncolor{colorcbet7!70}}c>{\columncolor{colorcbet7!70}}c>{\columncolor{colorcbet4!70}}c>{\columncolor{colorcbet4!70}}c|}
\hline
 & {C-BET (Resets)} & {C-BET (No Resets)} & {Count (Resets)} & {Count (No Resets)} 
\\
\hline
Default & \textcolor{black}{{11.37\%}} & \textcolor{black}{{1.68\%}} & 9.72\% & 5.92\%\\
\hline
NoisyTV & \textcolor{black}{{9.76\%}} & \textcolor{black}{{4.0\%}} & \textbf{17.90}\% & {4.04}\%\\
\hline
\end{tabular}
\end{center}
\end{table*}

Unsurprisingly, when the noisy ball is added C-BET performance slightly decreases. However, Count's almost doubles.
To explain these behaviors, we need to look at the policy distributions after pre-training, shown in Table~\ref{tab:supp_noisy_success}. Let's consider distributions after training without noise first. In this case, both policies assign high probability to `toggle'. `Toggle', indeed, is the only interesting action to do because there is nothing to pick and all doors are already unlocked (and `toggle' opens/closes doors). The only significant difference is that C-BET moves more and turns less. 
The reason is that C-BET's panoramic changes give little reward to turns, as discussed in Appendix~\ref{supp:abl_pano_state} and~\ref{supp:abl_change_only}. Overall, when trained without noise C-BET transfer slightly better than Count, as shown in Table~\ref{tab:supp_noisy_success}.

Let's now consider distributions after training with noise. As expected, both policies assign much higher probability to `drop', the action randomly changing the ball color. Yet, other actions probabilities are significantly different between C-BET and Count.
\\
Indeed, C-BET assigns slightly lower probability to all other actions, but overall its distribution does not change much. This explains why its success rate in Table~\ref{tab:supp_noisy_success} is similar with/without noise.
\\
On the contrary, Count's `toggle' probability drastically decreases (from 24.7\% to 6.8\%) but `forward' increases.
Therefore, Count's policy toggles less and moves more. Thanks to this, the resulting policy explores better, despite the higher `drop' probability.

Finally, we can summarize the findings of this investigation as follows.
\begin{itemize}[nosep,leftmargin=*,before=\vspace{-4pt}]
\item Count resets are important for learning exploration policies, especially if the training environment lacks diversity.
\item A policy `toggling too much' performs poorly.
\item In the presence of noise, C-BET performs worse, albeit to a limited extent.
\item The presence of noise actually helps the Count baseline over `overfit less' to some actions (in this case, the policy learns not to `toggle too much').
\end{itemize}


\begin{table*}[h]
\begin{center}
\setlength{\tabcolsep}{5pt}
\vspace*{5pt}
\caption{\label{tab:supp_noisy_actions}\textbf{Policy distributions} after pre-training in different setups. Without resets (\textcolor{colorcbet2!70}{gold} rows) both C-BET and Count policies are essentially random regardless of the noise, as shown by their entropy. With resets and no noise, both C-BET and Count assign low probability to `drop' and high to `toggle' (highlighted in \textcolor{red!70}{red}), as there are only unlocked doors. With resets and noise, both assign much higher probability to `drop', the action triggering the noisy ball to change color.}
\textbf{From Default MultiRoom}
\\[2pt]
\begin{tabular}{@{\extracolsep{\fill}}|l|>{\columncolor{colora0!0}}c>{\columncolor{colora1!0}}c>{\columncolor{colora2!0}}c>{\columncolor{colora3!0}}c>{\columncolor{colora4!0}}c>{\columncolor{colora5!0}}c>{\columncolor{colora6!0}}c||c|}
    \hline
     & Left & Right & Forward & Pick & \textcolor{red}{\textbf{Drop}} & \textcolor{red}{\textbf{Toggle}} & Done & Entropy \\
    \hline
    C-BET (Resets) & {4.0\%} & 4.3\% & \textcolor{black}{{48.6\%}} & 5.3\% & \textcolor{red}{{\textbf{6.0}\%}} & \textcolor{red}{\textbf{26.1}}\% & 5.6\% & 0.75
    \\
    \rowcolor{colorcbet2!50} C-BET (No Resets) & {13.6\%} & 13.6\% & \textcolor{black}{{16.0\%}} & 14.1\% & \textcolor{black}{{14.1\%}} & 14.4\% & 14.0\% & 0.99
    \\
    Count (Resets) & {12.4\%} & 9.2\% & \textcolor{black}{{33.2\%}} & 5.2\% & \textcolor{red}{{\textbf{5.6}\%}} & \textcolor{red}{\textbf{29.0}}\% & 5.0\% & 0.86
    \\
    \rowcolor{colorcbet2!50} Count (No Resets) & {14.3\%} & 14.8\% & \textcolor{black}{{23.7\%}} & 12.7\% & \textcolor{black}{{10.6\%}} & 12.6\% & 11.1\% & 0.98
    \\
    \hline
\end{tabular}
\\[1em]
\textbf{From MultiRoom with NoisyTV}
\\[1.7pt]
\begin{tabular}{@{\extracolsep{\fill}}|l|>{\columncolor{colora0!0}}c>{\columncolor{colora1!0}}c>{\columncolor{colora2!0}}c>{\columncolor{colora3!0}}c>{\columncolor{colora4!0}}c>{\columncolor{colora5!0}}c>{\columncolor{colora6!0}}c||c|}
    \hline
     & Left & Right & Forward & Pick & \textcolor{red}{\textbf{Drop}} & \textcolor{red}{\textbf{Toggle}} & Done & Entropy \\
    \hline
    C-BET (Resets) & {2.9\%} & 3.0\% & \textcolor{black}{{48.4\%}} & 5.3\% & \textcolor{red}{{\textbf{12.8}\%}} & \textcolor{red}{\textbf{24.7}}\% & 2.9\% & 0.74
    \\
    \rowcolor{colorcbet2!50} C-BET (No Resets) & {11.5\%} & 11.3\% & \textcolor{black}{{20.3\%}} & 13.0\% & \textcolor{black}{{15.8\%}} & 15.7\% & 12.3\% & 0.99
    \\
    Count (Resets) & {10.7\%} & 11.6\% & \textcolor{black}{{42.3\%}} & 4.6\% & \textcolor{red}{{\textbf{22.2}\%}} & \textcolor{red}{\textbf{6.8}}\% & 2.1\% & 0.82
    \\
    \rowcolor{colorcbet2!50} Count (No Resets) & {10.9\%} & 12.1\% & \textcolor{black}{{16.5\%}} & 9.3\% & \textcolor{black}{{32.1\%}} & 11.3\% & 7.6\% & 0.94
    \\
    \hline
\end{tabular}
\end{center}
\end{table*}

\clearpage

\section{Habitat Supplemental Results}

\subsection{C-BET With Egocentric Change}
\label{supp:hab_cbet_ego}
\begin{figure*}[h]
\centering
\begin{subfigure}[t]{\linewidth}
\centering
\includegraphics[width=\linewidth]{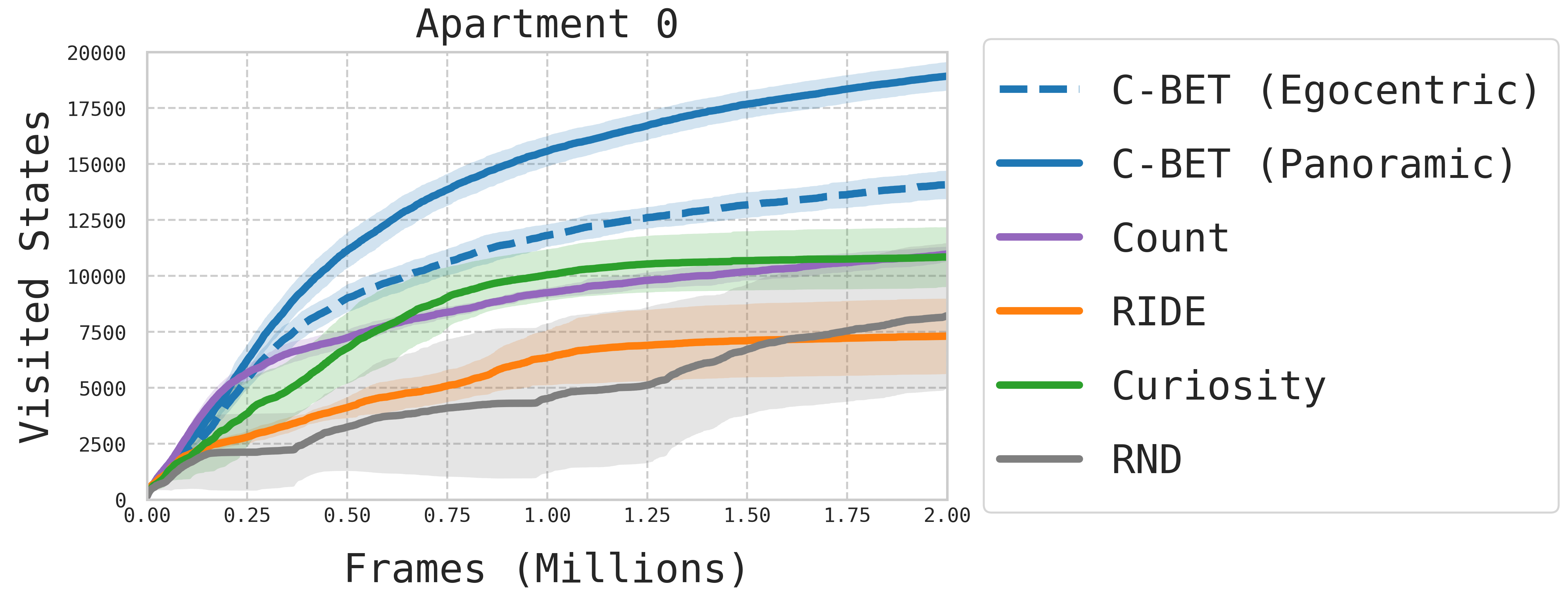}
\vspace*{-1.1em}
\caption{Visited states throughout pre-training.}
\end{subfigure}
\\[1.2em]
\begin{subfigure}[t]{\linewidth}
\centering
\textbf{\small \hspace*{0.5em} C-BET (Ego) \hspace*{1em} C-BET (Pano) \hspace*{2.5em} RIDE \hfill Count \hfill Curiosity \hfill RND \hspace*{2.5em}}
\\[0.em]
\includegraphics[width=0.16\linewidth]{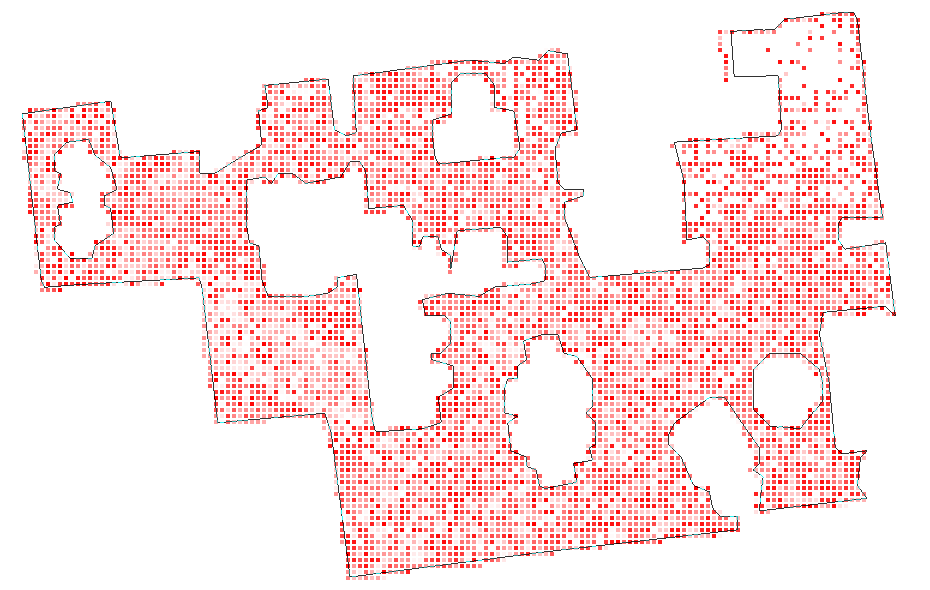}
\hfill
\includegraphics[width=0.16\linewidth]{heats/heat_pre_cbet.png}
\hfill
\includegraphics[width=0.16\linewidth]{heats/heat_pre_ride.png}
\hfill
\includegraphics[width=0.16\linewidth]{heats/heat_pre_count.png}
\hfill
\includegraphics[width=0.16\linewidth]{heats/heat_pre_curiosity.png}
\hfill
\includegraphics[width=0.16\linewidth]{heats/heat_pre_rnd.png}
\vspace*{-0.8em}
\caption{Scene coverage after pre-training.}
\end{subfigure}
\\[1.2em]
\begin{subfigure}[t]{\linewidth}
    \centering
    \includegraphics[width=\linewidth]{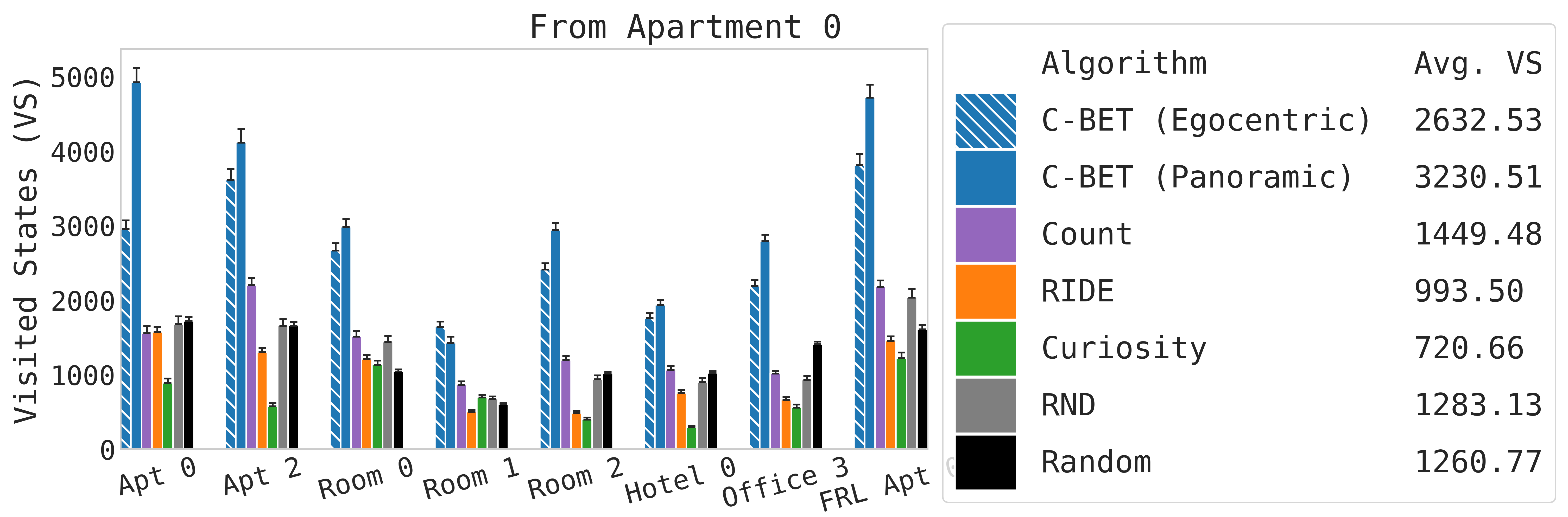}
    \vspace{-1.5em}
    \caption{Visited states at offline transfer.}
\end{subfigure}
\caption{\label{fig:habitat_pretrain_supp}\textbf{Habitat pre-training.} Even with egocentric views, C-BET clearly outperforms all baselines. Not only it still visits almost all Apartment 0 during pre-training, but its policy also transfers well to all unseen scenes.}
\end{figure*}

Here, we report C-BET results when environment changes are encoded with egocentric views rather than panoramic views.
\\
Figure~\ref{fig:habitat_pretrain_supp} shows that egocentric views do not perform as well as panoramic views. Yet, C-BET still outperforms all baselines. Indeed, it visits more of Apartment 0 (at pre-training) and of all scenes (at transfer) than all baselines.
This is a further confirmation of what stated in Appendix~\ref{supp:abl_ego_vs_pano}: \textit{C-BET does not perform better because of panoramic change counts, but because of how it combines them with egocentric state counts.} Taking into account \textbf{both} agent-centric and environment-centric components is the key for better exploration.
Nonetheless, these results also suggest that how we encode \textit{what changes in the environment} as opposed to the agent's state can further improve exploration.

\clearpage

\subsection{Scene Coverage at Transfer on All Scenes}
\label{supp:scenes}

\begin{figure*}[h]
\centering
\textbf{\small \hspace*{2.9em} C-BET \hfill RIDE \hfill Count \hfill Curiosity \hfill RND \hfill Random \hspace*{1.5em}}
\\[0.em]
\raisebox{14pt}{\rotatebox[origin=t]{90}{\textbf{Apt 0}}}
\includegraphics[width=0.155\linewidth]{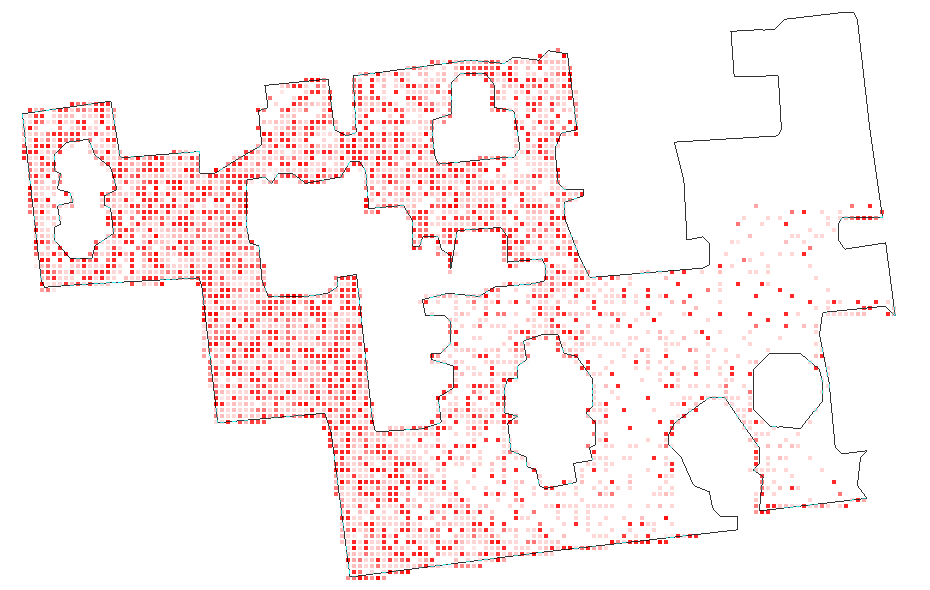}
\hfill
\includegraphics[width=0.155\linewidth]{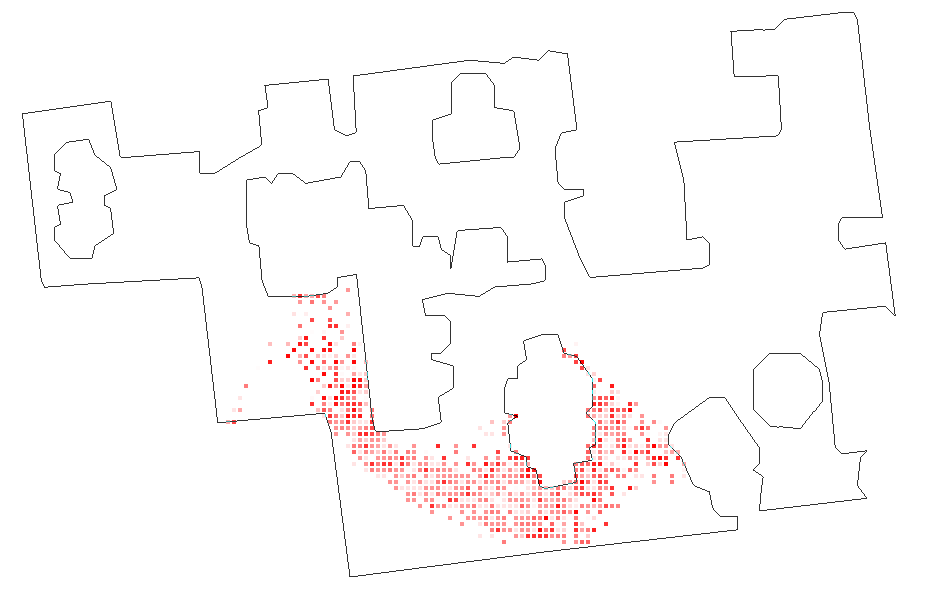}
\hfill
\includegraphics[width=0.155\linewidth]{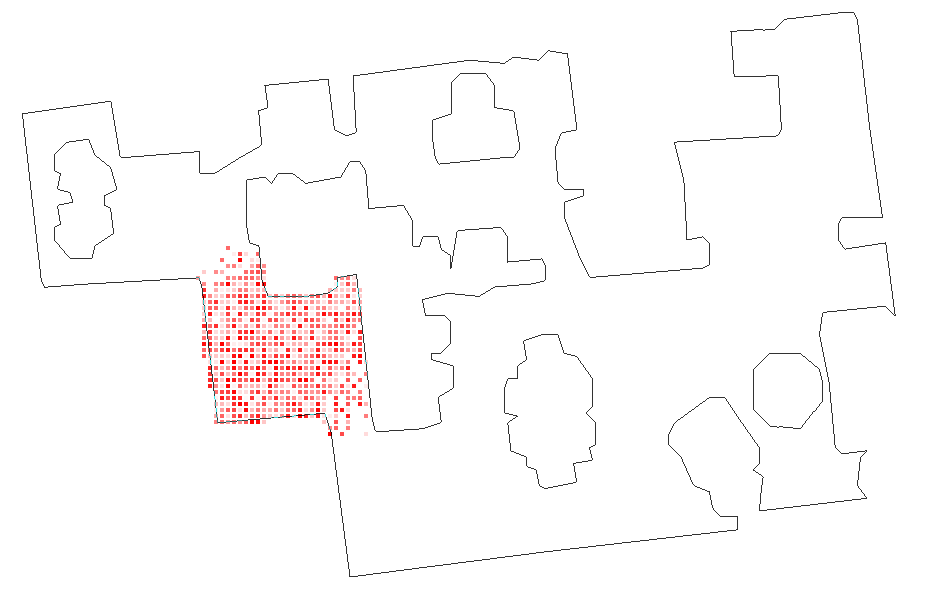}
\hfill
\includegraphics[width=0.155\linewidth]{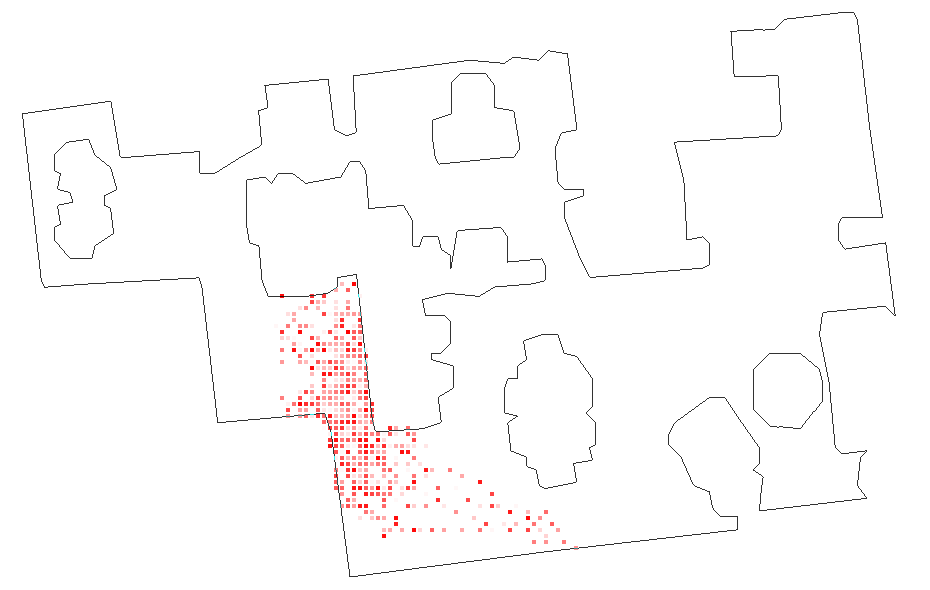}
\hfill
\includegraphics[width=0.155\linewidth]{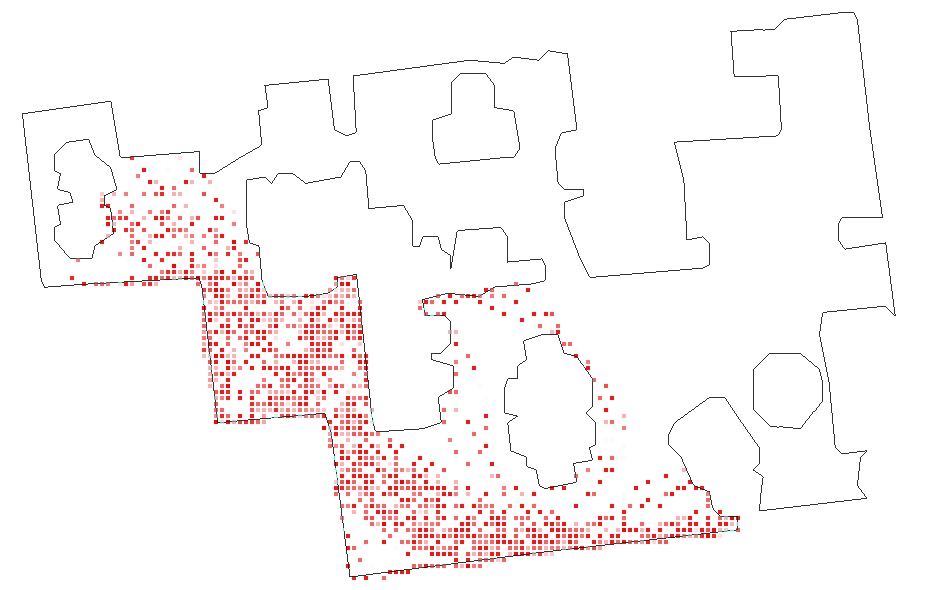}
\hfill
\includegraphics[width=0.155\linewidth]{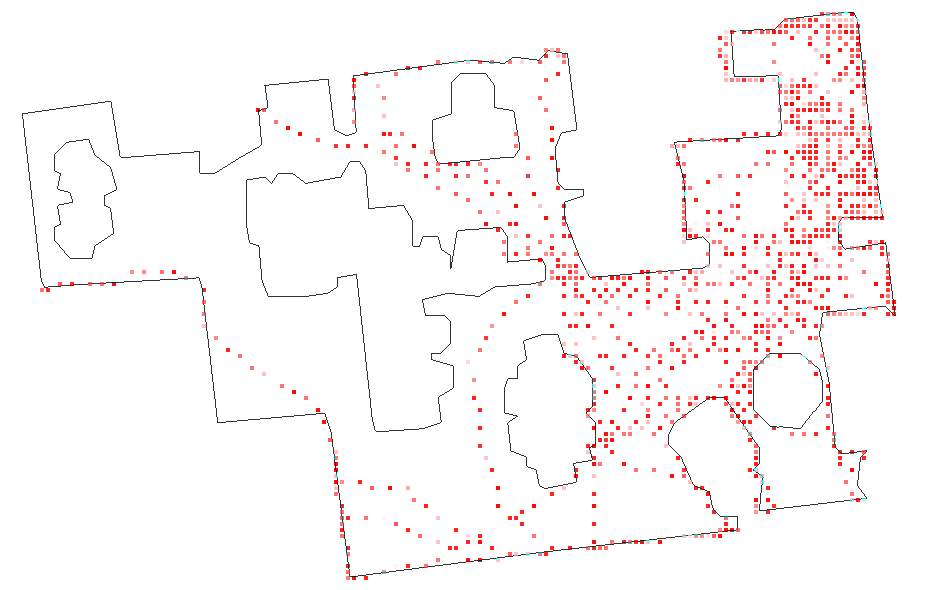}
\\[0.00em]
\raisebox{19pt}{\rotatebox[origin=t]{90}{\textbf{Apt 2}}}
\includegraphics[width=0.155\linewidth]{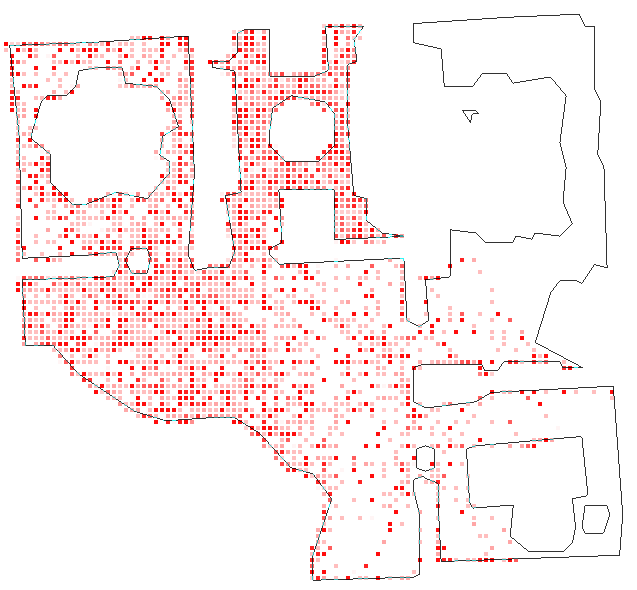}
\hfill
\includegraphics[width=0.155\linewidth]{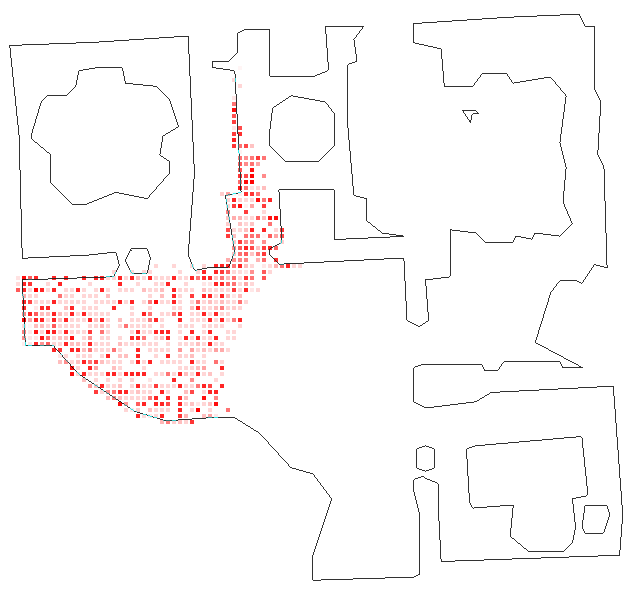}
\hfill
\includegraphics[width=0.155\linewidth]{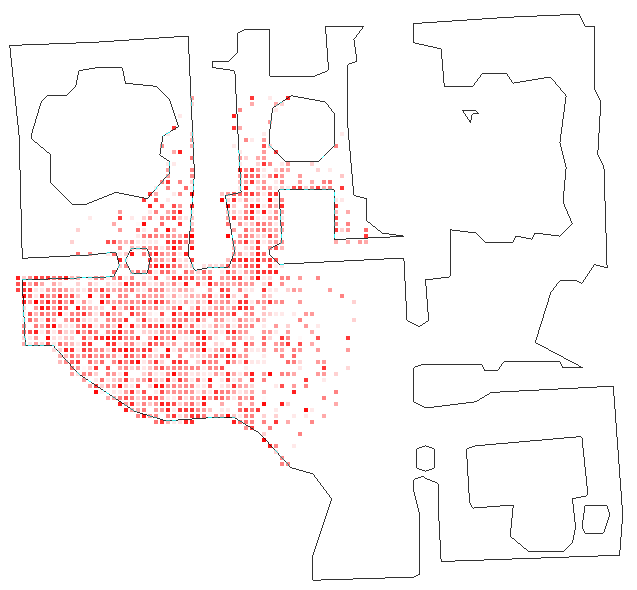}
\hfill
\includegraphics[width=0.155\linewidth]{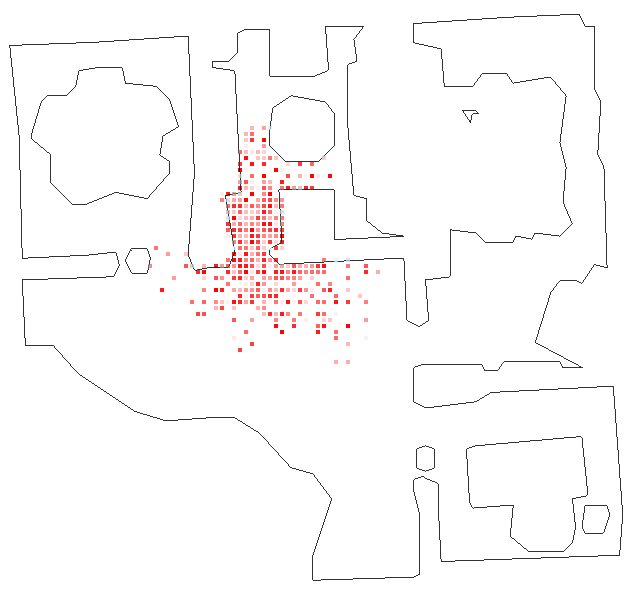}
\hfill
\includegraphics[width=0.155\linewidth]{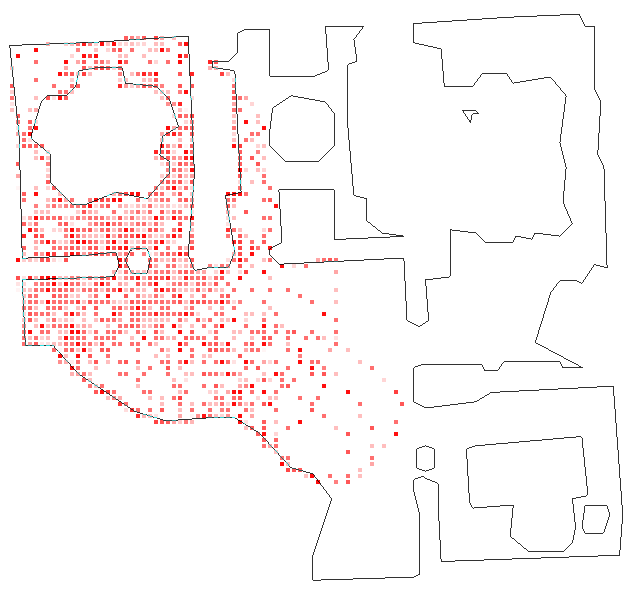}
\hfill
\includegraphics[width=0.155\linewidth]{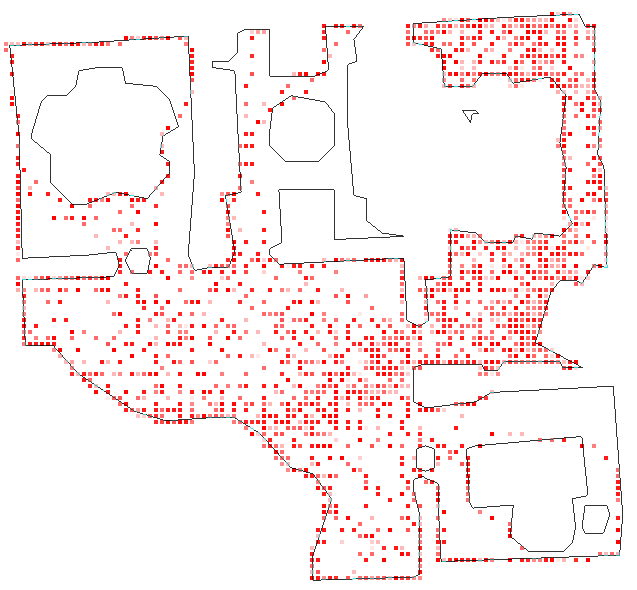}
\\[0.00em]
\raisebox{25pt}{\rotatebox[origin=t]{90}{\textbf{Room 0}}}
\includegraphics[width=0.155\linewidth,height=2.2cm]{heats/heat_test_cbet_room_0.png}
\hfill
\includegraphics[width=0.155\linewidth,height=2.2cm]{heats/heat_test_ride_room_0.png}
\hfill
\includegraphics[width=0.155\linewidth,height=2.2cm]{heats/heat_test_count_room_0.png}
\hfill
\includegraphics[width=0.155\linewidth,height=2.2cm]{heats/heat_test_curiosity_room_0.png}
\hfill
\includegraphics[width=0.155\linewidth,height=2.2cm]{heats/heat_test_rnd_room_0.png}
\hfill
\includegraphics[width=0.155\linewidth,height=2.2cm]{heats/heat_test_random_room_0.png}
\\[0.00em]
\raisebox{25pt}{\rotatebox[origin=t]{90}{\textbf{Room 1}}}
\includegraphics[width=0.155\linewidth,height=2.2cm]{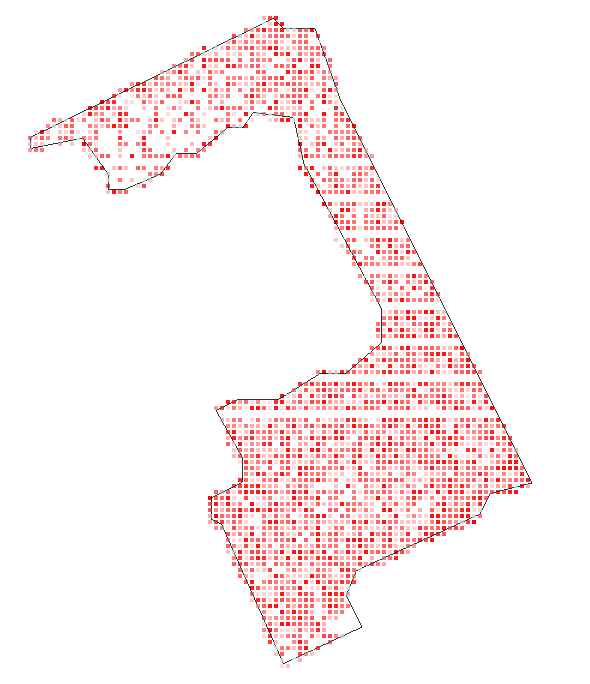}
\hfill
\includegraphics[width=0.155\linewidth,height=2.2cm]{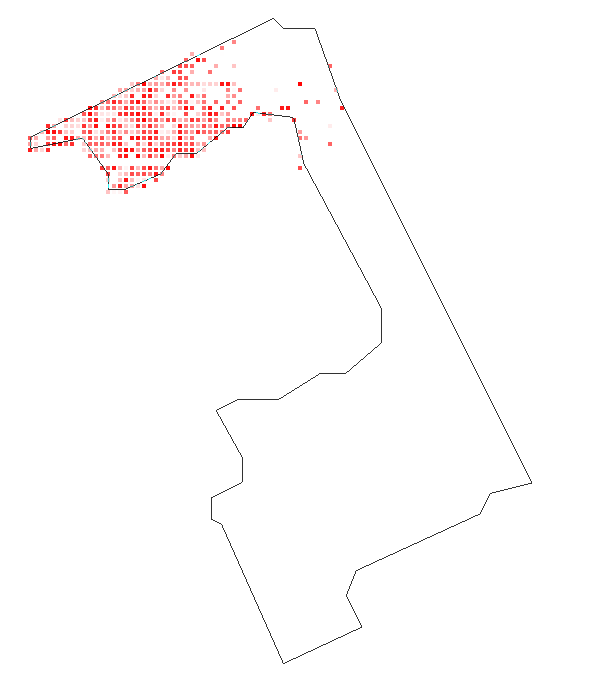}
\hfill
\includegraphics[width=0.155\linewidth,height=2.2cm]{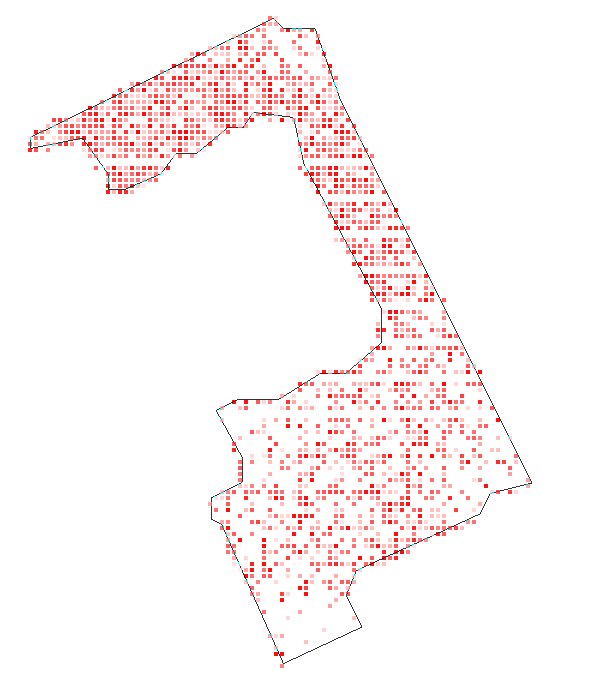}
\hfill
\includegraphics[width=0.155\linewidth,height=2.2cm]{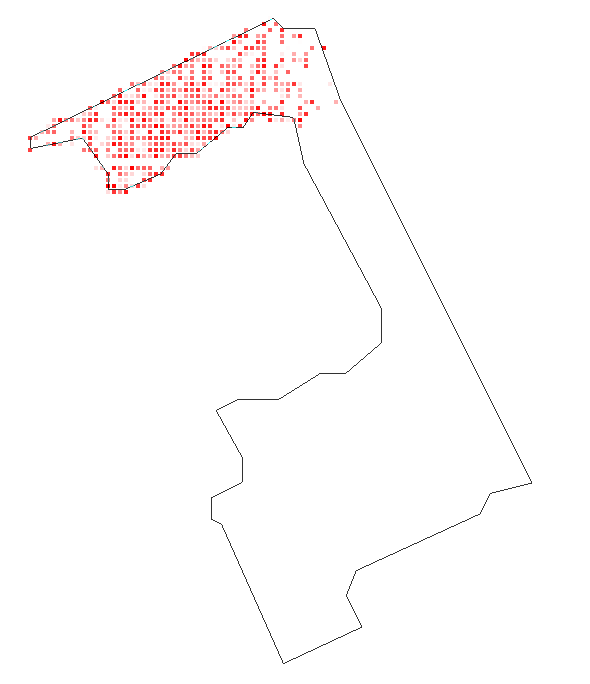}
\hfill
\includegraphics[width=0.155\linewidth,height=2.2cm]{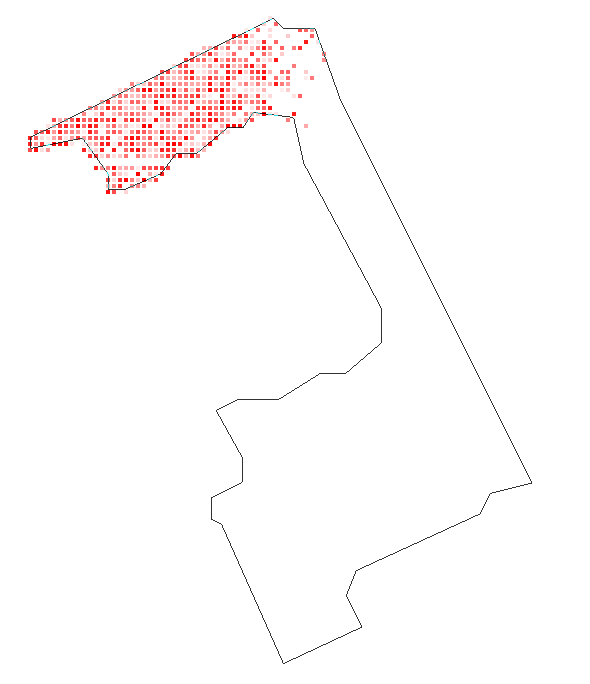}
\hfill
\includegraphics[width=0.155\linewidth,height=2.2cm]{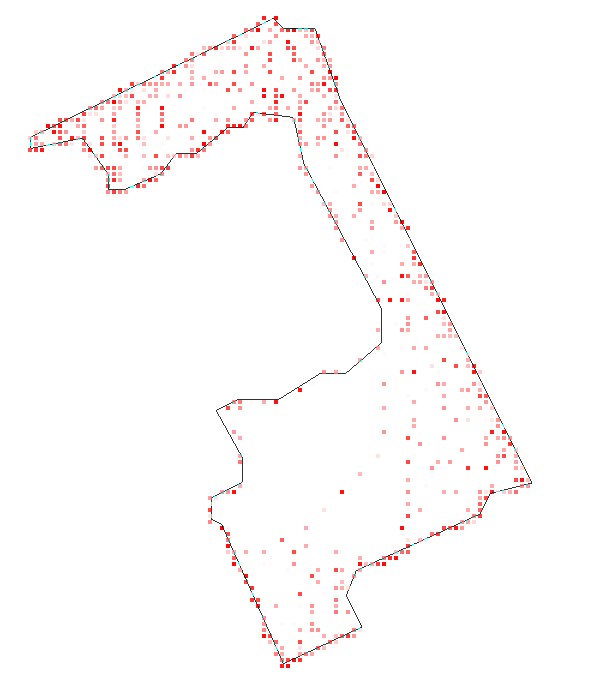}
\\[0.00em]
\raisebox{25pt}{\rotatebox[origin=t]{90}{\textbf{Room 2}}}
\includegraphics[width=0.155\linewidth,height=2.2cm]{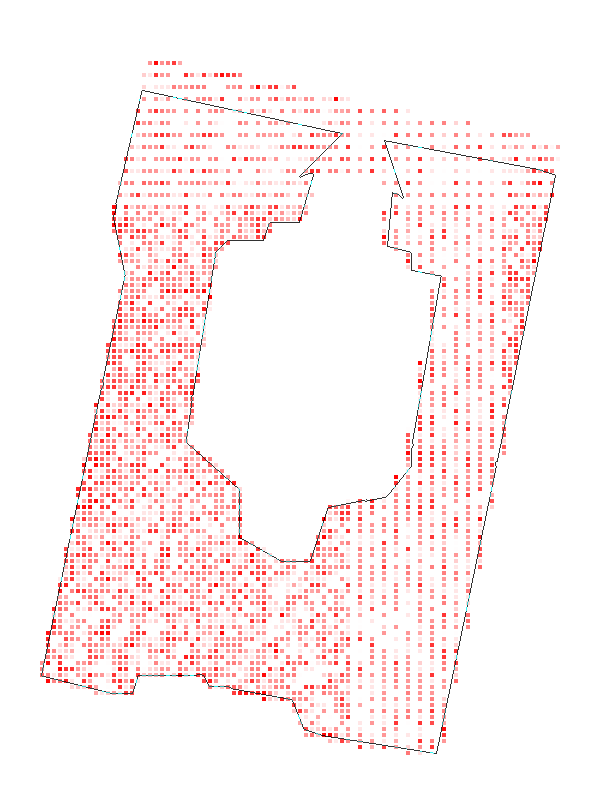}
\hfill
\includegraphics[width=0.155\linewidth,height=2.2cm]{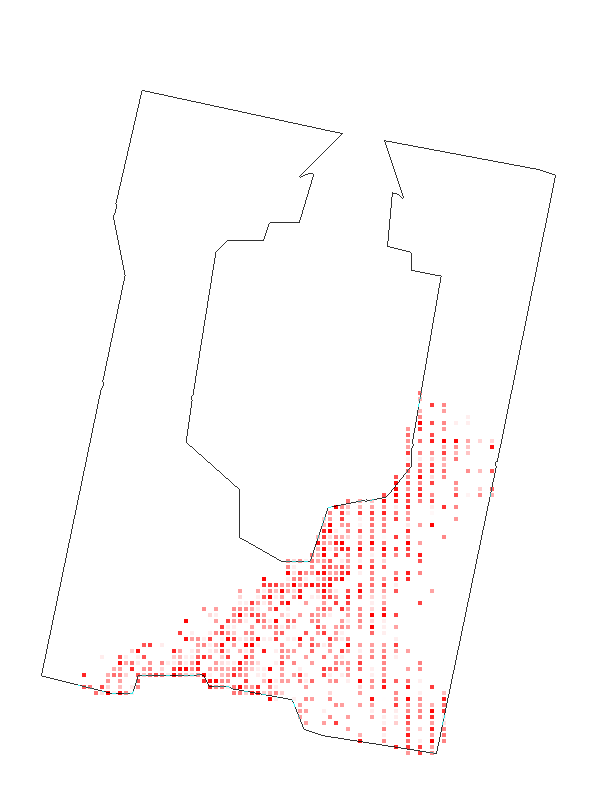}
\hfill
\includegraphics[width=0.155\linewidth,height=2.2cm]{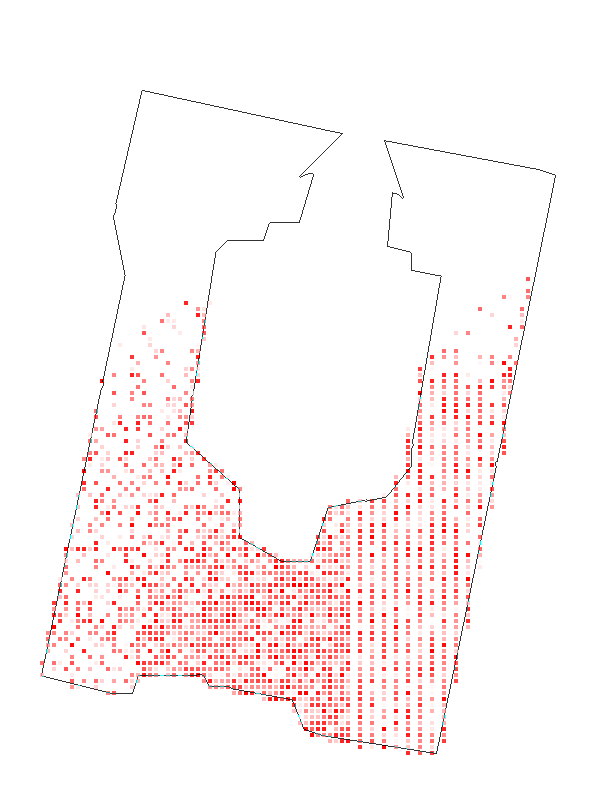}
\hfill
\includegraphics[width=0.155\linewidth,height=2.2cm]{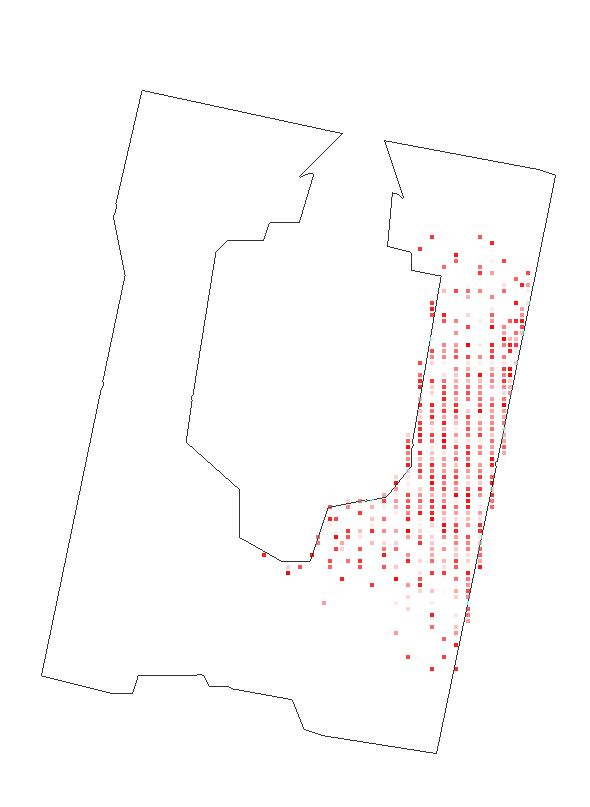}
\hfill
\includegraphics[width=0.155\linewidth,height=2.2cm]{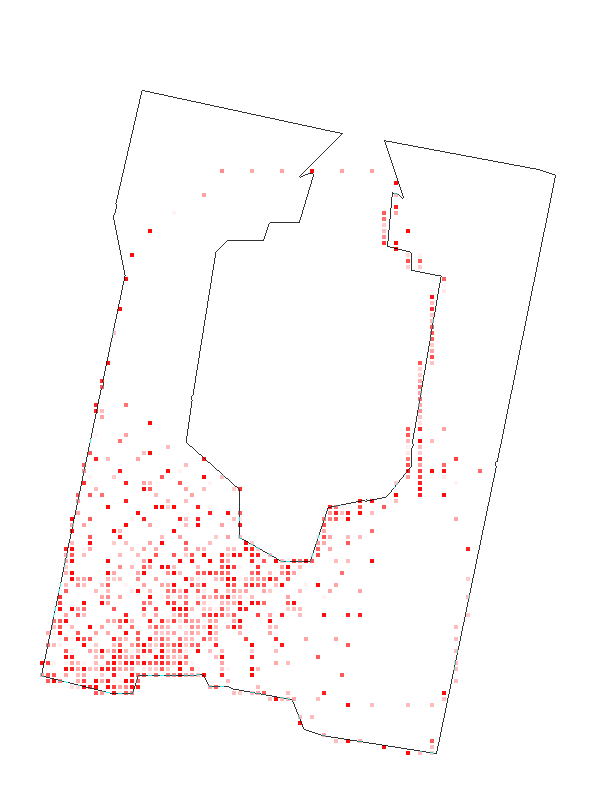}
\hfill
\includegraphics[width=0.155\linewidth,height=2.2cm]{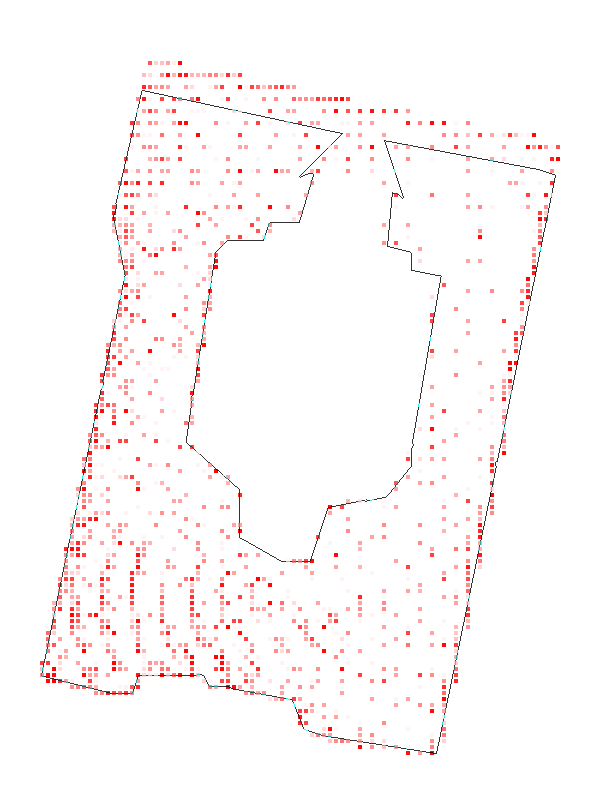}
\\[0.00em]
\raisebox{40pt}{\rotatebox[origin=t]{90}{\textbf{Hotel 0}}}
\includegraphics[width=0.155\linewidth,height=3.3cm]{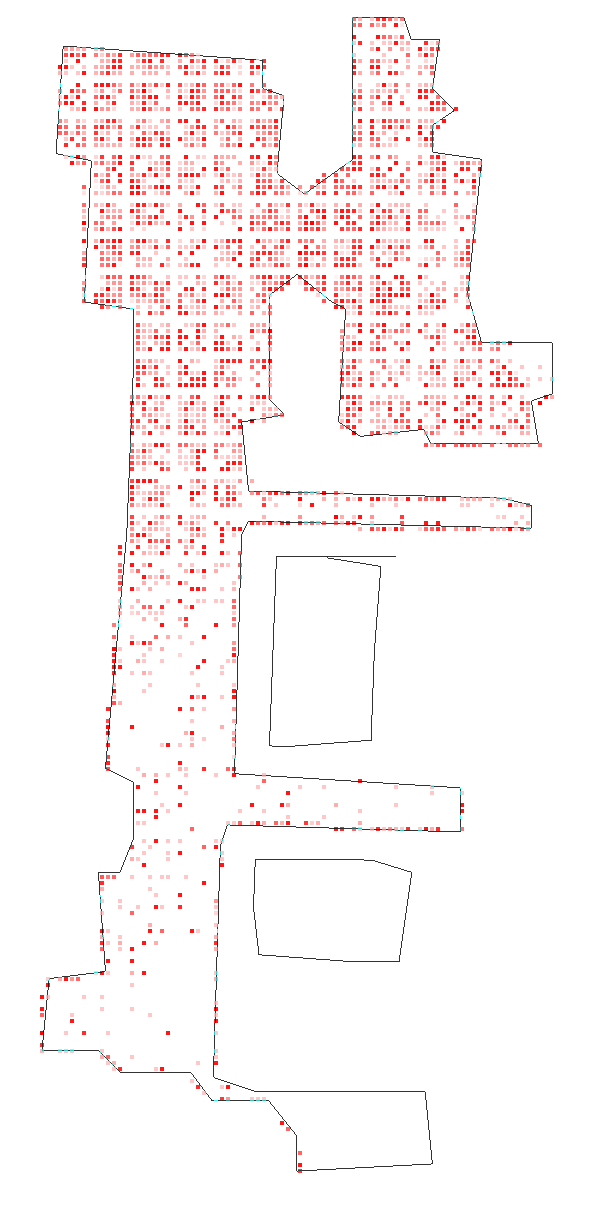}
\hfill
\includegraphics[width=0.155\linewidth,height=3.3cm]{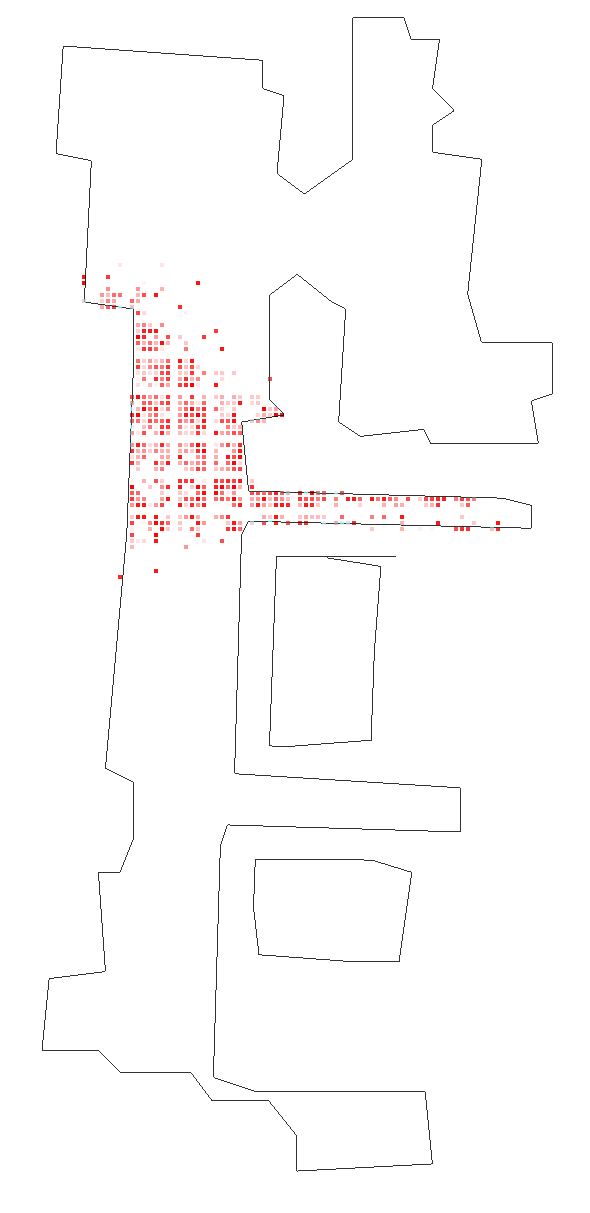}
\hfill
\includegraphics[width=0.155\linewidth,height=3.3cm]{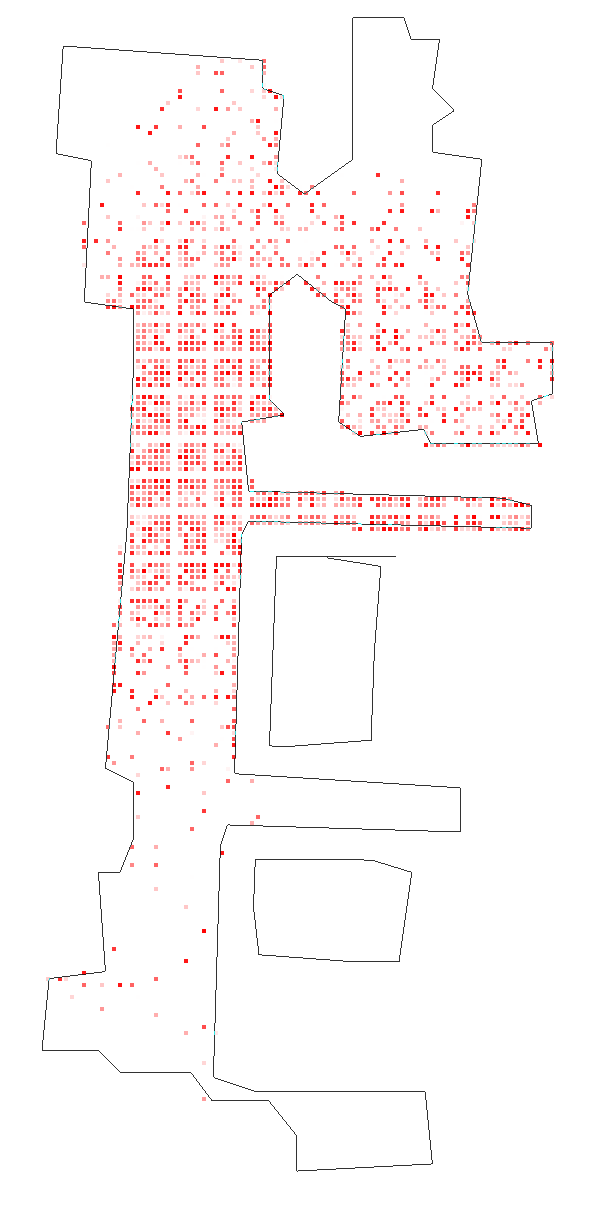}
\hfill
\includegraphics[width=0.155\linewidth,height=3.3cm]{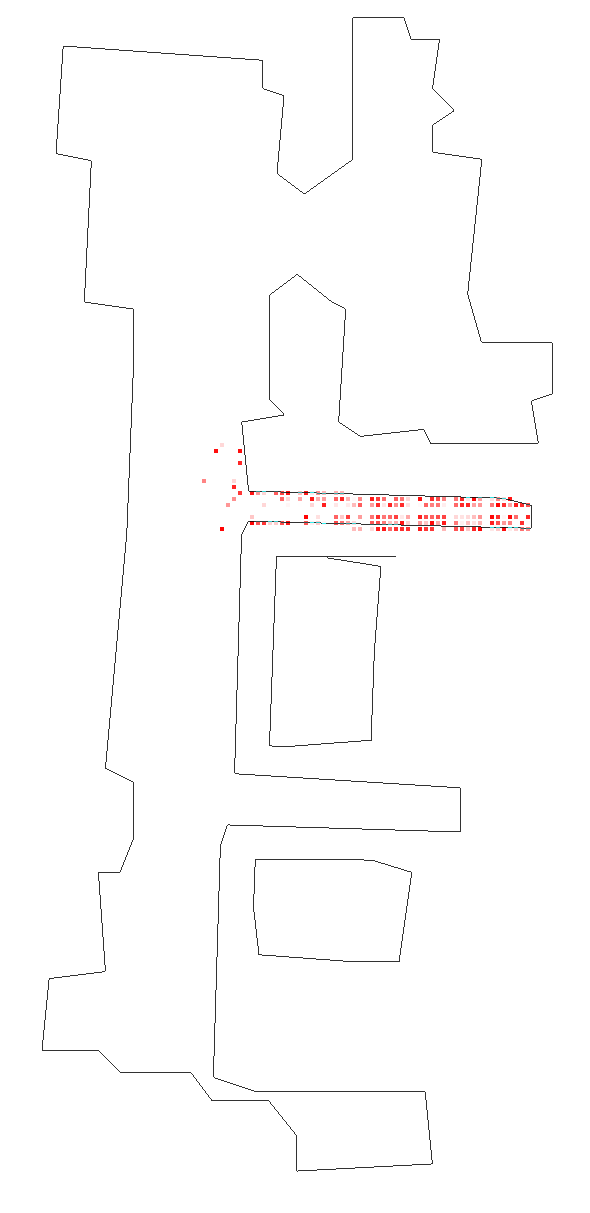}
\hfill
\includegraphics[width=0.155\linewidth,height=3.3cm]{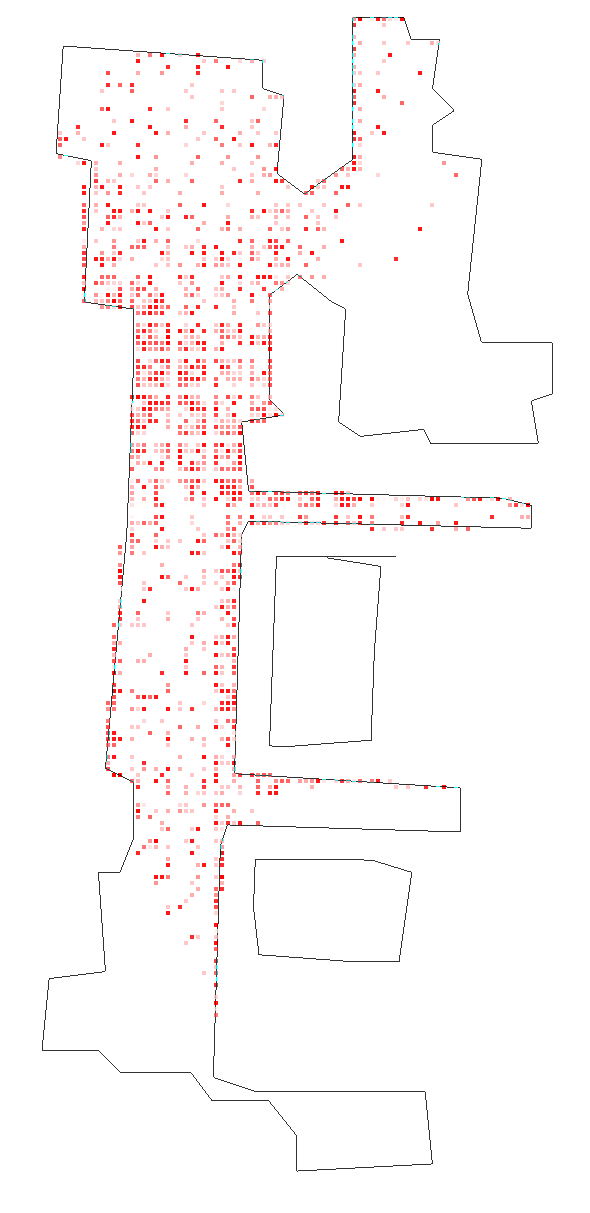}
\hfill
\includegraphics[width=0.155\linewidth,height=3.3cm]{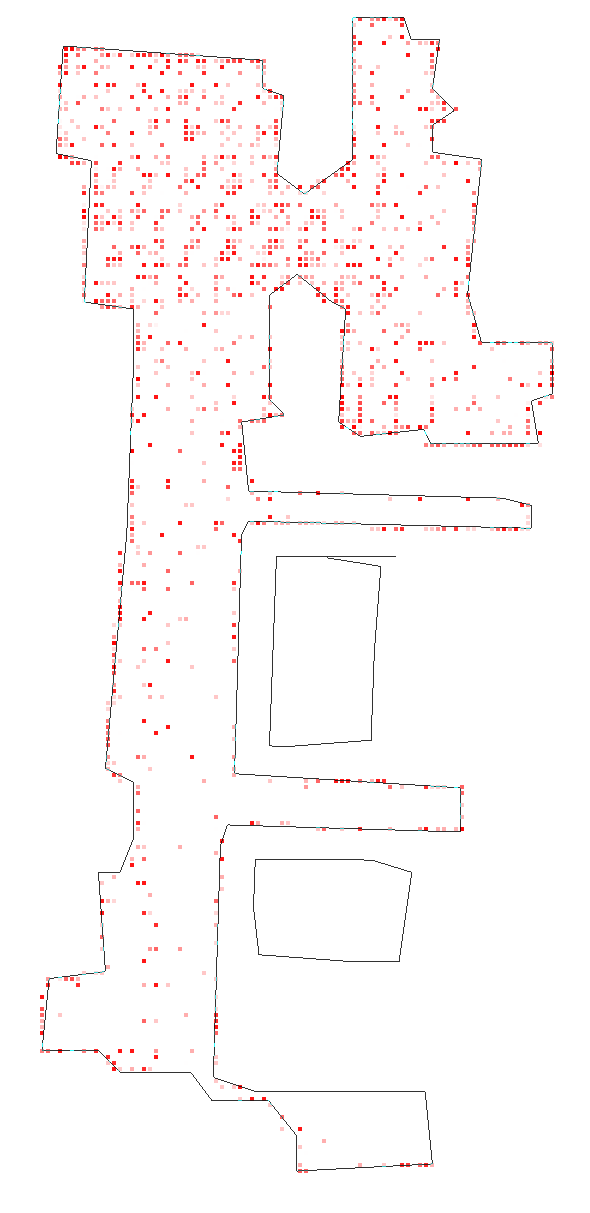}
\\[0.00em]
\raisebox{26pt}{\rotatebox[origin=t]{90}{\textbf{Office 3}}}
\includegraphics[width=0.155\linewidth]{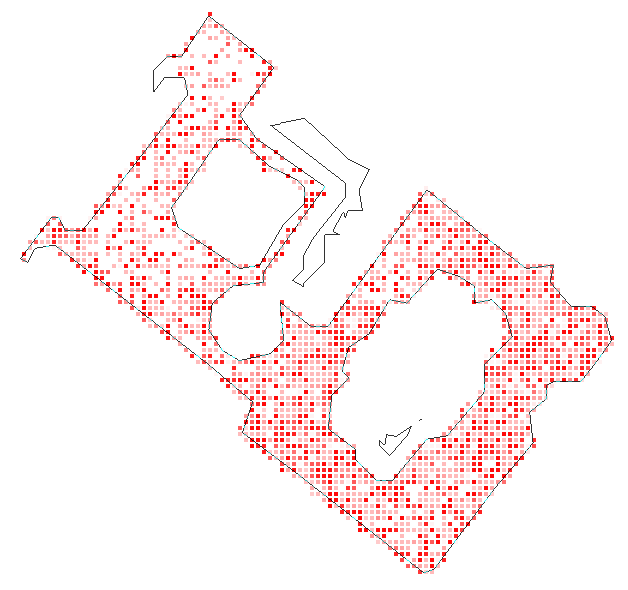}
\hfill
\includegraphics[width=0.155\linewidth]{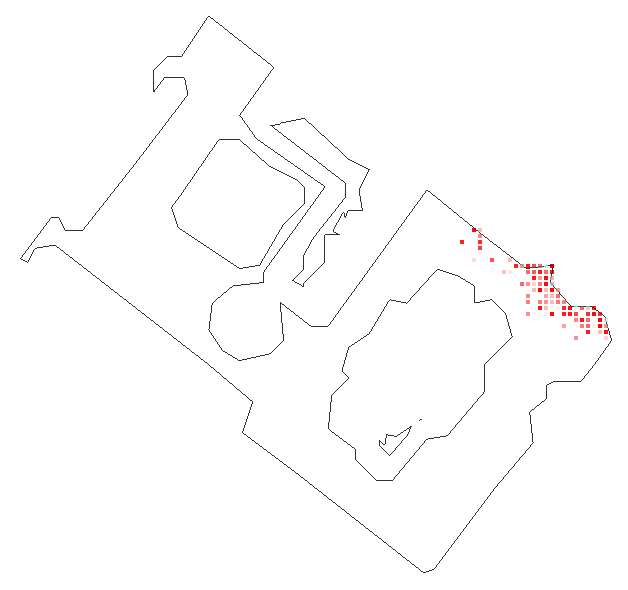}
\hfill
\includegraphics[width=0.155\linewidth]{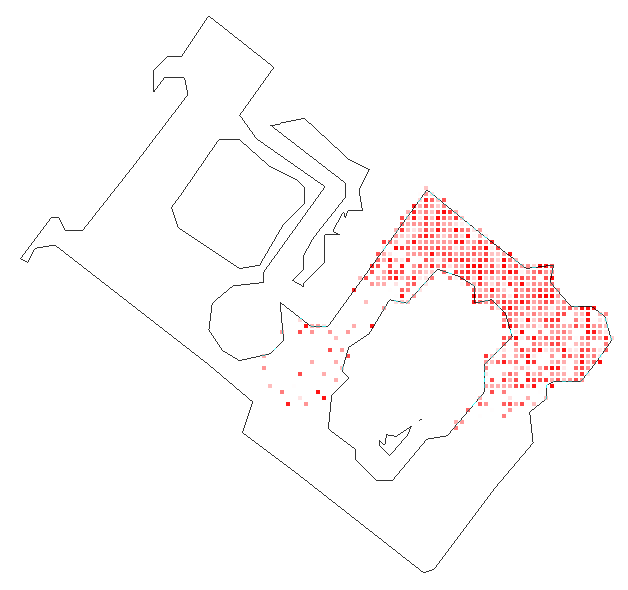}
\hfill
\includegraphics[width=0.155\linewidth]{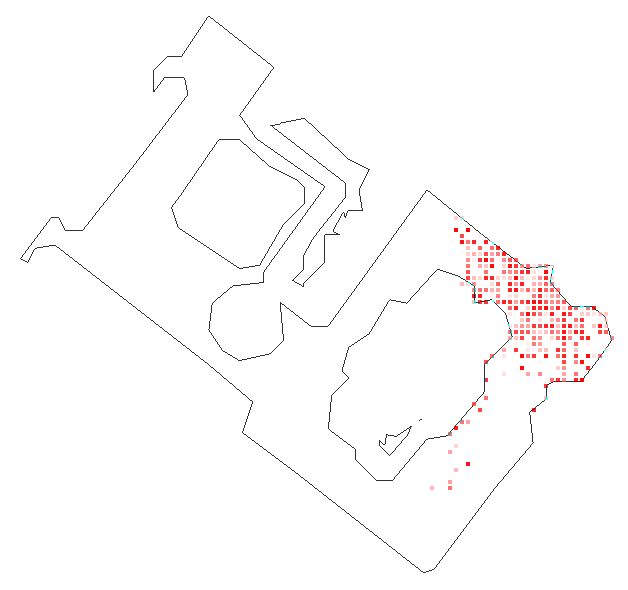}
\hfill
\includegraphics[width=0.155\linewidth]{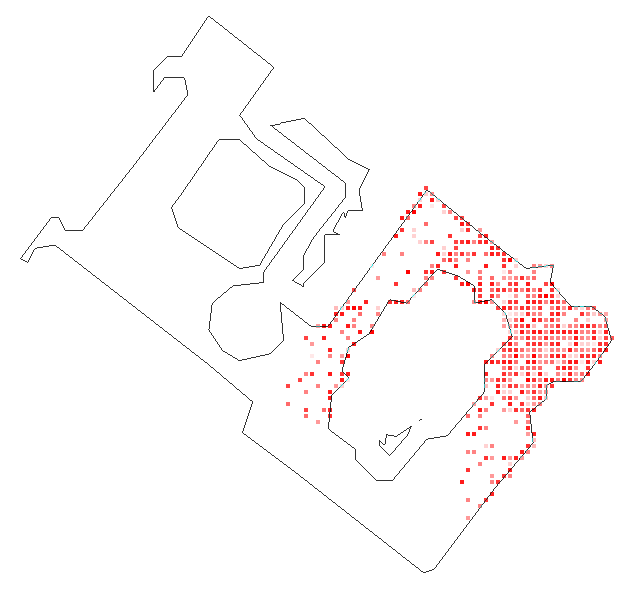}
\hfill
\includegraphics[width=0.155\linewidth]{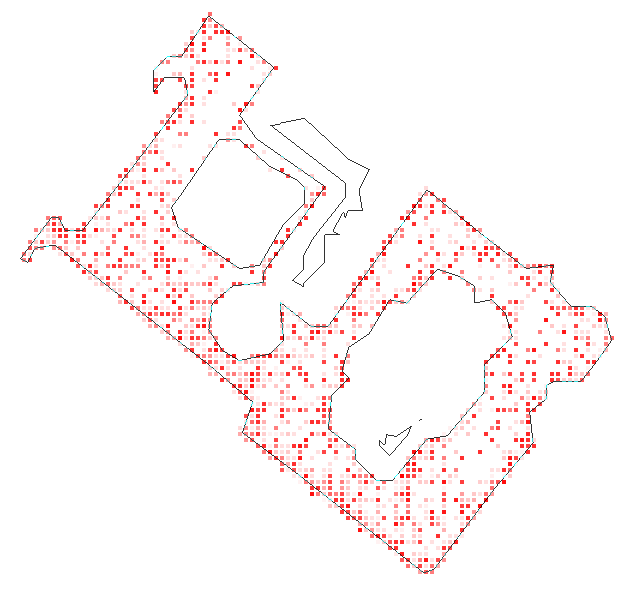}
\\[0.00em]
\raisebox{15pt}{\rotatebox[origin=t]{90}{\textbf{FRL Apt 0}}}
\includegraphics[width=0.155\linewidth,height=1.7cm]{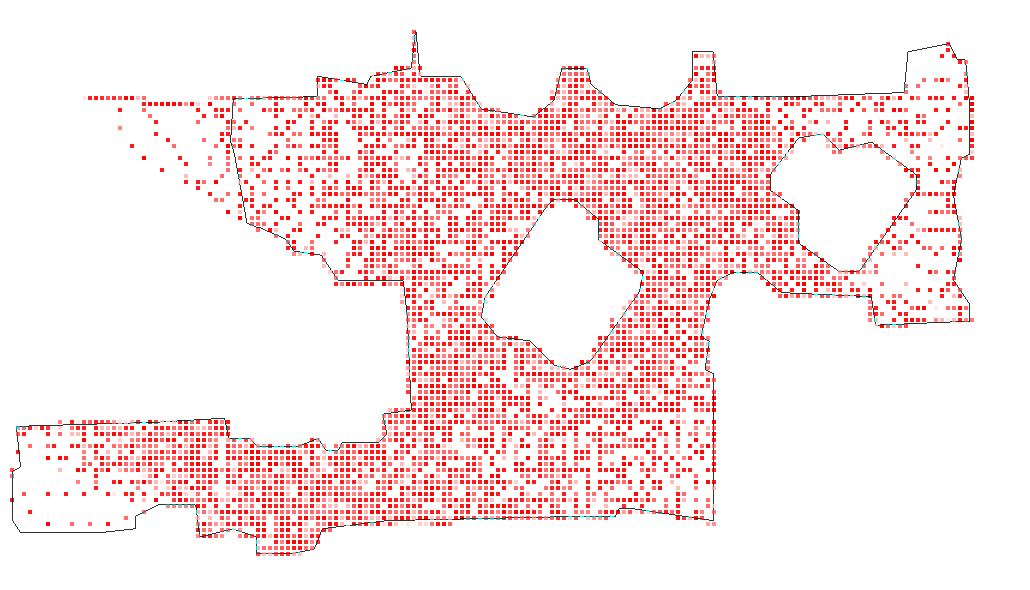}
\hfill
\includegraphics[width=0.155\linewidth,height=1.7cm]{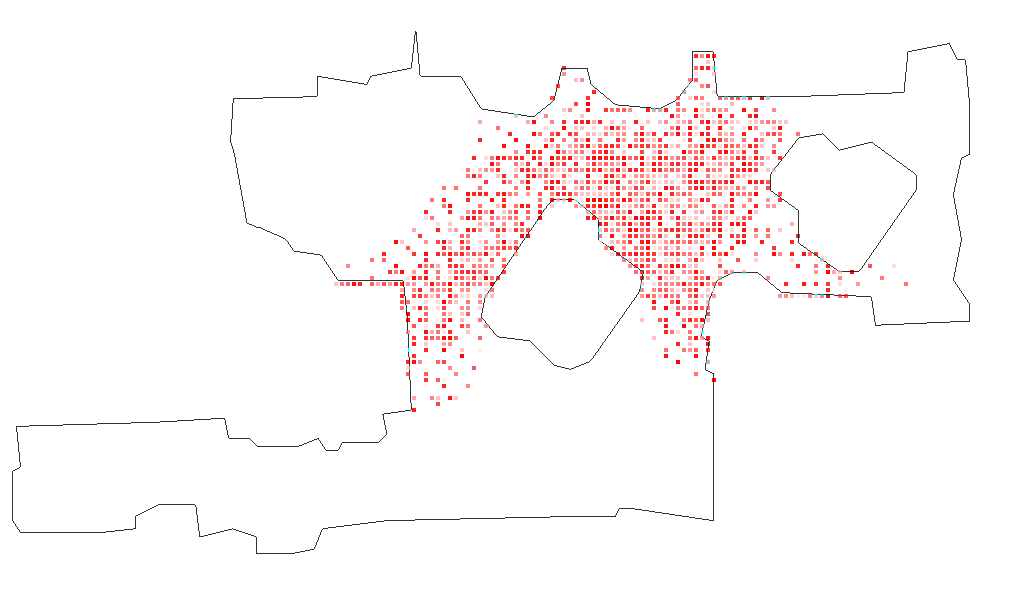}
\hfill
\includegraphics[width=0.155\linewidth,height=1.7cm]{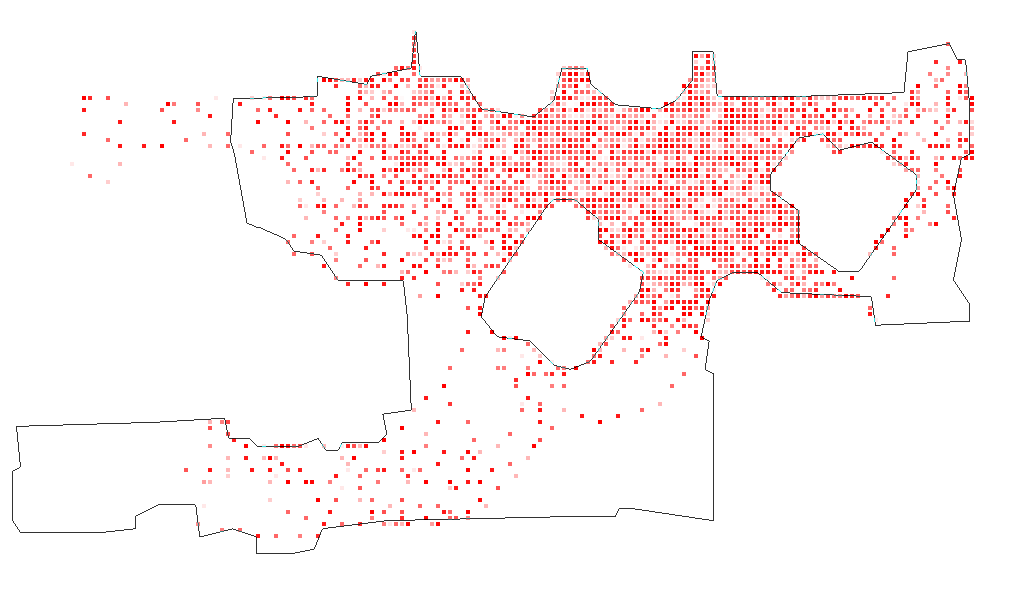}
\hfill
\includegraphics[width=0.155\linewidth,height=1.7cm]{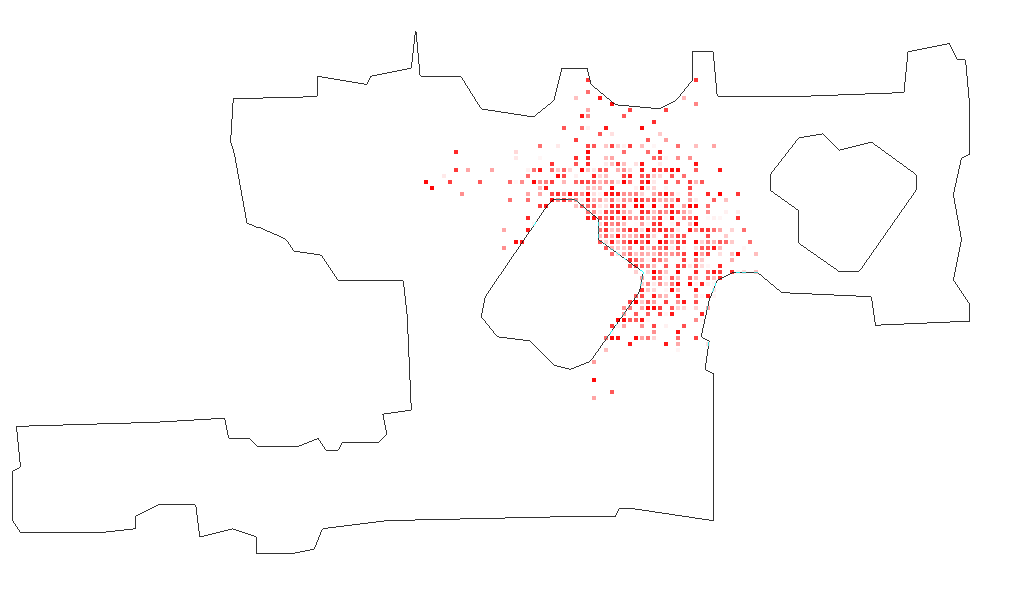}
\hfill
\includegraphics[width=0.155\linewidth,height=1.7cm]{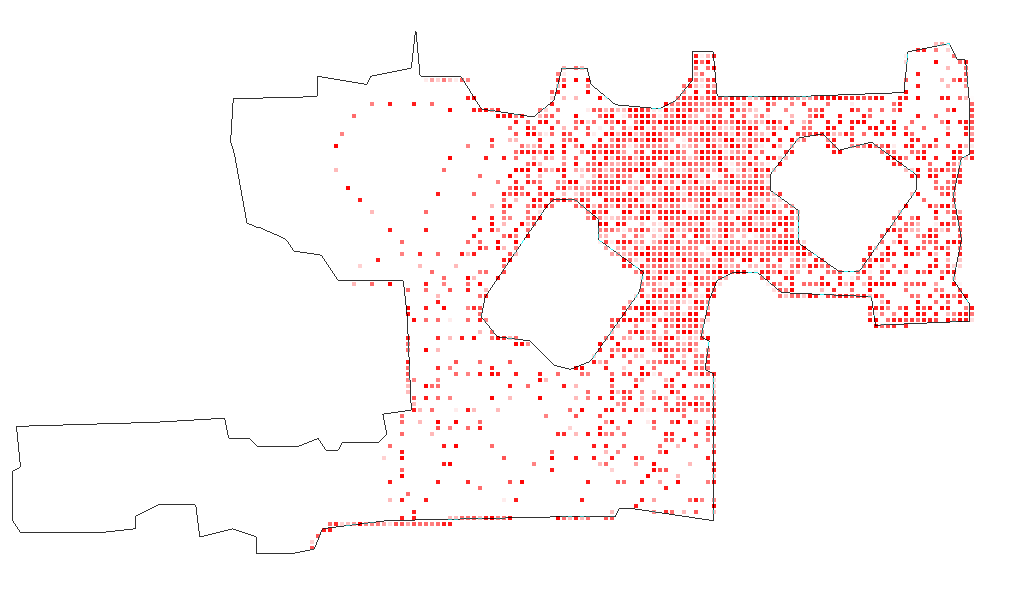}
\hfill
\includegraphics[width=0.155\linewidth,height=1.7cm]{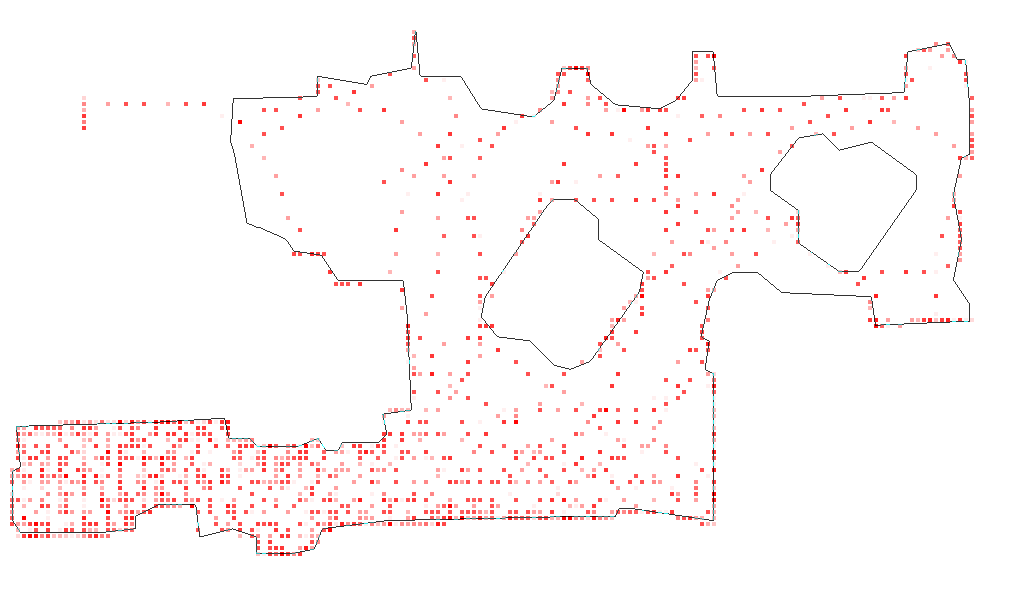}
\vspace*{-0.5em}
\caption{\label{fig:habitat_heatmap_test_all}
\textbf{Scene coverage} of exploration policies at offline transfer to new scenes. Each scene is explored for 100 episodes (50,000 total steps), with the agent always starting at the same location. Darker red cells denote higher visitation rates. C-BET outperforms baselines and exhibits great transfer to all scenes. It is the only algorithm visiting smaller scenes completely (Room 0, Room 1, Room 2, Office 3, FRL Apt. 0), and most of the bigger ones (Apt. 0, Apt. 2, Hotel 0) within only one episode. Furthermore, its visitation count is uniform in all scenes.}
\end{figure*}




\end{document}